\pdfoutput=1
\documentclass{article}
\usepackage[T2A]{fontenc}
\usepackage{amsmath,amssymb}
\usepackage[english]{babel}
\usepackage[colorlinks, linkcolor=black]{hyperref}
\usepackage[numbers]{natbib}
\usepackage[dvipsnames]{xcolor}
\usepackage{titlesec}
\usepackage{graphicx}
\usepackage{changepage}
\usepackage{booktabs}
\usepackage{enumitem}
\usepackage{multirow}
\usepackage[top=3cm, bottom=3cm, right=3cm, left=3cm]{geometry}
\usepackage{caption}
\usepackage{subcaption}
\definecolor{Green}{HTML}{00BD6B}

\usepackage{tgadventor}
\usepackage[T1]{fontenc}

\makeatletter
\renewcommand{\maketitle}{\bgroup\setlength{\parindent}{0pt}
\begin{flushleft}
  \textbf{\LARGE\textcolor{Green}{\fontfamily{qag}\selectfont \@title}}
  
  \vspace{0.3cm}
  {\fontfamily{qag}\selectfont \large \@author}
  \vspace{0.3cm}
\end{flushleft}\egroup
}
\makeatother

\title{Kandinsky 6.0 Video: Foundation Models for Synchronized Video and Audio Generation}
\author{{\color{Green}\textbf{Kandinsky Lab}$^*$} \\
\small $^*$ A detailed list of the contributors can be found at the end of this paper.}

\titleformat{\section}
  {\normalfont\sffamily\Large\bfseries\color{Green}}
  {\thesection}{1em}{}
\titleformat{\subsection}
  {\normalfont\sffamily\large\bfseries\color{Green}}
  {\thesubsection}{1em}{}
\titleformat{\subsubsection}
  {\normalfont\sffamily\normalsize\bfseries\color{Green}}
  {\thesubsubsection}{1em}{}
  
\begin{document}

\maketitle

\begin{figure}[h!]
    \centering
    \includegraphics[width=\textwidth]{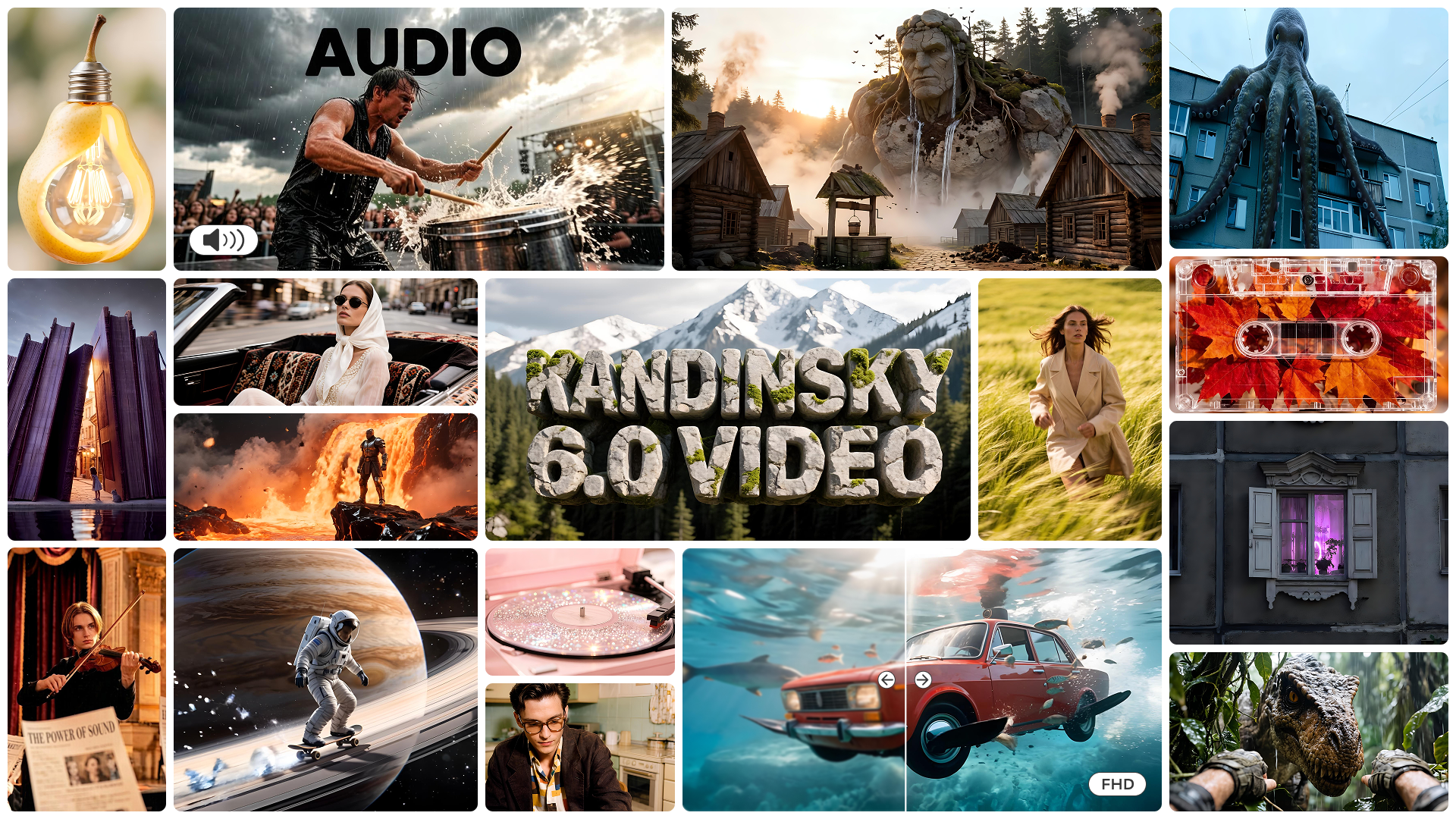}
\end{figure}

\noindent
\textbf{\textcolor{Green}{Abstract:}} We present Kandinsky~6.0 Video, a family of foundation diffusion models for synchronized text-to-audio-video generation, comprising Kandinsky~6.0 Video Lite (3B parameters) and Kandinsky~6.0 Video Pro (29B parameters). Both models generate 5-second video clips with synchronized 44~kHz audio, including lip-sync, in text-to-audio-video (T2AV) and image-to-audio-video (I2AV) modes; a built-in super-resolution model raises the output resolution to Full-HD (1920$\times$1080). Building on the video generation capabilities of Kandinsky~5.0, Kandinsky~6.0 Video employs a dual-stream CrossDiT architecture that connects a pretrained video stream and a newly trained audio stream through bidirectional cross-attention for temporal and semantic alignment. Our continuous pretraining strategy first trains the audio stream from scratch on large-scale audio corpora and then trains both streams jointly on paired audio-video data while preserving unimodal fidelity; pretraining is followed by supervised fine-tuning, reinforcement-learning-based post-training, and distillation. In side-by-side human evaluation, Kandinsky~6.0 Video Pro clearly outperforms its predecessor, Kandinsky~5.0 Video Pro, and remains competitive with leading audio-video generation models, outperforming LTX~2.5 and Seedance~2.0 in speech quality. To accelerate open research and deployment in multimedia generation, we release the code, model checkpoints, and diffusers integration under the MIT license.

\noindent
\textbf{\textcolor{Green}{Date:}} October 6, 2026

\noindent
\textbf{\textcolor{Green}{Website: }} \href{https://kandinskylab.ai}{https://kandinskylab.ai}

\noindent
\textbf{\textcolor{Green}{GitHub:}} \href{https://github.com/kandinskylab/kandinsky-6}{https://github.com/kandinskylab/kandinsky-6}

\noindent
\textbf{\textcolor{Green}{Hugging Face:}} \href{https://huggingface.co/collections/kandinskylab/kandinsky-60-diffusers}{https://huggingface.co/collections/kandinskylab/kandinsky-60-diffusers}

\newpage
\tableofcontents

\newpage

\section[Introduction]{\color{Green}Introduction}\label{sec:introduction}

Diffusion models~\cite{ho2020ddpm,song2021scorebased} and flow matching methods~\cite{lipman2023flow} have proven effective for image synthesis~\cite{rombach2022high, podell2024sdxl}, offering fine detail and controllability. Extending them to video generation has required specialized architectures based on diffusion transformers (DiT)~\cite{peebles2023dit} and efficient spatio-temporal attention mechanisms that model temporal dependencies without prohibitive computational cost. Today, text-to-video (T2V) generation is represented by both closed-source systems (Sora~\cite{openai2024sora}, Veo~\cite{Veo}) and open-source models (HunyuanVideo~\cite{kong2024hunyuanvideo}, Wan~\cite{wan2025}, Kandinsky~5.0~\cite{kandinsky5}). Kandinsky~5.0, in particular, combines the CrossDiT architecture with sparse NABLA attention~\cite{mikhailov2026nabla} to achieve high efficiency with modest computational resources.

Generating video with synchronized audio (text-to-audio-video, T2AV) is substantially more challenging than generating either modality alone, as it requires semantic, temporal, and emotional consistency across modalities. Closed commercial systems such as Veo~3.1~\cite{Veo}, Sora~2~\cite{openai2024sora2}, and Wan~2.6~\cite{wan2025} achieve impressive results in unified audio-visual synthesis, including natural lip synchronization, dynamic background sounds, and multilingual speech. However, their source code is unavailable, hindering reproducibility and further research.

Recent works target joint audio-video generation within a single architecture, including ALIVE~\cite{alive2025}, SyncFlow~\cite{syncflow2024}, JavisDiT~\cite{javis2025}, 3MDiT~\cite{3mdit2025}, LTX-2 and its successor LTX-2.5~\cite{ltx2025}, MiniMax-H3~\cite{MiniMax}, and NVIDIA Cosmos~3~\cite{cosmos3}. These models align the modalities temporally and semantically through mechanisms ranging from bidirectional cross-attention and hierarchical synchronization to flow-based architectures and omnimodal transformers. Considerable effort has also gone into scaling model size (to about 30B parameters and beyond), supporting additional input modes (image conditioning, multiple references), and developing standardized benchmarks for synchronization and quality, such as VABench~\cite{vabench2025}, T2AV-Compass~\cite{t2avcompass2025}, and AVGen-Bench~\cite{avgenbench2025}. Despite this progress, open-source solutions still lag behind closed ones in clip duration, resolution, and synchronization accuracy, keeping this area highly active and open to new approaches.

Continuing this research, \textbf{we present Kandinsky~6.0 Video, a family of foundation diffusion models for synchronized text-to-audio-video generation}. Our contributions are as follows:

\begin{itemize}
    \item We introduce two models of different scale: Kandinsky~6.0 Video Lite (3B parameters) and Kandinsky~6.0 Video Pro (29B parameters). Both generate 5-second video clips with synchronized 44~kHz audio, including lip-sync, and operate in text-to-audio-video (T2AV) and image-to-audio-video (I2AV) modes. Clips are generated in SD resolution and can be upscaled to Full-HD using a built-in super-resolution model.

    \item We employ a continuous pretraining strategy: the audio stream is first pretrained on large-scale audio corpora, then connected to the pretrained video stream through newly initialized cross-attention layers, and finally both streams are trained jointly on paired audio-video data.

    \item We describe a multi-stage post-training pipeline consisting of domain-specific supervised fine-tuning combined via model soup, reinforcement learning adapted from OmniNFT~\cite{zhang2026omninft}, and two-stage distillation ($\pi$-Flow followed by adversarial refinement) to 10 function evaluations.

    \item We evaluate Kandinsky~6.0 Video on VABench~\cite{vabench2025} and in side-by-side human evaluation. It outperforms the open-source LTX~2.5 on most VABench metrics, clearly surpasses its predecessor, Kandinsky~5.0 Video Pro, and remains competitive with proprietary systems, outperforming Seedance~2.0 in speech quality.

    \item We release all model weights, source code, and integration with the diffusers library to the research community under the MIT license.

\end{itemize}
\newpage

\section[Data Processing Pipeline]{\color{Green}Data Processing Pipeline}\label{sec:data_processing_pipeline}

The development of Kandinsky 6.0 Video required a substantial expansion of our data processing infrastructure and methodology. Building upon the foundations established for Kandinsky 5.0, we introduced a novel audio modality and constructed a comprehensive set of datasets for pretraining, fine-tuning, and downstream optimization. For pretraining, we prepared image (T2I), video (T2V), audio (T2A), joint audio-video (T2AV), and image-conditioned (I2AV) datasets to bootstrap the model across all modalities. For supervised fine-tuning, we developed dedicated SFT datasets for video and audio-visual tasks to enhance visual quality and achieve reliable audio-video synchronization. Additionally, we curated a Reinforcement Learning dataset for online optimisation, a Super-Resolution dataset to improve output detail, and an expanded Russian Cultural Code dataset to support culturally aware generation.

\subsection{Data Processing Infrastructure}
The core data processing infrastructure was upgraded: table partitioning and load-balancing mechanisms were enhanced to support concurrent operation of thousands of data processors, and vector indexes were extended to support clustering of audio-visual embeddings.

\subsection{Text-to-Image Dataset}

The Kandinsky T2I dataset for Kandinsky 6.0 Video is a 50-million-image subset of the Kandinsky 5.0 collection, re-annotated with a new captioner.

\subsubsection*{Dataset Properties}
The subset used for Kandinsky 6.0 Video inherits the characteristics of the original Kandinsky 5.0 filtering pipeline. It consists of high-quality, diverse images with minimal text content: watermarked and text-heavy scenes were removed by the original filters, and perceptual deduplication~\cite{zauner2010perceptualhashing} ensures broad visual variety. The dataset is balanced across object and scene categories (via YOLOv8~\cite{yolov8} and CLIP-based classifiers~\cite{radford2021clip}), and technical and aesthetic quality scores (TOPIQ~\cite{chen2023topiq}, Q-Align~\cite{wu2023qalign}) guarantee a high standard. Complexity filtering (SAM 2~\cite{ravi2024sam2}) retains visually rich and non-trivial compositions.

\subsubsection*{Content Annotation}

\begin{itemize}
    \item Captions are generated using \textbf{Qwen3-235B-A22B}~\cite{yang2025qwen3technicalreport}.
    \item Descriptions are produced in both \textbf{Russian and English}.
    \item Using \textbf{Qwen3-30B-A3B}, each description is rewritten into \textbf{4 levels of detail}: long, medium, short, and shortest.
\end{itemize}

\subsection{Text-to-Video Dataset}
For Kandinsky 6.0 Video, we selected a subset of 20 million video scenes from the original Kandinsky 5.0 T2V collection. Additional filtering was applied to ensure quality and temporal consistency: clips with non-standard frame rates or containing scene transitions (detected via TransNetV2~\cite{soucek2020transnetv2}) were discarded. All selected scenes were re-captioned with new, detailed captions.

\subsubsection*{Dataset Properties}
The subset inherits the quality characteristics of the Kandinsky 5.0 pipeline while incorporating stricter temporal filtering. It consists of stable, single-scene clips without internal scene breaks, ensuring coherent visual narratives, and with standardized frame rates. Watermarks and text-heavy scenes are excluded, and perceptual deduplication ensures diverse content. Technical and aesthetic quality is guaranteed by DOVER~\cite{wu2023exploring} and Q-Align scores, while YOLOv8 and CLIP-based classifiers maintain balance across object and scene categories. Camera and object dynamics (VideoMAE~\cite{wang2023videomaev2}) provide controlled motion diversity, and structural dynamics (MS-SSIM~\cite{wang2003msssim}) ensures sufficient motion without excessive redundancy.

\subsubsection*{Content Annotation}
For the video dataset, we applied the same captioning pipeline as for images. Captions are generated using \textbf{Qwen3-235B-A22B} in both Russian and English. Using \textbf{Qwen3-30B-A3B}, each description is rewritten into \textbf{4 levels of detail}: long, medium, short, and shortest. Post-processing includes cleaning of introductory phrases and filtering of non-English and non-Russian characters.

\subsection{Text-to-Audio Dataset}
For pretraining the audio branch, we constructed a separate audio-only dataset Kandinsky T2A. It consists of \textbf{40 million audio tracks} (5–20 seconds in length) extracted from the Kandinsky 5.0 video collection. Low-quality, silent, or corrupted tracks were discarded. Each track was annotated with a textual description of its audio content using \textbf{Qwen3-30B-A3B}, covering sounds, music, speech, and environmental acoustics. This dataset was used to bootstrap the audio modality before joint audio-video training.

\subsection{Text-to-Audio-Video Dataset}
For training joint video generation with sound, we constructed the Kandinsky T2AV dataset. It consists of \textbf{7 million audio-video segments} derived from the Kandinsky 5.0 video collection, filtered to ensure high-quality audio, meaningful audio--visual correspondence, and suitability for lip-sync modeling.

\subsubsection*{Dataset Properties}
The dataset was built by extracting audio tracks and applying a multi-stage filtering pipeline:
\begin{itemize}
    \item \textbf{Audio quality filters:} Scenes with silent, corrupted, or low-quality audio tracks were discarded. Remaining clips have sufficient dynamic range, low clipping ratio, and adequate amplitude.
    \item \textbf{Lip-sync and synchronization filters:} We applied two key metrics following the VABench benchmark~\cite{vabench2025}: Lip-Sync alignment score (LSA) and Desync. Clips with detected people were retained only if 
    their LSA was at least 3 on a scale from 0 to 5, and clips with audio-visual events were retained only if Desync was below 0.1 on a scale from 0 to 1. Clips without visible faces were kept as they do not require lip-sync (about 25\% of T2AV dataset). The combined filter ensures that retained clips either have good synchronization or contain no faces.
    \item \textbf{Visual quality:} The subset inherits the visual quality filters from the T2V pipeline (DOVER, Q-Align, CRAFT~\cite{baek2019craft}, YOLOv8, and CLIP-based classifiers), ensuring high-resolution, text-free, and diverse visual content.
\end{itemize}

\subsubsection*{Captioning Pipeline}
For joint audio-video training, we generated comprehensive captions describing both visual and audio content. The process involved:
\begin{itemize}
    \item Extracting speech transcripts using Gemma-4-E4B-it~\cite{gemmateam2026gemma4}, enclosed in \texttt{<S>...<E>} markers.
    \item Generating audio descriptions using Qwen3-30B-A3B, enclosed in \texttt{<AUDCAP> ... <ENDAUDCAP>} markers.
    \item Combining these components and feeding them into Qwen3.5-35B-A3B~\cite{qwen3.5}, which produced a unified detailed caption covering both the visual scene and its accompanying audio.
\end{itemize}

\subsection{Image-to-Audio-Video Dataset}
For the image-conditioned generation task (I2AV), we constructed a dedicated dataset. It consists of \textbf{116,000} specifically selected samples derived from the Kandinsky 5.0 video collection. Each sample is a pair: a starting frame (first frame of a video scene) and a corresponding combined caption describing both the visual content of that frame and the expected audio-visual continuation.

\subsubsection*{Captioning Pipeline}
Captions were generated using a two-stage local pipeline. First, audio descriptions were produced using \textbf{Gemma-4-E4B-it} and \textbf{Qwen3-Omni-30B-A3B}~\cite{xu2025qwen3omnitechnicalreport}. Second, these audio descriptions, together with camera motion classification labels, were fed into \textbf{Qwen3.5-35B-A3B} to produce a unified video+audio caption. All captions were subsequently rewritten into four detail levels (long, medium, short, shortest) using a dedicated rewriter.

\subsubsection*{Dataset Properties}
The dataset underwent additional filtering applied to the starting frames to ensure semantic and technical suitability for image-conditioned generation. The selection criteria included:
\begin{itemize}
    \item \textbf{Zero-shot embedding quality} (Qwen3-VL-Embedding-2B~\cite{li2026qwen3vlembedding}): overall quality, sharpness, exposure, artifacts.
    \item \textbf{No-reference image quality assessment} (using PyIQA library~\cite{chen2022pyiqa}): multiple metrics including clipiqa+, topiq\_nr, musiq-koniq, dbcnn, qalign, niqe, brisque, ilniqe, pi, and others.
    \item \textbf{Classical CV detectors}: blur detection via variance of Laplacian.
    \item \textbf{Semantic suitability for I2V}: \textbf{Qwen3-VL-8B-Instruct}~\cite{qwen3vl2025} verified that the frame represents a beginning or pause of an action (not a mid-action peak), that the main object is clearly visible and not cropped, and that the frame is not empty or transitional.
\end{itemize}
The filtering (union of all criteria) ensured that only high-quality, semantically appropriate starting frames were retained, resulting in a diverse and reliable dataset for both pretraining and subsequent fine-tuning stages.

\subsection{Supervised Fine-Tuning Datasets}
For supervised fine-tuning, we constructed two datasets: for video-only tasks and for audio-visual tasks. Both datasets share the same 11-domain taxonomy (see Table~\ref{tab:sft_domain_counts}) and are used in a two-stage training procedure: Stage~1 focuses on visual quality enhancement, Stage~2 on audio-video synchronization and lip-sync.
\begin{table}[ht]
\centering
\caption{Domain breakdown of SFT datasets for the audio-visual model.}
\label{tab:sft_domain_counts}
\begin{tabular}{@{} l r r @{}}
\toprule
\textbf{Domain} & \textbf{Stage 1 (Visual)} & \textbf{Stage 2 (Sync)} \\
\midrule
people action & 1,410 & 1,460 \\
animals & 591 & 840 \\
cartoons & 550 & 998 \\
people speech & 493 & 1,069 \\
music & 466 & 466 \\
food & 347 & 662 \\
nature & 214 & 668 \\
art & 159 & 307 \\
architecture & 147 & 474 \\
interiors & 147 & 357 \\
tech & 139 & 368 \\
\midrule
\textbf{Total} & \textbf{4,663} & \textbf{7,686} \\
\bottomrule
\end{tabular}
\end{table}

The datasets were built with the following common pipeline:
\begin{itemize}
    \item \textbf{Technical quality filtering:} Video scenes are filtered using DOVER, Q-Align, CRAFT, and motion dynamics models (same as in the main T2V pipeline). For audio, we apply Meta Audiobox Aesthetics~\cite{tjandra2025meta} (production quality, complexity, content enjoyment, usefulness) and discard silent or corrupted tracks.

    \item \textbf{Domain classification:} Scenes are classified into 11 domains (people action, animals, cartoons, people speech, music, food, nature, art, architecture, interiors, tech) using combined visual (Qwen2.5-VL-32B-Instruct~\cite{bai2025qwen25vltechnicalreport}) and audio (Qwen3-Omni-30B-A3B) embeddings.
    \item \textbf{Expert evaluation:} Two-stage evaluation is performed by technical screeners and cinematography experts, with emphasis on visual aesthetics, cinematographic quality, audio clarity, semantic audio-visual alignment, and lip-sync accuracy for speech scenes.
    \item \textbf{Captioning:} Audio descriptions are produced with Gemma-4-E4B-it and Qwen3-Omni-30B-A3B, while joint video+audio captions are generated by Qwen3.5-35B-A3B incorporating motion classification labels. All captions are subsequently rewritten into four detail levels: long, medium, short, shortest.
\end{itemize}

The domain breakdown for the datasets is given in Table~\ref{tab:sft_domain_counts}. Stage~1 (4,663 scenes) is primarily used for visual enhancement, while Stage~2 (7,686 scenes) expands the data with additional speech and music content to improve synchronization and lip-sync.

\subsection{Reinforcement Learning Dataset}
The RL dataset comprises \textbf{2,984 curated clips} sourced from the SFT collection. Each record contains T2AV-specific captions in English and Russian, with 100\% of samples containing an audio caption (\texttt{<AUDCAP>...<ENDAUDCAP>}). Precomputed text embeddings are provided for efficient online training. The dataset is used for online RL with per-prompt rollouts and multi-objective rewards.

\subsection{Super-Resolution Data}\label{sec:sr_data}

For super-resolution, we constructed a combined dataset comprising 150,000 unique clips from three sources:
\begin{itemize}
    \item \textbf{4K video collection}: 100,000 diverse
high-resolution video clips used to construct paired
HQ/LQ examples for super-resolution training.

    \item A high-resolution subset of the \textbf{Kandinsky~5.0 T2V} video collection, comprising approximately 35,000 clips, selected through artifact filtering and Q-Align-based quality filtering.

    \item 15,000 synthetic camera-motion clips constructed from
    high-resolution images from the \textbf{Aesthetic-4K}
    dataset~\cite{zhang2025diffusion4k}. Each clip is produced by moving a spatial crop along a sampled camera trajectory over a still image, simulating camera motion.
\end{itemize}

\paragraph{Quality filtering.}
We filter the two video datasets using different quality checks.
For the Kandinsky~5.0 T2V collection, we check for compression artifacts
and visible bands in smooth color gradients, and use Q-Align
scores to select high-quality examples. For the 4K video collection, we combine Q-Align scores with a separate ripple detector to remove low-quality examples and those with strong noise-like
patterns.

The preparation of paired HQ/LQ training examples, including
crop selection, online filtering, and degradation pipelines,
is described in Section~\ref{sec:sr_degradations}.

\subsection{Russian Cultural Code Dataset}
For Kandinsky 6.0 Video, we expanded and updated the Russian Cultural Code
(RCC) dataset~\cite{vasilev2024craft,vasilev2025ruscode}. The dataset comprises
\textbf{1.2M images} and \textbf{530K video scenes}, specifically
curated to represent the Russian cultural code --- covering distinctive
features related to faith, language, historical memory, nature, architecture, and other culturally significant elements. All samples are
annotated with accurate object names and comprehensive scene descriptions,
ensuring high-quality data for culturally aware generation. The dataset is
used during the T2V pre-training phase and is also injected through images
during joint T2AV training, in both cases with English and Russian captions,
to support the model's ability to generate content that aligns with Russian
cultural context.

%
\section[Kandinsky 6.0 Video Architecture]{Kandinsky 6.0 Video Architecture}\label{sec:architecture}

\subsection{Overview}

Kandinsky~6.0 Video is released in two versions: Kandinsky~6.0 Video Lite (3B parameters) and Kandinsky~6.0 Video Pro (29B parameters). Both versions share the same architecture and are trained with flow matching~\cite{lipman2023flow}, so we describe it in general terms below and point out version-specific differences where they apply.

The core of the model is a \textbf{CrossDiT} network that inherits the architecture of Kandinsky~5.0~\cite{kandinsky5} and extends it to two modalities, video and audio. This core component is described in detail in Section~\ref{sec:core_model_architecture}.

Both versions use the Hunyuan VAE~\cite{kong2024hunyuanvideo} for video encoding and decoding and the MMAudio autoencoder~\cite{mmaudio}, which supports a sampling rate of 44~kHz, for audio. Text is encoded with Qwen2.5-VL-7B-Instruct~\cite{bai2025qwen25vltechnicalreport} and CLIP ViT-L/14~\cite{radford2021clip}.

\begin{figure}[ht]
    \centering
    \includegraphics[width=\textwidth]{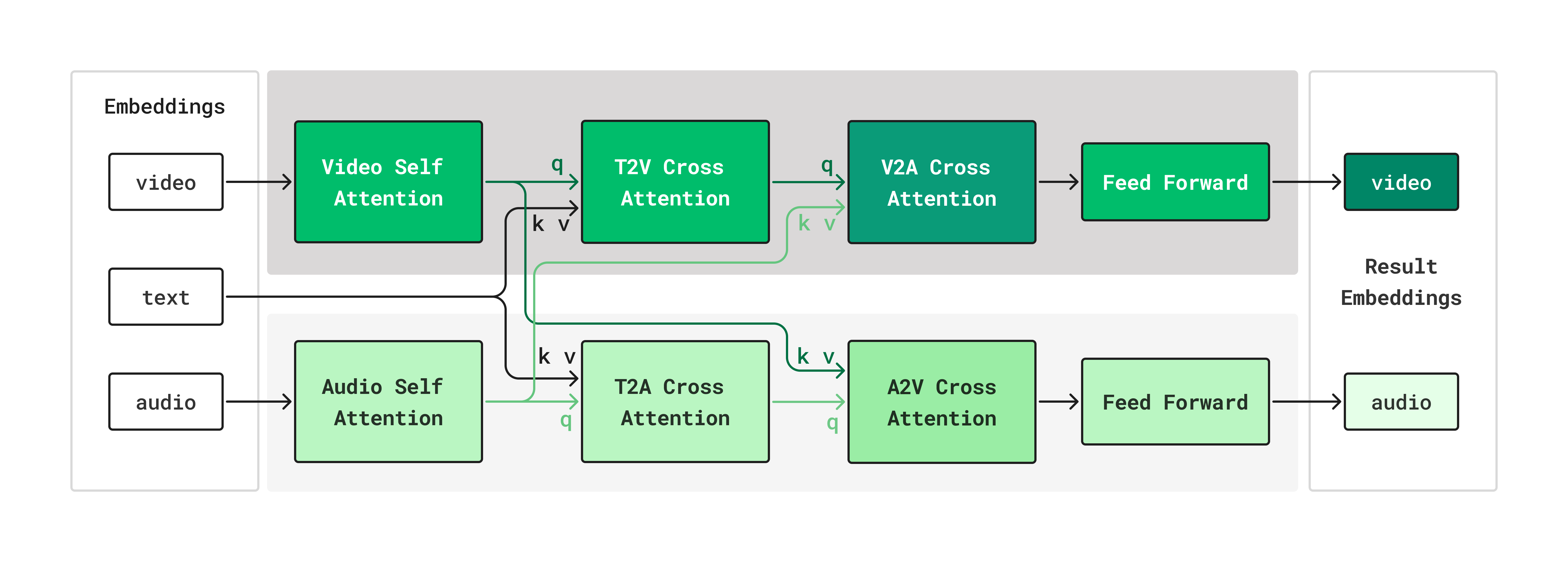}
    \caption{Simplified scheme of a single Kandinsky~6.0 Video transformer block. The video (top) and audio (bottom) streams each apply self-attention, cross-attention to the text embeddings (T2V and T2A), cross-modal attention, and a feed-forward layer. In the V2A block, video queries attend to audio keys and values; in the A2V block, audio queries attend to video keys and values.}
    \label{fig:CrossDiTFusedBlock}
\end{figure}

The \textbf{super-resolution} model, which is applied to the output of the trained model to obtain Full-HD resolution, is described in Section~\ref{sec:SR_architecture}.

\subsection{Core Model}\label{sec:core_model_architecture}

\begin{figure}[h!]
    \centering
    \includegraphics[width=0.93\textwidth]{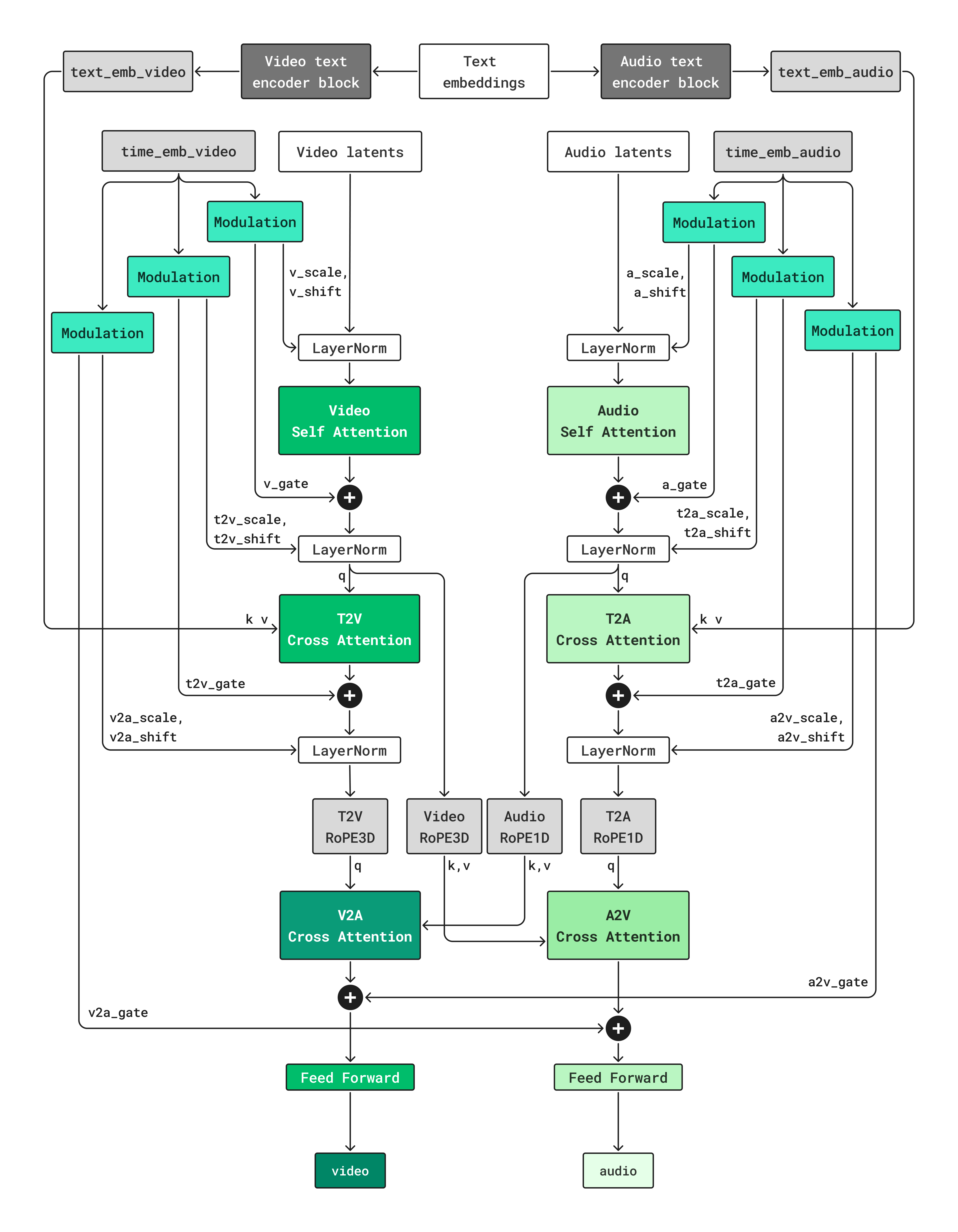}
    \caption{Detailed scheme of the Kandinsky~6.0 Video transformer block. Each stream has its own text encoder block and timestep embedding; the timestep embedding modulates every sub-layer through scale, shift, and gate parameters. In the cross-modal attention layers, video tokens use 3D RoPE and audio tokens use 1D RoPE.}
    \label{fig:CrossDiTFusedBlock_detailed}
\end{figure}

The core of Kandinsky~6.0 Video is a dual-stream CrossDiT architecture that processes the video and audio modalities, similar to Ovi~\cite{low2025ovi} and LTX-2~\cite{ltx2025}. It comprises asymmetric DiT backbones for video and audio, connected by blockwise bidirectional cross-attention.

Figure~\ref{fig:CrossDiTFusedBlock} shows a simplified scheme of a single block with both streams and the cross-attention that connects them; a detailed diagram is given in Figure~\ref{fig:CrossDiTFusedBlock_detailed}.

Each stream inherits its architecture from Kandinsky~5.0~\cite{kandinsky5}. The two streams have the same number of transformer blocks but differ in hidden and feed-forward (FFN) dimensions and, consequently, in parameter count.

In Kandinsky~6.0 Video Lite (3B parameters in total), the video stream has 2B parameters, the audio stream 0.6B, and the cross-attention layers between the streams 0.4B. In Kandinsky~6.0 Video Pro (29B in total), the video stream has 19B parameters, the audio stream 5B, and the cross-attention layers 5B.

Positions are encoded with Rotary Position Embedding (RoPE)~\cite{su2023roformerenhancedtransformerrotary}: video tokens are indexed by frame number and audio tokens by temporal position. For video, the frame index is normalized by the frame rate relative to 24~fps:
\[
\text{RoPE}\left(\frac{N_{\text{frame}}}{\text{fps}/24}\right),
\]
where $N_{\text{frame}}$ is the input frame index and $\text{fps}$ is the frames per second of the input video.

Table~\ref{tab:model_details} lists the number of blocks and other architectural hyperparameters for both versions at each training stage (see Section~\ref{sec:training}).

\begin{table}[htbp]
\centering
\caption{Architectural hyperparameters of Kandinsky~6.0 Video Lite and Pro (values given as Lite\,/\,Pro) at each training stage: video-only (T2V) and audio-only (T2A) pre-training, joint text-to-audio-video (T2AV) training, and image-to-audio-video (I2AV) training. The Qwen2.5-VL and CLIP text encoders are the same in all configurations.}
\label{tab:model_details}
\resizebox{\textwidth}{!}{%
\begin{tabular}{l | c c c c}
\toprule
Parameter & T2V (Lite/Pro) & T2A (Lite/Pro) & T2AV (Lite/Pro) & I2AV (Lite/Pro) \\
\midrule
& \multicolumn{4}{c}{\textbf{CrossDiT}} \\
\midrule
video\_model\_dim & 1792 / 4096 & -- & 1792 / 4096 & 1792 / 4096 \\
audio\_model\_dim & -- & 896 / 2048 & 896 / 2048 & 896 / 2048 \\
ff\_dim & 7168 / 16384 & 3584 / 7168 & 3584 (audio), & 3584 (audio), \\
 & & & 7168 (video) / & 7168 (video) / \\
 &  & & 7168 (audio), & 7168 (audio), \\
  &  & & 16384 (video) & 16384 (video) \\
num\_visual\_blocks & 32 / 60 & -- & 32 / 60 & 32 / 60 \\
num\_audio\_blocks & -- & 32 / 60 & 32 / 60 & 32 / 60 \\
num\_text\_blocks & 2 / 4 & 2 / 4 & 2 / 4 & 2 / 4 \\
in\_visual\_dim & 16 & 16 & 16 & 16 \\
out\_visual\_dim & 16 & 16 & 16 & 16 \\
in\_audio\_dim & -- & 40 & 40 & 40 \\
time\_dim & 512 / 1024 & 512 / 1024 & 512 / 1024 & 512 / 1024 \\
patch\_size & (1, 2, 2) & (1, 2, 2) & (1, 2, 2) & (1, 2, 2) \\
head\_dims\_video & (16, 24, 24) / & -- & (16, 24, 24) / & (16, 24, 24) / \\
 & (32, 48, 48) & -- & (32, 48, 48) & (32, 48, 48) \\
head\_dim\_audio & -- & 64 / 128 & 64 / 128 & 64 / 128 \\
\midrule
& \multicolumn{4}{c}{\textbf{Qwen}} \\
\midrule
version & \multicolumn{4}{c}{Qwen2.5-VL-7B-Instruct} \\
max\_length & \multicolumn{4}{c}{1024} \\
emb\_size & \multicolumn{4}{c}{3584} \\
\midrule
& \multicolumn{4}{c}{\textbf{CLIP}} \\
\midrule
max\_length & \multicolumn{4}{c}{77} \\
emb\_size & \multicolumn{4}{c}{768} \\
\bottomrule
\end{tabular}%
}
\end{table}

\newpage

\subsection{Super-Resolution Model}\label{sec:SR_architecture}

To obtain Full-HD output, videos generated by Kandinsky~6.0 Video
are upscaled by a separate video super-resolution (SR) model.
Unlike the base generator, which uses the Hunyuan VAE, the SR model
operates in the latent space of
K-VAE~\cite{shutkin2026kvaefamilytokenizersmultimodal},
a video autoencoder with 64 latent channels and $16\times$ spatial
and $4\times$ temporal compression.
For spatially aligned inputs with $T = 4n + 1$ frames, a clip of
$T \times H \times W$ pixels is encoded into a latent of size
$T' \times \frac{H}{16} \times \frac{W}{16} \times 64$,
where $T' = (T-1)/4 + 1$.
Thus, a 121-frame clip corresponds to 31 latent frames.

The SR model consists of two components trained separately:
\begin{itemize}
    \item the \textbf{Latent Upscaler} (LU), a convolutional network
    that deterministically upsamples a low-quality (LQ) latent
    by a factor of $2$ or $4$ to the working resolution of the
    SR-DiT (Section~\ref{sec:latent_upscaler_arch});
    \item the \textbf{SR-DiT}, a flow-matching diffusion transformer
    that removes degradations and restores detail in the
    upsampled latent (Section~\ref{sec:sr_dit_arch}).
\end{itemize}
A single SR-DiT is trained jointly for the $\times 2$ and $\times 4$
routes, while the LU uses scale-specific branches.

In our inference pipeline, the SR-DiT processes spatial tiles (height $\times$ width)
corresponding to $512 \times 768$~px for landscape content,
$768 \times 512$~px for portrait content, and
$512 \times 512$~px for square content.
These correspond to latent grids of $32 \times 48$,
$48 \times 32$, and $32 \times 32$ cells, respectively,
with 64 channels.
For a fixed scale and working tile size, larger input videos
require more tiles (Section~\ref{sec:inference}).
The SR model supports scale factors of $\times 2$ and $\times 4$;
the $\times 2.25$ route used for Full-HD output combines bilinear
$\times 1.125$ pre-upscaling in pixel space with the $\times 2$
route.

\subsubsection{SR-DiT}\label{sec:sr_dit_arch}

\begin{figure}[h!]
    \centering
    \includegraphics[width=\textwidth]{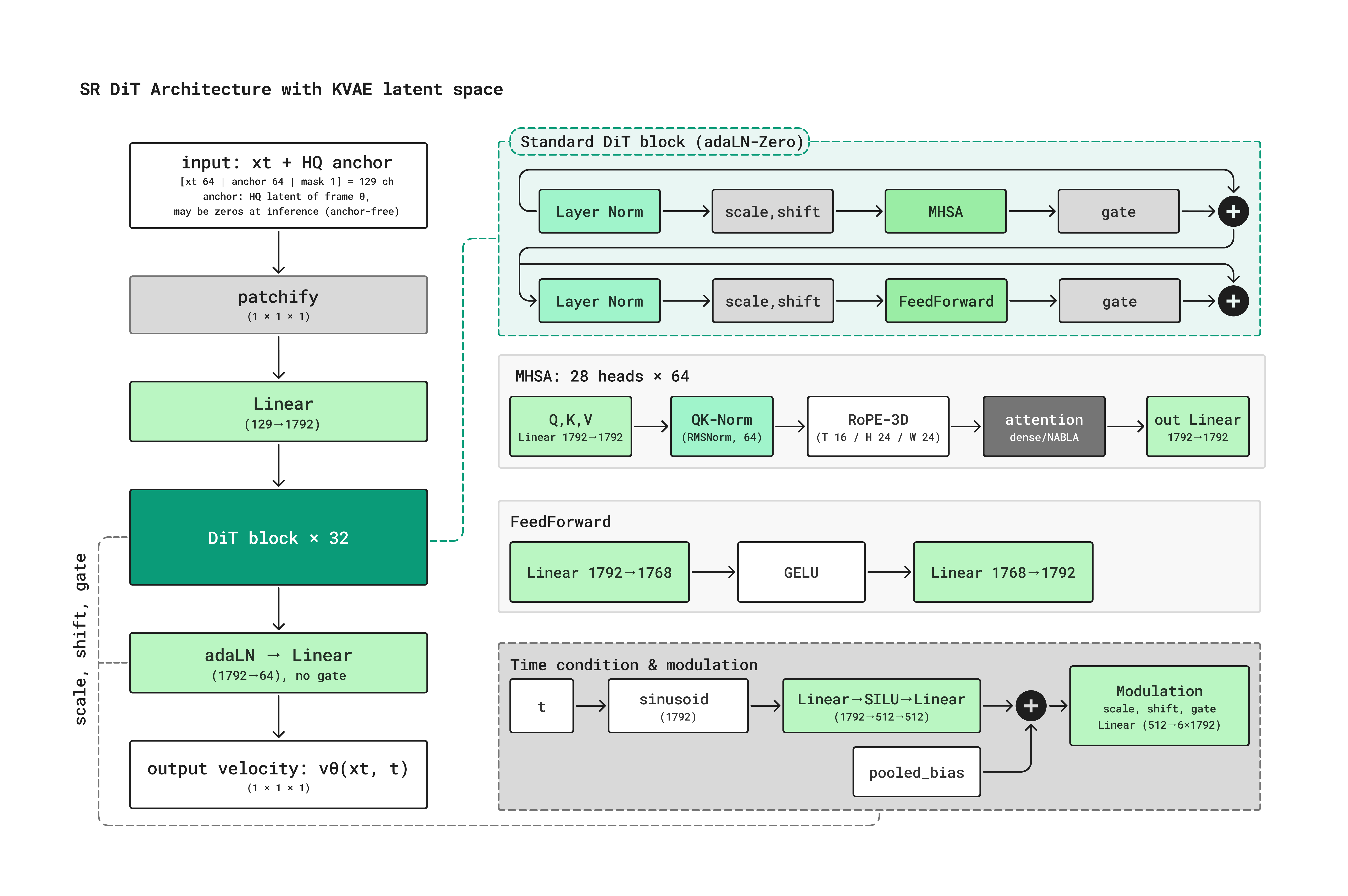}
    \caption{
    SR-DiT architecture. The noisy video latent, first-frame HQ anchor, and binary anchor mask are concatenated along the channel dimension and processed by timestep-modulated DiT blocks. The output head predicts the flow-matching velocity. The model operates without text conditioning.
    }
    \label{fig:sr_dit}
\end{figure}

The SR-DiT (Figure~\ref{fig:sr_dit}) is based on
Kandinsky~5.0 Video Lite~\cite{kandinsky5}.
It is initialized from a text-free Kandinsky~5.0 Video Lite
(SFT) checkpoint operating in the K-VAE latent space.
The model has 32 transformer blocks with a hidden dimension
of 1792, 28 attention heads of dimension 64, an FFN dimension
of 7168, and a timestep embedding dimension of 512.
The backbone with the standard velocity head contains
approximately 1.41B parameters.

\paragraph{Text-free design.}
The SR-DiT does not use text encoders or text cross-attention
and requires no prompt.
The pooled text contribution to the timestep conditioning
is replaced with a learnable constant (\emph{pooled bias}).
Each block uses six modulation vectors: scale, shift, and gate
for self-attention and for the FFN, instead of the nine vectors
used by the text-conditioned architecture.
A SiLU activation followed by a linear projection
$512 \to 6 \times 1792$ produces these vectors.
The pretrained modulation weights are inherited from the
initialization checkpoint.

\paragraph{Input and tokenization.}
The model receives a 64-channel noisy latent $x_t$ together
with a 64-channel anchor tensor and a binary anchor mask.
When an HQ anchor is provided, its latent occupies the first
temporal position of the anchor tensor, and the mask is one
at that position; the remaining positions are zero.
When the anchor is absent, both the anchor tensor and its
mask are zero.
The concatenated input
$[x_t \,|\, \text{anchor} \,|\, \text{mask}]$
therefore always has $64 + 64 + 1 = 129$ channels.

The patch size is $(1, 1, 1)$, so each token corresponds to
one latent cell, at a spatial stride of 16~px.
A $512 \times 768$ tile yields 1536 tokens per latent frame,
while a $512 \times 512$ tile yields 1024.
A linear projection $129 \to 1792$ maps the input tokens
to the model width.
The $\times 2$ and $\times 4$ routes use the same SR-DiT
weights and input dimensions, with their inputs prepared
by the corresponding LU branches.

\paragraph{Transformer block.}
Each block contains self-attention and an FFN with residual
connections, LayerNorm without learnable affine parameters,
and adaLN-Zero modulation.
Queries and keys are normalized with RMSNorm within each head.
The model uses 3D RoPE, with the head dimension split into
$(16, 24, 24)$ channels for the temporal, height, and width axes,
corresponding to 8, 12, and 12 rotary frequency pairs.
The positional rotations are shared across all 32 blocks.

Attention is dense (FlashAttention) in the first training stage
and is replaced with sparse NABLA
attention~\cite{mikhailov2026nabla} during the second stage
(Section~\ref{sec:sr_training}).
The FFN consists of two bias-free linear layers,
$1792 \to 7168 \to 1792$, with a GELU activation between them.
It accounts for approximately 58\% of each block's parameters.

\paragraph{Output head and parameter count.}
Before distillation, the output head consists of a LayerNorm
modulated by the timestep conditioning through scale and shift,
followed by a linear projection $1792 \to 64$ that predicts
the velocity.
Each transformer block contains approximately 44.06M parameters:
5.52M in modulation, 12.85M in attention, and 25.69M in the FFN.
Including the input projection, timestep embedding, pooled bias,
and output head, the model contains approximately 1.413B
parameters.

For $\pi$-Flow DX distillation, the final projection is expanded
to $1792 \to 640$, producing ten 64-channel outputs per token
for constructing the DX policy
(Section~\ref{sec:sr_training}).
The transformer backbone is unchanged, and the expanded head
adds approximately 1.03M parameters.
Figure~\ref{fig:sr_dit} shows the architecture with the standard
velocity head.

\subsubsection{Latent Upscaler}\label{sec:latent_upscaler_arch}

\begin{figure}[h!]
    \centering
    \includegraphics[width=\textwidth]{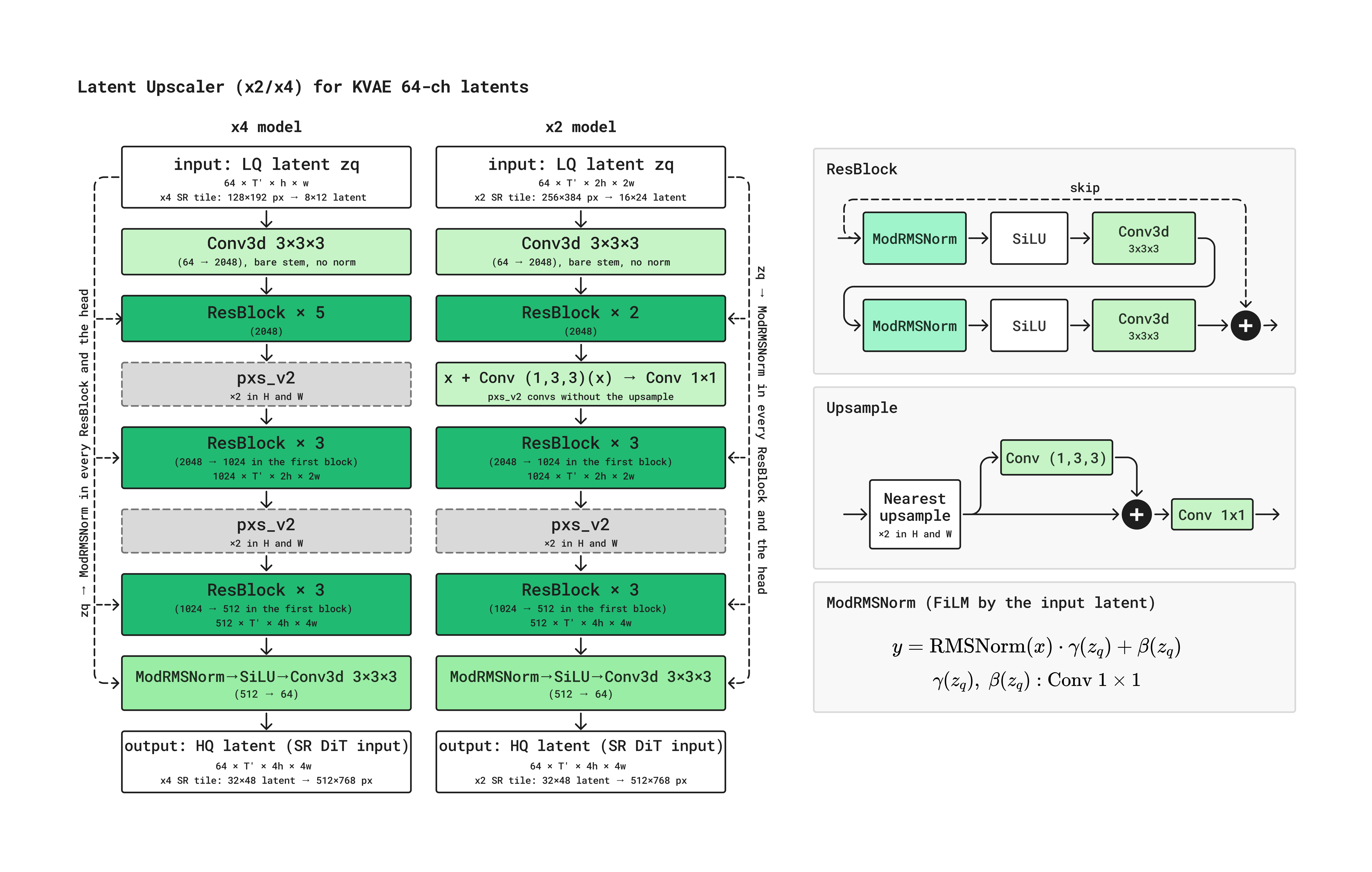}
    \caption{
    Latent Upscaler architecture: the $\times 4$ branch (left), the $\times 2$ branch (middle), and their building blocks (right). Both branches follow the K-VAE decoder geometry and map 64-channel LQ latents to upsampled 64-channel latents, preserving the temporal resolution. The outputs labeled ``HQ latent'' refer to the higher-resolution
latent grid; they retain degradations, which are subsequently
removed by SR-DiT.
    }
    \label{fig:latent_upscaler}
\end{figure}

The Latent Upscaler (LU; Figure~\ref{fig:latent_upscaler}) brings the LQ latent to the working resolution of the SR-DiT directly in latent space, without decoding to pixels. Its architecture is derived from the K-VAE decoder: the $\times 4$ model follows the decoder's channel widths, residual blocks, modulated normalization, and spatial upsampling modules over two upsampling stages. Unlike the decoder, it outputs 64-channel latents instead of pixels and preserves the temporal resolution. The $\times 2$ branch is constructed from the trained $\times 4$ model with a shorter front section and one spatial upsampling stage.

The LU replaces pixel-space bilinear resizing followed by K-VAE encoding of each upscaled tile. Instead, the source video is encoded once, and spatial tiles are extracted and upscaled directly in latent space. This avoids separately encoding each upscaled pixel tile and reduces inference time (Section~\ref{sec:inference}).

For a landscape tile, the LU maps an $8 \times 12$ latent grid ($128 \times 192$~px of the source video) to $32 \times 48$ for the $\times 4$ route, and a $16 \times 24$ grid ($256 \times 384$~px) to $32 \times 48$ for the $\times 2$ route.

\paragraph{$\times 4$ model.}
The architecture follows the widths and block counts of the
K-VAE decoder over the part spanning two spatial upsampling
stages.
The input stem, residual blocks, and upsampling modules are
initialized from the decoder; the output heads are initialized
randomly.
The $\times 4$ inference path consists of:
\begin{itemize}
    \item an input stem: a $3 \times 3 \times 3$ convolution
    $64 \to 2048$ without normalization or activation
    (3.5M parameters);
    \item five ResBlocks at width 2048, operating at the input
    resolution (1135.2M parameters);
    \item the first $\times 2$ spatial upsampling module
    (41.9M);
    \item three ResBlocks, with the first reducing the width
    from 2048 to 1024 (201.2M);
    \item the second $\times 2$ spatial upsampling module
    (10.5M);
    \item three ResBlocks, with the first reducing the width
    from 1024 to 512 (50.5M);
    \item an output head consisting of ModRMSNorm, SiLU,
    and a $3 \times 3 \times 3$ convolution $512 \to 64$
    (1.0M).
\end{itemize}
These modules contain approximately 1.44B parameters,
of which about 79\% belong to the five input-resolution
ResBlocks.

Each ResBlock consists of two ModRMSNorm--SiLU--Conv3d
units with a residual connection and no bottleneck.
Blocks that reduce the channel width use a
$1 \times 1 \times 1$ projection in the residual connection.
All normalization layers, including the one in the output
head, are modulated by the input LQ latent $z_q$.
The conditioning latent is resized with nearest-neighbor
interpolation to match the current feature grid, and
$1 \times 1 \times 1$ convolutions produce the modulation
scale and shift.

Each spatial upsampling module doubles the height and width
using nearest-neighbor resizing, followed by a residual
$(1, 3, 3)$ convolution and a $1 \times 1 \times 1$ convolution.
The $3 \times 3 \times 3$ convolutions use symmetric temporal
padding with replicated edge frames.
When transferring weights from the causal K-VAE decoder,
the temporal kernels are shifted to align them with this
symmetric window.

\paragraph{$\times 2$ model.}
For the same output tile, the $\times 2$ input contains four
times as many spatial latent cells as the $\times 4$ input.
Applying all five 2048-wide ResBlocks on this larger grid
would substantially increase computation.
The $\times 2$ branch therefore uses a shorter front section
and only one spatial upsampling stage.

The $\times 2$ branch is trained after the $\times 4$ model
and initialized from its trained checkpoint.
All branch modules are copied from the $\times 4$ model,
with no randomly initialized parameters:
\begin{itemize}
    \item the input stem is copied from the $\times 4$ stem;
    \item two 2048-wide adapter ResBlocks are copied from
    the first and fifth ResBlocks of the $\times 4$ model;
    \item the convolutions of the first upsampling module
    are copied and applied without nearest-neighbor resizing;
    \item the remaining modules are separate copies of the three $2048 \to 1024$ ResBlocks, the second upsampling module, the three $1024 \to 512$ ResBlocks, and the output head.
\end{itemize}
All modulated normalization layers in this branch are
conditioned on the $\times 2$ input latent.
The original $\times 4$ path remains frozen throughout
$\times 2$ training, while all copied $\times 2$ modules
are optimized from the first training step.
The $\times 2$ inference path contains approximately
0.763B parameters and does not execute the original
$\times 4$ path. %
\newpage

\section[Training Stages]{Training Stages}\label{sec:training}

\begin{figure}[h!]
    \centering
    \includegraphics[width=\textwidth]{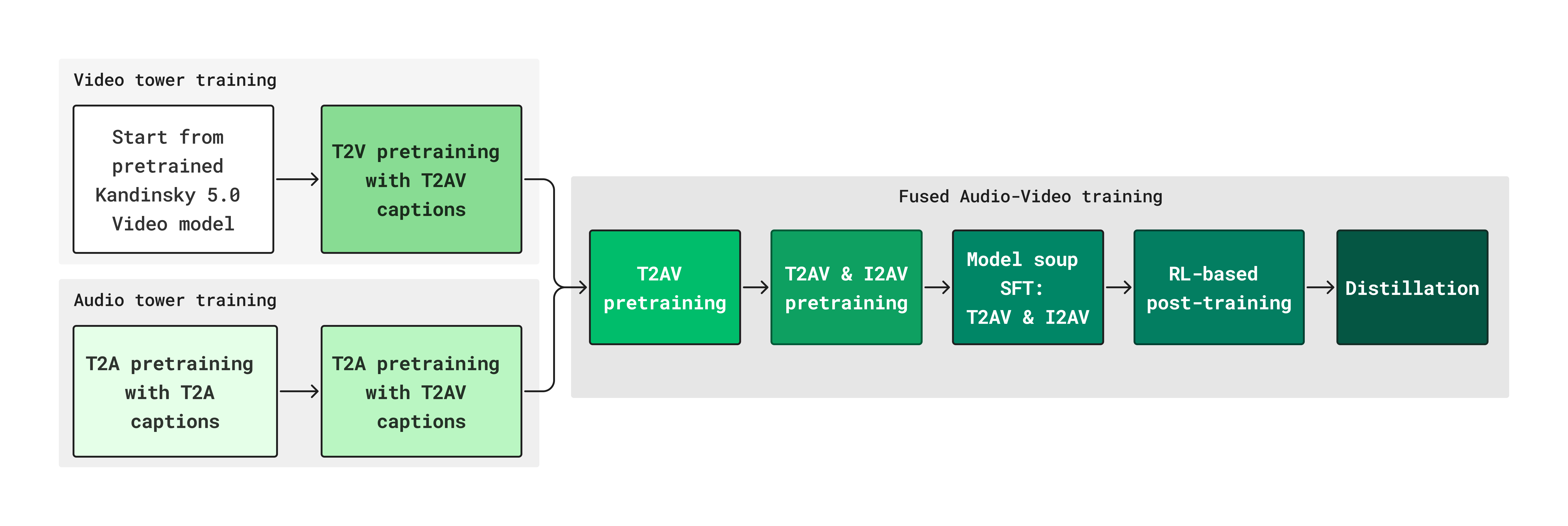}
    \caption{Training pipeline of Kandinsky~6.0 Video. The video stream (initialized from the pretrained Kandinsky~5.0 Video model) and the audio stream are first pre-trained separately, then fused and trained jointly in T2AV mode and in mixed T2AV/I2AV mode. Joint pre-training is followed by supervised fine-tuning with model soup, RL-based post-training, and distillation.}
    \label{fig:training_pipeline}
\end{figure}

\subsection{Overview}

Kandinsky~6.0 Video is trained in several sequential stages (Figure~\ref{fig:training_pipeline}). Pre-training comprises three phases: separate training of the video stream, separate training of the audio stream from scratch, and joint training of both streams. Pre-training is followed by supervised fine-tuning (SFT), RL-based post-training (alignment), and distillation. Kandinsky~6.0 Video Lite and Kandinsky~6.0 Video Pro follow the same training scheme and differ mainly in the number of steps, batch size, and learning rates; other differences are stated where they apply (e.g., the RL setup in Section~\ref{sec:rl}).

\subsection{Continuous Pre-training}

All pre-training stages use flow matching~\cite{lipman2023flow} with an MSE loss. The audio and video streams are first trained separately and are then connected through cross-attention layers and trained jointly. The audio stream is trained from scratch, with weights randomly initialized from a normal distribution. The video stream is initialized from the corresponding Kandinsky~5.0 Video (Lite or Pro) pre-training checkpoint (SD resolution, 45k steps). At each stage, the Lite and Pro models are trained on the same data.

\subsubsection{Separate Training}

\textbf{Kandinsky~6.0 Video Lite:}
\begin{itemize}
    \item \textbf{Text-to-audio (T2A):} 40k steps on T2A captions; batch size $128 \times 640$~seconds; sample length from 5 to 20~seconds.
    \item \textbf{T2A (continued):} 2k steps on T2AV captions; batch size $256 \times 640$~seconds; sample length from 5 to 20~seconds.
    \item \textbf{Text-to-video (T2V), in parallel with T2A:} starting from the Kandinsky~5.0 Video Lite SD pre-training checkpoint (45k steps on T2V captions), 20k steps of T2V pre-training on T2AV captions; batch size $2 \times 512 = 1024$.
\end{itemize}

\textbf{Kandinsky~6.0 Video Pro:}
\begin{itemize}
    \item \textbf{T2A:} 30k steps on T2A captions; batch size $128 \times 640$~seconds.
    \item \textbf{T2A (continued):} 4k steps on T2AV captions; batch size $256 \times 640$~seconds.
    \item \textbf{T2V, in parallel with T2A:} starting from the Kandinsky~5.0 Video Pro SD pre-training checkpoint (45k steps on T2V captions), 14k steps of T2V pre-training on T2AV captions; batch size $1 \times 1024 = 1024$.
\end{itemize}

\subsubsection{Joint T2AV Training}

\textbf{Kandinsky~6.0 Video Lite:}
\begin{itemize}
    \item \textbf{T2AV:} 57.5k steps; batch size 256; up to 121 frames at 23--25~fps (at most 5~seconds of video).
    \item \textbf{T2AV + I2AV:} 4k additional steps with learning rate $1\mathrm{e}{-5}$, with 25\% of samples in I2AV mode.
\end{itemize}

\textbf{Kandinsky~6.0 Video Pro:}
\begin{itemize}
    \item \textbf{T2AV:} 46k steps with learning rate $5\mathrm{e}{-5}$, followed by 3k steps with learning rate $2.5\mathrm{e}{-5}$ (49k steps in total).
    \item \textbf{T2AV + I2AV:} 5k additional steps with learning rate $2.5\mathrm{e}{-5}$, with 25\% of samples in I2AV mode.
\end{itemize}

During joint T2AV training, we use the \texttt{independent\_time} mode, in which the diffusion time is sampled independently for video and audio, so that the two modalities are noised to different levels $t$. This is required for audio-to-video and video-to-audio generation, where the model must take into account information from the other modality. In addition, a training batch may contain audio only, video only, images only, or video with audio; images come from the image dataset, which is also used to inject knowledge of the Russian Cultural Code (RCC). Training is performed in bfloat16 precision. Training details and hyperparameters for all continuous pre-training stages are given in Table~\ref{tab:training}.

\begin{table}[htbp]
\centering
\caption{Training hyperparameters and data sampling settings for the continuous pre-training stages of Kandinsky~6.0 Video Lite and Pro (values given as Lite\,/\,Pro where they differ): separate T2V and T2A pre-training (the latter first on audio captions and then on audio-video captions), joint T2AV training, and the final T2AV stage with 25\% of I2AV samples.}
\label{tab:training}
\resizebox{\textwidth}{!}{%
\begin{tabular}{l | c c c c c}
\toprule
Parameter & T2V & T2A, & T2A, & T2AV & T2AV + \\
& (Lite/Pro) & Audio captions & AV captions & (Lite/Pro) & I2AV (25$\%$) \\
& & (Lite/Pro) & (Lite/Pro) & & (Lite/Pro) \\
\midrule
num\_steps & 45k (T2V captions) & 40k / 30k & 2k / 4k & 57.5k / 49k & 4k / 5k\\
& 20k/14k (T2AV captions) & & & & \\
\midrule
& \multicolumn{5}{c}{\textbf{Optimizer}} \\
\midrule
learning\_rate & 3e-05 & 0.0001 & 3e-05 & 5e-05 / 5e-05 (46k steps) & 1e-05 / 2.5e-05 \\
& & & & \& 2.5e-05 (3k steps) & \\
weight\_decay & 1e-3 & 0.001 & 0.001 & 0.001 & 0.001 \\
betas & (0.9, 0.95) & (0.9, 0.999) & (0.9, 0.999) & (0.9, 0.95) & (0.9, 0.95) \\
ema\_decay & 0.999 & 0.999 & 0.999 & 0.999 & 0.999 \\
num\_warmup\_steps & 1000 & 1000 & 1000 & 1000 & 1 / 0 \\
\midrule
& \multicolumn{5}{c}{\textbf{Data}} \\
\midrule
uncond\_prob & 0.1 & 0 & 0.1 & 0.1 & 0.1 \\
\midrule
& \multicolumn{5}{c}{Image} \\
\midrule
data\_prob & 0.02 / 0.01 & -- & -- & 0 / 0.01 & -- \\
max\_seq\_len & 62 / 31 & -- & -- & 62 / 31 & -- \\
resolutions\_probs & 768$\times$512: 0.502; & -- & -- & 768$\times$512: 0.502; & -- \\
& 512$\times$768: 0.272; & -- & -- & 512$\times$768: 0.272; & -- \\
& 512$\times$512: 0.226 & -- & -- & 512$\times$512: 0.226 & -- \\

long\_prompts\_prob & 0.7 & -- & -- & as for video & -- \\
medium\_prompts\_prob & 0.15 & -- & -- & as for video & -- \\
short\_prompts\_prob & 0.1 & -- & -- & as for video & -- \\
shortest\_prompts\_prob & 0.05 & -- & -- & as for video & -- \\
RCC\_en\_caption\_prob & 0.5 & -- & -- & 0.5 & -- \\
RCC\_ru\_caption\_prob & 0.5 & -- & -- & 0.5 & -- \\
\midrule
& \multicolumn{5}{c}{Video} \\
\midrule
video\_len (latent frames) & [5, 31] & -- & -- & [1, 31] & [5, 31] \\
data\_prob & 0.98 / 0.99 & -- & -- & 1.0 / 0.99 & 1.0 \\
fps\_range & [23, 25] & -- & -- & [23, 25] & [23, 25] \\
resolutions\_probs & 768$\times$512: 0.989; & -- & -- & 768$\times$512: 0.973; & 768$\times$512: 0.999; \\
& 512$\times$768: 0.006; & -- & -- & 512$\times$768: 0.024; & 512$\times$768: 0.001; \\
& 512$\times$512: 0.005 & -- & -- & 512$\times$512: 0.003 & 512$\times$512: 0 \\

long\_prompts\_prob & 0.7 & -- & -- & 0.7 & 0.7 \\
medium\_prompts\_prob & 0.15 & -- & -- & 0.15 & 0.15 \\
short\_prompts\_prob & 0.1 & -- & -- & 0.1 & 0.1 \\
shortest\_prompts\_prob & 0.05 & -- & -- & 0.05 & 0.05 \\
RCC\_en\_caption\_prob & 0.5 & -- & -- & 0.5 & 0.5 \\
RCC\_ru\_caption\_prob & 0.5 & -- & -- & 0.5 & 0.5 \\
\midrule
& \multicolumn{5}{c}{Audio} \\
\midrule
audio\_len (sec) & -- & [5, 20] & [2, 31] & as for video & as for video \\
data\_prob & -- & 1.0 & 1.0 & as for video & as for video \\
long\_prompts\_prob & -- & 0.7 & 0.7 & as for video & as for video \\
medium\_prompts\_prob & -- & 0.15 & 0.15 & as for video & as for video \\
short\_prompts\_prob & -- & 0.1 & 0.1 & as for video & as for video \\
shortest\_prompts\_prob & -- & 0.05 & 0.05 & as for video & as for video \\
\bottomrule
\end{tabular}%
}
\end{table}

\subsubsection{Image-to-Audio-Video (I2AV) Mode}

\begin{figure}[htbp]
  \centering
  \begin{minipage}{\textwidth}
    \centering
    \includegraphics[width=\textwidth]{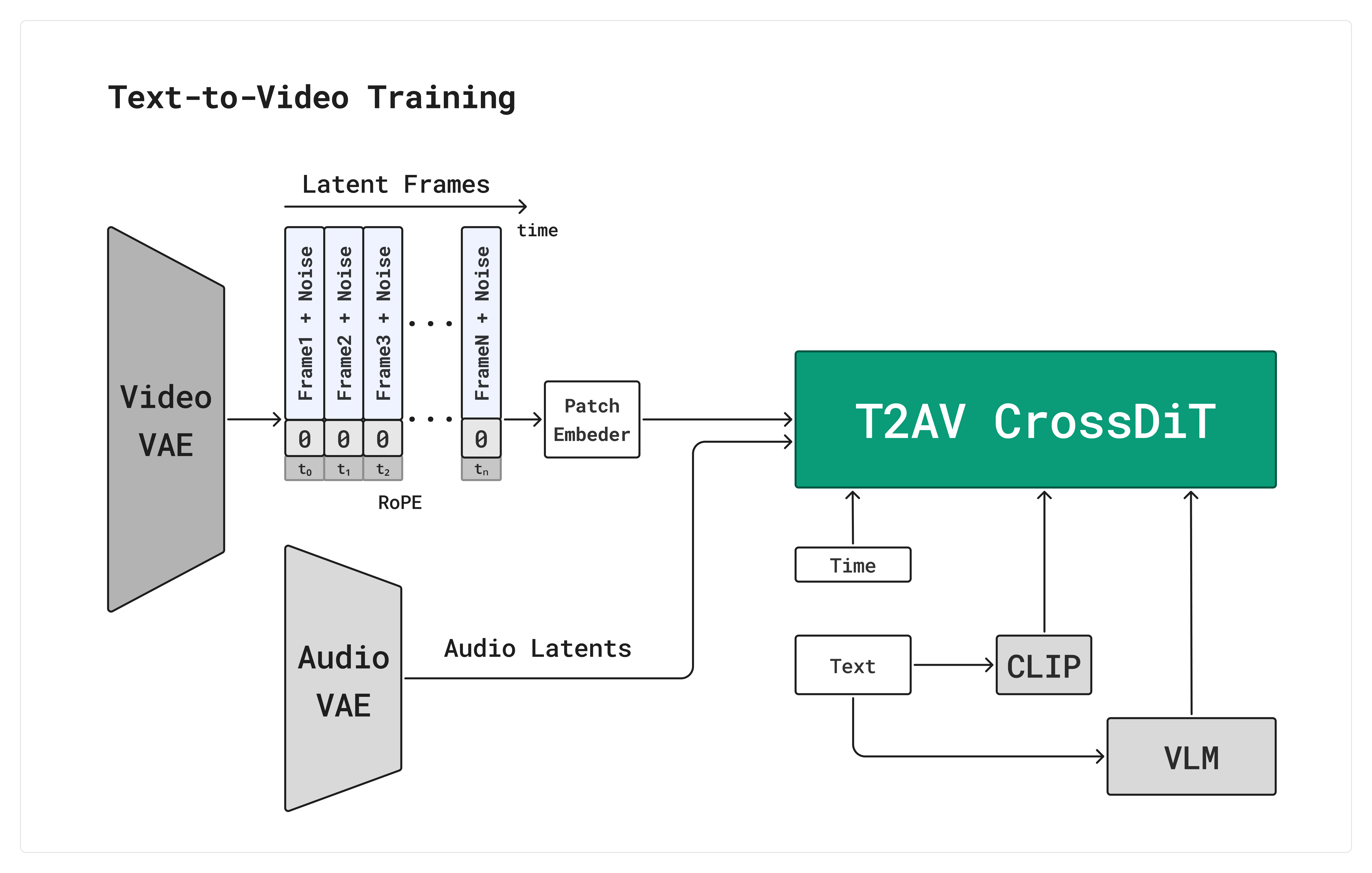}
  \end{minipage}
  \begin{minipage}{\textwidth}
    \centering
    \includegraphics[width=\textwidth]{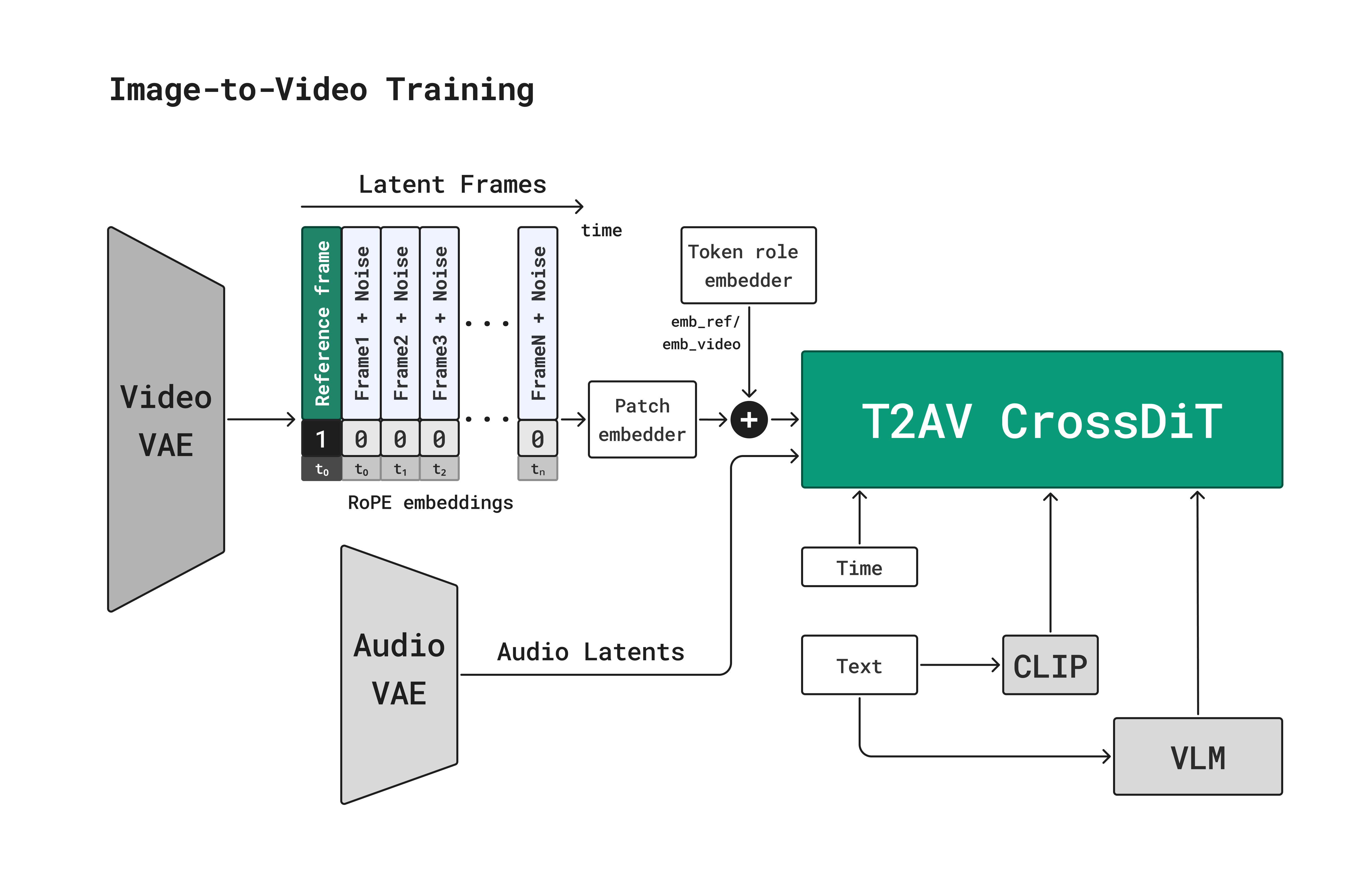}
  \end{minipage}
  \caption{Input construction in T2AV (top) and I2AV (bottom) training modes. In I2AV mode, the latent of the reference frame is concatenated with the noisy video latents along the temporal axis and marked with mask value~1 (0 for the frames to be generated); a learnable Token Role Embedding (\texttt{emb\_ref} / \texttt{emb\_video}) is added to the patch-embedded tokens. In both modes, the text prompt is encoded by CLIP and a VLM, and audio latents are produced by the audio VAE.}
  \label{fig:t2av_i2av_training}
\end{figure}

Both model versions are trained in I2AV mode with a probability of 25\% during the final pre-training stage. The input scheme is compared with that of T2AV mode in Figure~\ref{fig:t2av_i2av_training}. The latent of the conditioning frame is concatenated with the video latents along the temporal axis and receives its own temporal RoPE position. A binary channel mask distinguishes the conditioning frame (value~1) from the frames to be denoised (value~0). In addition, we introduce two learnable embeddings in the token space, the Token Role Embeddings (TRE): one for the noisy video tokens and one for the reference frame. They are added to the input tokens and provide an additional learnable signal indicating whether a token is noise or conditioning.

Inference-time ablations showed that TRE has a relatively weak effect on the result. Disabling the embeddings, or decomposing them into shared and role-specific components, changed the generations only slightly; a noticeable effect appeared only when the role-specific component was artificially amplified many times. This indicates that TRE carries discriminative information about token roles but is not the primary mechanism for recognizing the reference frame, since other cues largely duplicate its function. At the same time, inference-time ablations cannot fully determine how useful TRE is during training; this would require a separate training run without these embeddings.

The channel mask and the temporal RoPE had the strongest effects. Setting the mask to one for all tokens almost completely blocked denoising, whereas an all-zero mask still allowed the video to be denoised but produced artifacts, mainly in the first latent frame. Changing the temporal RoPE position of the reference frame made the model use it at the specified position: a reference with RoPE position $t_4$, $t_8$, and so on acted as a condition for the corresponding part of the video. Thus, RoPE is the primary mechanism that binds the reference frame to a position in the generated sequence, the mask explicitly separates conditioning tokens from tokens to be denoised, and TRE serves as an additional, partially redundant signal.

\subsection{Supervised Fine-Tuning}\label{sec:sft}

Supervised fine-tuning is performed in two stages on the SFT datasets described in Section~\ref{sec:data_processing_pipeline} (see Table~\ref{tab:sft_domain_counts}). In both stages, a separate model is fine-tuned on each of the 11 domains. The domain-specific models are trained independently through both stages and are merged into a model soup by uniform weight averaging only at the end of Stage~2. The same procedure is applied to both Kandinsky~6.0 Video Lite and Kandinsky~6.0 Video Pro.

\textbf{Stage 1 (visual quality).} The goal of this stage is to improve visual quality. Training starts from the pre-trained checkpoint. The audio stream, the fusion modules, and the audio-text modules are frozen, and only the video loss is optimized: audio inputs still pass through the frozen audio stream but do not contribute to the loss. We use a learning rate of \(1\times10^{-5}\) and the same mixture of prompt detail levels as in pre-training (long: 0.7, medium: 0.15, short: 0.1, shortest: 0.05). The model is trained in T2AV and I2AV modes (75\% and 25\% of steps, respectively). Training is stopped after 7,000 steps, the visually best iteration before overfitting begins. To evaluate the result of this stage, the 11 domain-specific checkpoints are uniformly averaged into an intermediate Stage~1 soup. This soup is used only for quality measurement and is not used in further training.

\textbf{Stage 2 (synchronization and lip-sync).} This stage improves audio-video synchronization, with particular emphasis on lip-sync. Training continues separately from each domain-specific Stage~1 checkpoint rather than from the Stage~1 soup. All modules are unfrozen and trained with different learning rates: \(1\times10^{-5}\) for the video stream and the shared modules and \(1\times10^{-6}\) for the audio components. The loss is a weighted sum of the video loss (weight 0.85), the audio loss (0.15), and a new face loss (1.0). The face loss is the video loss restricted to the facial region in the latent space; it encourages more accurate lip-sync. As in Stage~1, 75\% of steps are in T2AV mode and 25\% in I2AV mode. After 5,000 steps, the 11 domain-specific checkpoints are uniformly averaged to obtain the final SFT model.

\subsection{Reinforcement Learning}
\label{sec:rl}

\begin{figure}[h!]
    \centering
    \includegraphics[width=\linewidth]{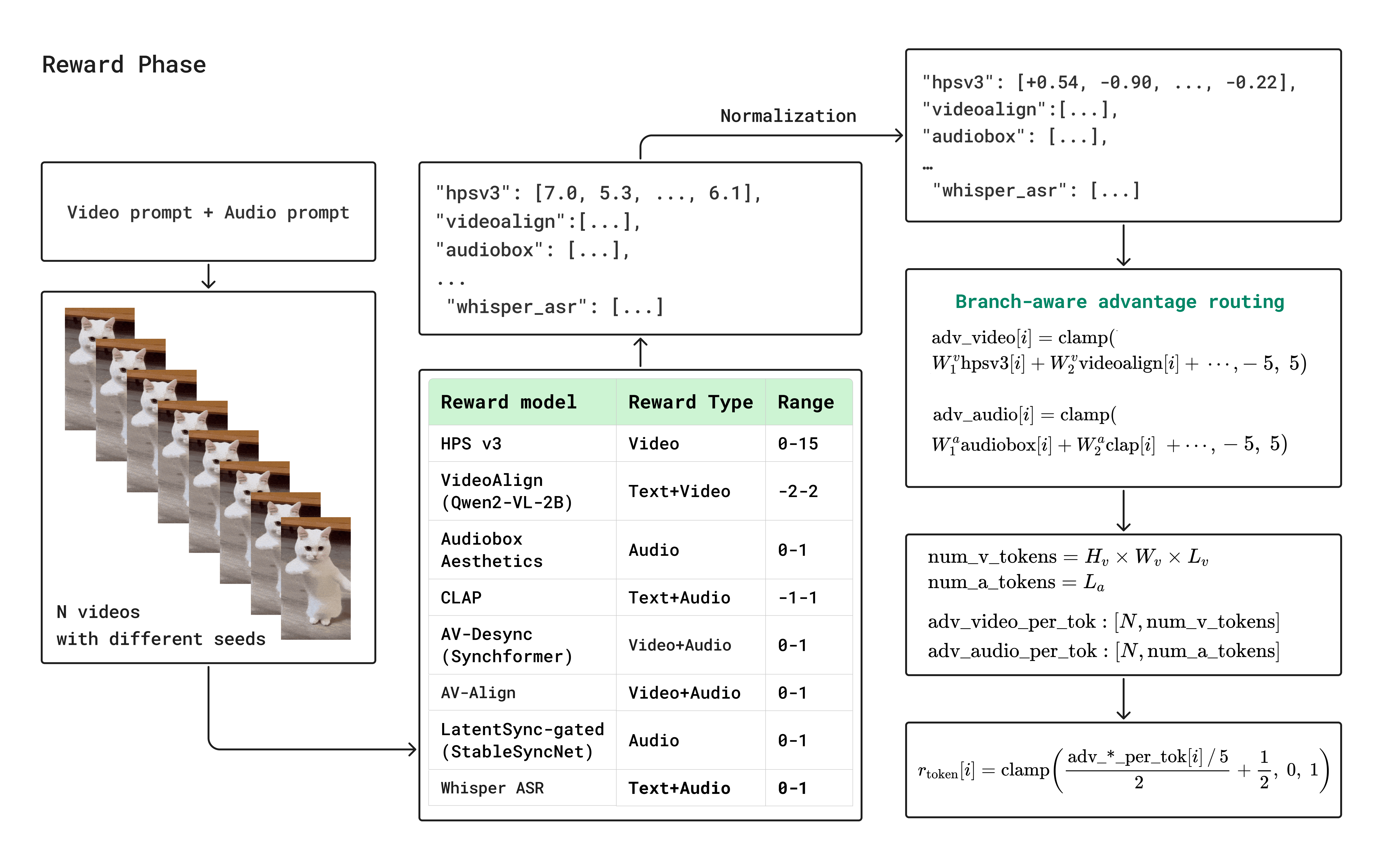}
    \caption{Reward phase of RL post-training. For each prompt, a group of videos with audio is generated with different random seeds and scored by eight reward terms. Each score is centered within the group and scaled by the batch-wide standard deviation, routed into separate video and audio advantages (sync terms enter both), clamped to $[-5,5]$, broadcast to the video and audio tokens, and rescaled to per-token rewards $r_\text{token} \in [0, 1]$, where 1 corresponds to a rollout far above the group mean and 0 far below it.}
    \label{fig:rl_reward}
\end{figure}

\begin{figure}[h!]
    \centering
    \includegraphics[width=\linewidth]{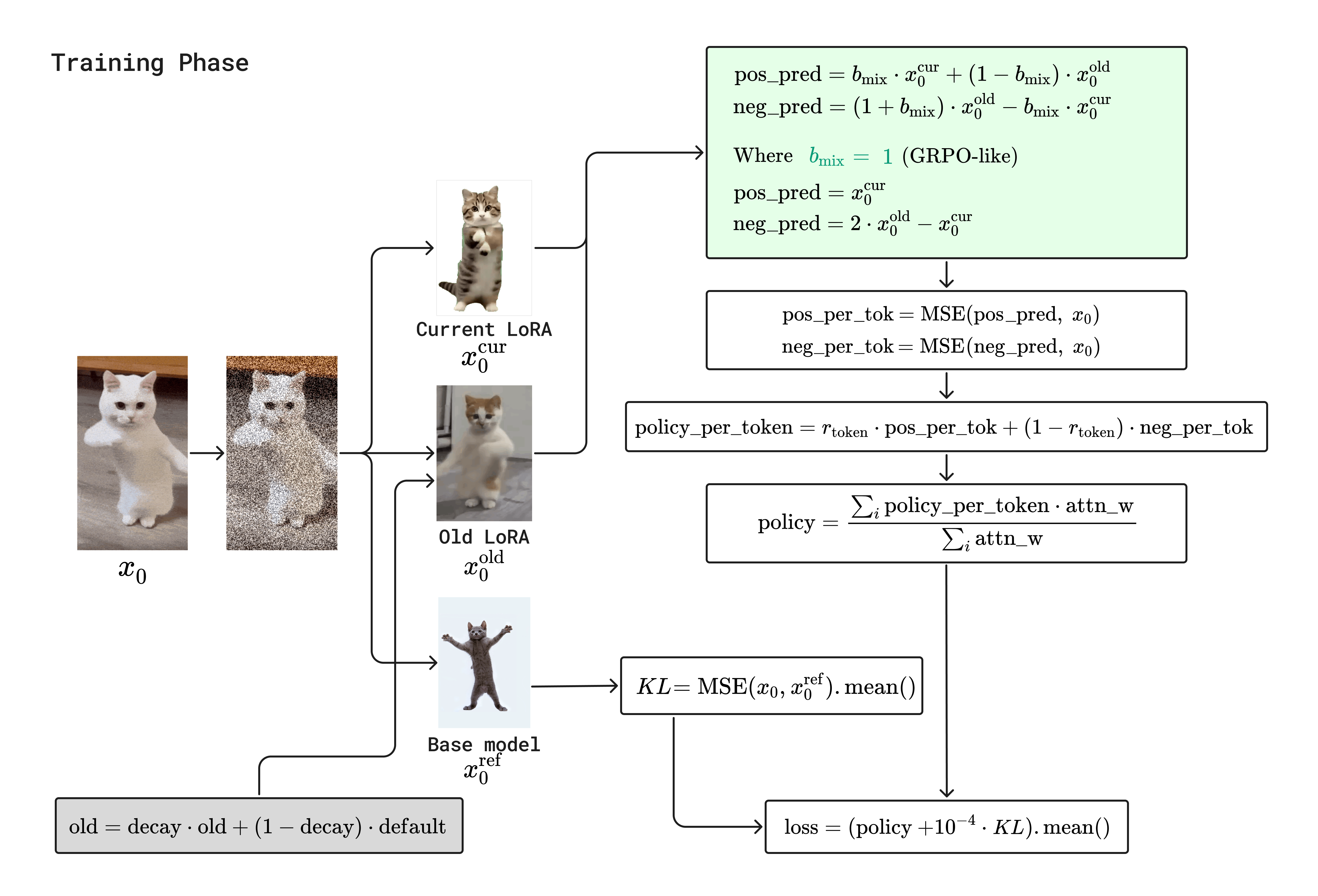}
    \caption{Training phase of RL post-training. The rollout latent is re-noised and denoised by the current LoRA, the old LoRA, and the frozen base (SFT) model. Predictions of the current and old LoRA are combined into positive and negative predictions; their per-token MSE losses to the rollout latent are interpolated with the rewards $r_\text{token}$ from the reward phase and aggregated with the region-wise attention weights $\text{attn\_w}$ derived from A2V cross-attention. A KL term with respect to the base-model prediction regularizes training; the old LoRA is an exponential moving average of the current one.}
    \label{fig:rl_training}
\end{figure}

For RL post-training, we adapt the OmniNFT method~\cite{zhang2026omninft}, originally developed for LTX-2 19B, to the Kandinsky~6.0 Video backbone. The approach combines modality-wise advantage routing, layer-wise gradient surgery, and region-wise loss reweighting. The reward phase and the training phase are illustrated in Figures~\ref{fig:rl_reward} and~\ref{fig:rl_training}, respectively. A side-by-side comparison of the RL and SFT checkpoints is given in Section~\ref{sec:ablation-rl-vs-sft}.

\paragraph{Data sampling and conditioning.}
The RL dataset is used online. All RL training is performed in pure T2AV mode; image-conditioned generation is not used. In each training epoch, prompts are sampled as follows:
\begin{itemize}
    \item \textbf{Bilingual prompting:} each prompt is given to the model in Russian or English with equal probability. The language is chosen deterministically from a hash of the sample index and the epoch, which keeps the choice consistent across distributed ranks. Reward models always receive English metadata to ensure stable scoring.
    \item \textbf{Speech oversampling:} clips containing speech are oversampled by a factor of 1.5 to provide sufficient signal for the speech-related rewards (Whisper-ASR, lip-sync).
    \item \textbf{Prompt format:} visual and audio descriptions are combined into the native Kandinsky format:
    \begin{verbatim}
    <visual description> <AUDCAP><audio description><ENDAUDCAP>
    \end{verbatim}
    For scenes with literal speech, the spoken text is enclosed in the markers \texttt{<S>...<E>}.
\end{itemize}

\paragraph{Reward functions and routing.}
The reward stack consists of eight objectives grouped into three branches: video, audio, and sync. Branch weights and per-reward weights are given in parentheses:
\begin{itemize}
    \item \textit{Video branch (2.5):} HPSv3 (1.5), VideoAlign (1.0);
    \item \textit{Audio branch (2.0):} AudioBox Aesthetics (0.5), CLAP (1.0), Whisper-ASR (0.5);
    \item \textit{Sync branch (2.75, affects both video and audio):} AV-desync (0.5), AV-align (0.75), LatentSync-gated (1.5). The last term is the StableSyncNet lip-sync score modulated by speech intelligibility, $s_\text{LS}\cdot(0.5 + 0.5\,s_\text{ASR})$, where $s_\text{ASR} = \max(0, 1 - \mathrm{WER})$ is the Whisper-ASR score of the same rollout.
\end{itemize}
Each raw reward is first converted into an advantage: it is centered by its mean within the rollout group and scaled by the batch-wide standard deviation, floored at $0.1$ to avoid advantage blow-up in near-uniform groups. The weighted advantages are then summed into two branch-aware advantages, one per modality. The sync-branch terms (AV-desync, AV-align, LatentSync-gated) are added to both the video and the audio advantages. Both advantages are clamped to $[-5, 5]$ for stability. The per-modality advantages are then broadcast to per-token rewards: the video advantage over the latent video token grid $H_v \times W_v \times L_v$ and the audio advantage over the $L_a$ audio tokens. Each is rescaled to $r_\text{token} = \mathrm{clamp}((\mathrm{adv}/5 + 1)/2,\, 0,\, 1)$ before being used in the policy update.

\paragraph{RL setup.}
We use GRPO groups of $K=10$ rollouts per prompt. Rollouts are generated at $512 \times 768$ resolution with $31$ latent frames (121 frames, 5~s at 24~fps), using the DPM2 solver with $20$ steps, $\mathrm{CFG}=1.0$, and noise level $0$. With these settings, the sampler is deterministic, so the random seed is the only source of diversity within a group. Note that RL is performed at a lower spatial resolution than Full-HD inference. The model is fine-tuned with LoRA ($r=16$, $\alpha=32$) on top of the frozen SFT model. LoRA is applied to every linear layer inside the transformer blocks of both streams --- self-attention, text cross-attention, feed-forward, and both cross-modal attention modules (Q, K, V, and output projections); text, time, and modulation embeddings are not adapted. This amounts to 0.69\% of the parameters for Pro (199M) and 1.57\% for Lite (47M). The LoRA is trained from scratch, without a warm-start checkpoint or transferred optimizer state.

Both model sizes undergo RL post-training with the same recipe. The hyperparameters in this section refer to Kandinsky~6.0 Video Pro (60 blocks); for Lite (32 blocks) we use $K=8$ rollouts per prompt and a gradient-surgery schedule recalibrated for its depth (see below). Unless stated otherwise, the numbers reported here are for Pro.

\paragraph{Optimization.}
The training phase is shown in Figure~\ref{fig:rl_training}. The clean rollout sample $x_0$ is re-noised at timesteps drawn from the 8 noisiest of the 20 sampling timesteps (8 random draws per rollout) and passed to three models: the current LoRA, the old LoRA, and the frozen base (SFT) model, which produce the predictions $x_0^\text{cur}$, $x_0^\text{old}$, and $x_0^\text{ref}$, respectively. The policy loss compares the predictions of the current and the old LoRA in a DPO-like manner. The positive and negative predictions are
\begin{gather*}
\text{pos\_pred} = b_\text{mix}\cdot x_0^\text{cur} + (1 - b_\text{mix})\cdot x_0^\text{old},\\
\text{neg\_pred} = (1 + b_\text{mix})\cdot x_0^\text{old} - b_\text{mix}\cdot x_0^\text{cur}.
\end{gather*}
We use the GRPO-like setting $b_\text{mix}=1$, for which $\text{pos\_pred} = x_0^\text{cur}$ and $\text{neg\_pred} = 2\cdot x_0^\text{old} - x_0^\text{cur}$. Both predictions are compared with the clean sample $x_0$,
\[
\text{pos\_per\_tok} = \mathrm{MSE}(\text{pos\_pred}, x_0), \qquad
\text{neg\_per\_tok} = \mathrm{MSE}(\text{neg\_pred}, x_0),
\]
each divided by the per-sample mean absolute deviation of the corresponding prediction from $x_0$ (as in OmniNFT), and weighted with the per-token rewards $r_\text{token}$:
\[
\text{policy\_per\_token} = r_\text{token}\cdot\text{pos\_per\_tok}
+ (1 - r_\text{token})\cdot\text{neg\_per\_tok}.
\]
The per-token terms are aggregated with attention weights,
\[
\text{policy} = \frac{\sum_i \text{policy\_per\_token}_i \cdot \text{attn\_w}_i}{\sum_i \text{attn\_w}_i},
\]
where $\text{attn\_w}$ is the region-wise reweighting mask described below. The old LoRA is an exponential moving average of the current one, updated once per epoch as
$\theta_\text{old} \leftarrow d\cdot\theta_\text{old} + (1 - d)\cdot\theta_\text{cur}$,
where the decay $d$ grows linearly from $0$ by $10^{-3}$ per optimizer step and is capped at $0.5$.
A KL term anchors the model to the frozen SFT model: $\mathrm{KL} = \mathrm{MSE}(x_0^\text{cur}, x_0^\text{ref})$, and the final loss is
\[
\text{loss} = \mathrm{mean}\left(\text{policy} + \beta\cdot\mathrm{KL}\right), \qquad \beta = 10^{-4}.
\]
We use AdamW, initialized from scratch, with a learning rate of $3\times10^{-5}$, linear warmup over 5 epochs, and cosine decay to $5\times10^{-6}$ over 300 epochs. Gradient norms are clipped at $1.0$.

\paragraph{Checkpoint selection.}
We do not take the last training epoch as the final RL checkpoint. Instead, we track the weighted sum of rewards during training and select the checkpoint with the best average over a sliding window of 5 epochs, which avoids picking a single lucky iteration.

\paragraph{Regularization techniques.}
\begin{itemize}
    \item \textit{Layer-wise gradient surgery.} To prevent video-branch gradients from disrupting audio representations, we scale down the gradients on the key-value path of the V2A cross-attention, in which video queries attend to audio keys and values (Figure~\ref{fig:CrossDiTFusedBlock}). The production Pro model uses a data-driven per-block schedule: scale 0.0 for block 0, 0.1 for blocks 1--10, and 0.3 for blocks 50--59 (a deep AV-sync zone identified by KV ablation), while blocks 11--49 are left unscaled (scale 1.0). For Lite (32 blocks) the recalibrated schedule is 0.0 for block 0, 0.1 for blocks 1--5, and 0.3 for blocks 22--26. The zones reported for LTX-2 did not transfer directly to our models, so we calibrated these schedules with our own probes: blockwise ablation of the cross-modal key--value input (measuring the resulting audio--video desynchronization) and measurement of the video-loss gradient leaking into the audio stream.
    \item \textit{Region-wise loss reweighting.} The weights $\text{attn\_w}$ used in the policy aggregation are derived from the attention maps of the A2V cross-attention, in which audio queries attend to video keys and values, sampled at a subset of denoising steps. Video patches that receive stronger attention from audio tokens (typically faces, hands, and musical instruments) are assigned higher weights. The reweighting factor is capped at 1.5$\times$ and is mapped from the DiT patch grid to the pre-patch loss grid by spatial unpatchifying.
\end{itemize}

\paragraph{Results.}
RL post-training substantially improves speech intelligibility. For Kandinsky~6.0 Video Pro, the word error rate (WER) on a held-out speech benchmark (64 prompts, 3 seeds each) decreases from 0.235 for the SFT model to 0.124, a relative reduction of 47\% (paired Wilcoxon test, $p < 10^{-3}$); for Lite it decreases from 0.179 to 0.127 ($-29\%$, $p < 10^{-3}$). WER is computed with Whisper-large-v3, which is also one of the reward models; the gain is confirmed on the public Harvard sentences (100 prompts), where Pro WER drops from 0.251 to 0.180 ($p = 0.005$).

\subsection{Distillation}

Following Kandinsky~5.0~\citep{kandinsky5}, we split distillation into two complementary stages:

\begin{enumerate}[label=\textbf{\arabic*.}]
    \item \textbf{Trajectory distillation:} the teacher's sampling trajectory is compressed into a few-step student.
    \item \textbf{Quality improvement:} the perceptual fidelity lost during aggressive trajectory compression is recovered using real data and adversarial training.
\end{enumerate}

\begin{figure}[htbp]
    \centering
    \begin{subfigure}[b]{\textwidth}
        \centering
        \includegraphics[width=\linewidth]{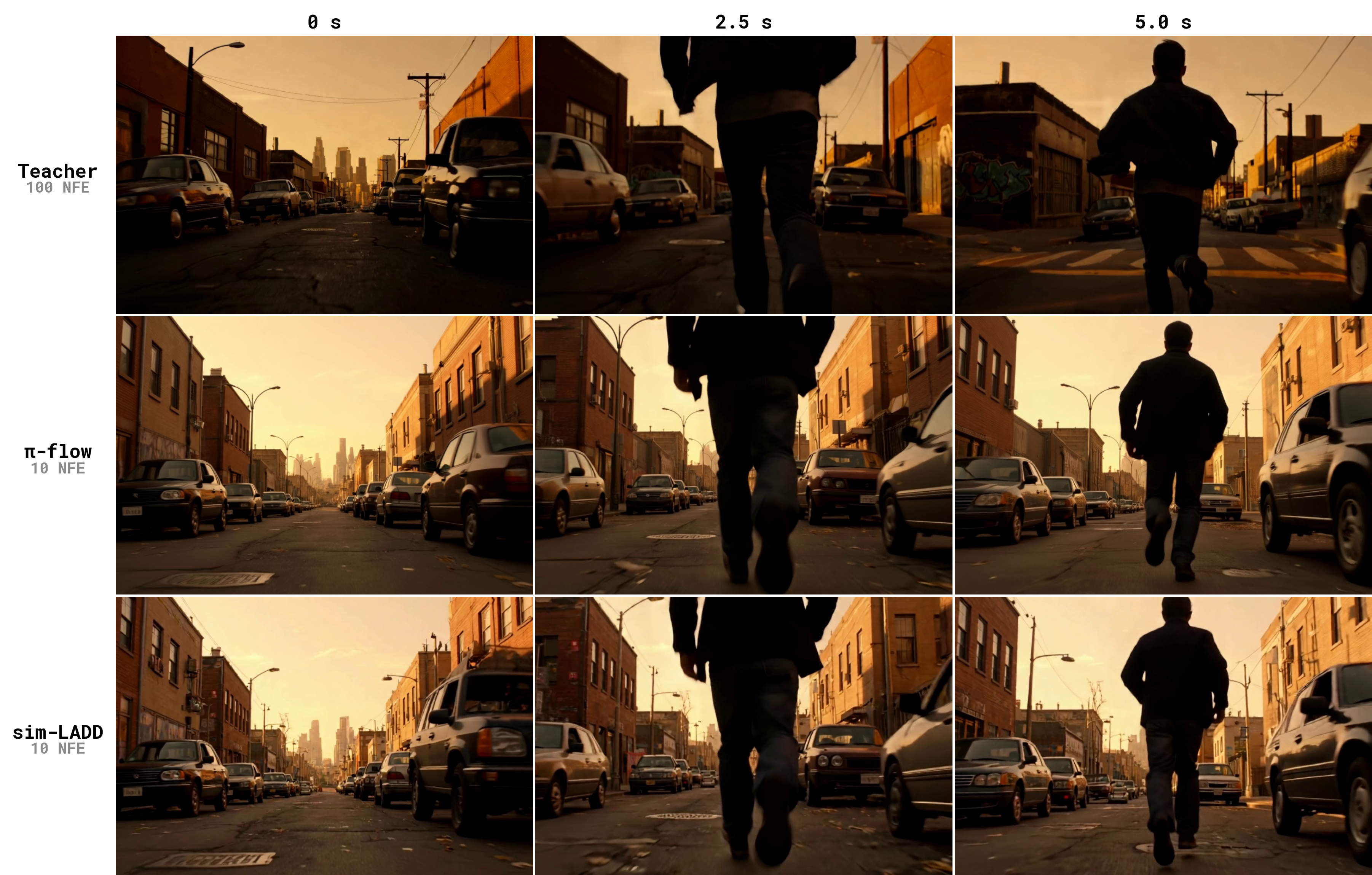}
    \end{subfigure}
    \\[1ex]
    \begin{subfigure}[b]{\textwidth}
        \centering
        \includegraphics[width=\linewidth]{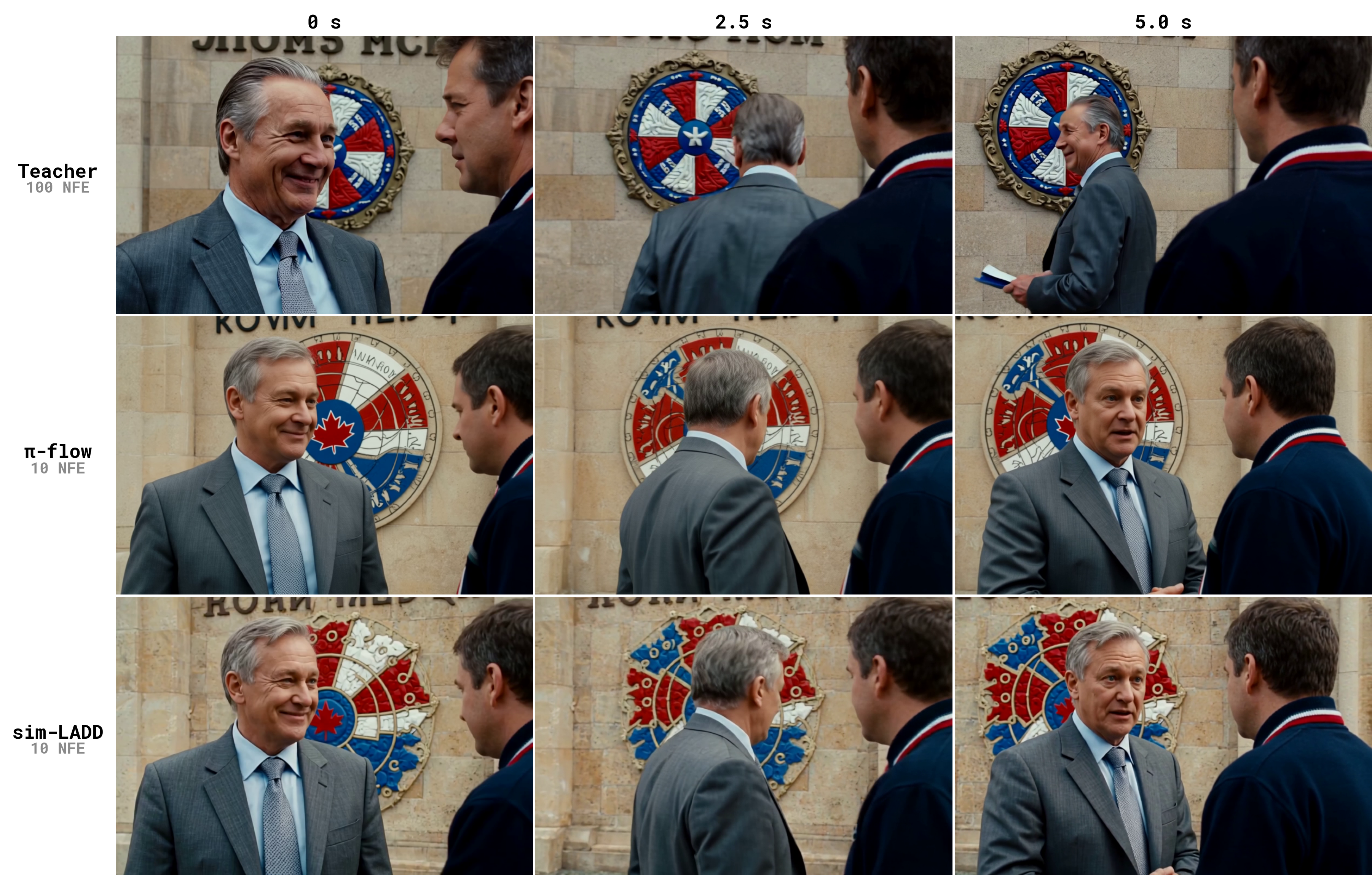}
    \end{subfigure}
    \caption{Qualitative video comparison. Storyboard frames at 0, 2.5, and 5.0 seconds comparing the 100-NFE diffusion teacher with the 10-NFE $\pi$-flow baseline and final stage sim-LADD. $\pi$-flow closely matches the teacher in terms of semantic content and scene composition, while sim-LADD further improves visual fidelity, producing noticeably sharper details and richer textures throughout the video.}
    \label{fig:distill_storyboards}
\end{figure}

See Fig.~\ref{fig:distill_storyboards} for a visual comparison of the teacher, $\pi$-flow, and sim-LADD generations.

\subsubsection{Motivation and Method Selection}

For trajectory distillation, we evaluated several families of methods: DMD~\citep{dmd}, DMD2~\citep{dmd2}, SID~\citep{sid}, Phased Consistency~\citep{pcm}, CFG distillation~\citep{cfg_distillation}, $\pi$-Flow~\citep{piflow}, and Parallel Decoding Distillation (PDD)~\citep{pdd}.

Distribution-matching methods (DMD, DMD2, SID) provide a direct objective for aligning the student and teacher distributions, but in our experiments they were prone to mode collapse and mode dropping. They also required substantially more GPUs because of expensive teacher evaluations, auxiliary score estimation, and optimization overhead. This cost is especially high for video, whose spatio-temporal latents are far more expensive to process than those of images.

CFG distillation and consistency-based methods are simple and effective baselines, but we observed that they can alter the teacher's generation behavior, including motion dynamics and the distribution of generated samples. They remained useful but were not flexible or performant enough for our few-step video generation setting.

We then evaluated two more recent trajectory-based methods, $\pi$-Flow and PDD. Of these, $\pi$-Flow showed the most reliable convergence and the best preservation of the teacher trajectory. Its imitation-learning formulation trains the student on-policy, i.e., on the states encountered during the student's own rollout, which is the same distribution the student faces at inference.

As with DMD, the quality of $\pi$-Flow is ultimately bounded by the teacher. Unlike distribution-matching methods, however, it avoids an aggressively mode-seeking objective: it encourages the student to reproduce the teacher's local decisions along the trajectory, which substantially reduces mode collapse and mode dropping. The guided teacher trajectory can also be distilled directly, which removes the need for a separate CFG distillation stage. For these reasons, we chose $\pi$-Flow for trajectory distillation.

Trajectory compression alone does not fully preserve model quality: as the number of sampling steps decreases, generated videos become smoother and less detailed. This effect grows with more aggressive compression even when the semantic composition and global motion remain correct: the student loses high-frequency texture, local contrast, and fine spatial detail.

Additional trajectory matching does not reliably fix this degradation. Therefore, the second stage uses real data and an adversarial objective to restore perceptual fidelity. We build on LADD~\citep{ladd} and introduce Simulated LADD (Sim-LADD), which aims to improve the realism of samples from the trajectory-compressed student.

In standard LADD, the intermediate state $x_t$ used for adversarial training is obtained via the forward noising process. Following the on-policy formulation of DMD2, Sim-LADD instead obtains $x_t$ from the student's own backward denoising rollout, so the discriminator sees the states actually encountered at inference, which reduces the train--inference gap.

Following DRaFT~\citep{draft}, the adversarial gradient can be propagated through the complete differentiable denoising chain rather than only through the final sample or a single transition. In practice, truncated backpropagation through the final $k$ solver steps (DRaFT-$k$) provided comparable quality at a substantially lower memory cost. We therefore use $k = 5$, i.e., half of the student's 10 NFE.

\subsubsection{Distillation Pipeline}

\begin{figure}[htbp]
  \centering
  \includegraphics[width=\textwidth]{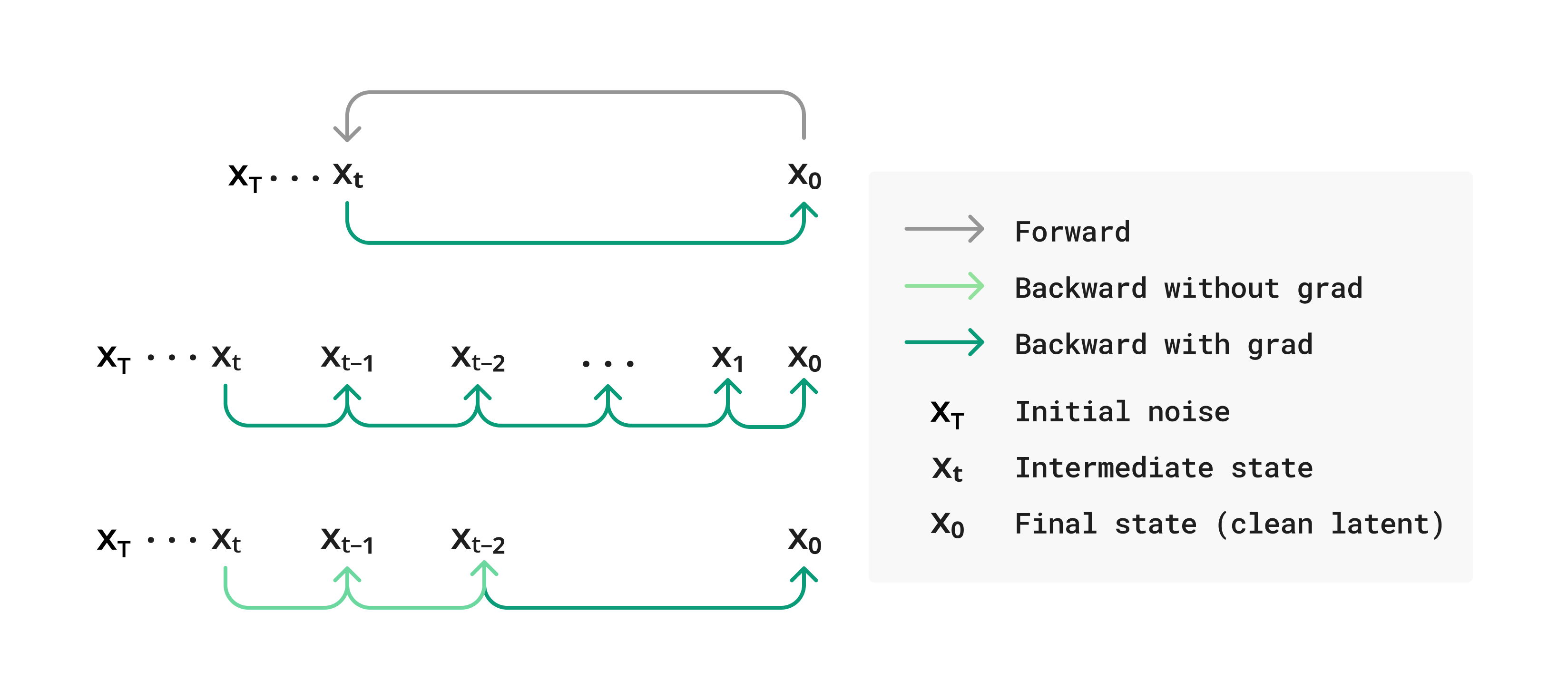}
  \caption{Rollout strategies for adversarial training. Top: standard LADD, where the intermediate state $x_t$ is obtained by forward noising of the clean sample $x_0$ and denoised in a single step with gradients. Middle: backward student rollout with gradients propagated through all denoising steps. Bottom: backward student rollout with gradients propagated only through the final steps.}
  \label{fig:rollouts}
\end{figure}

\begin{enumerate}[label=\textbf{\arabic*.}]
    \item \textbf{Trajectory compression via $\pi$-Flow.}

    We distill the teacher into a few-step student that requires only $10$ function evaluations (NFEs). Among the methods we tried, $\pi$-Flow gave the best convergence, motion and audio fidelity, and trajectory alignment, without mode collapse. Because it formulates trajectory distillation as on-policy imitation learning, it directly absorbs the teacher's CFG-guided trajectory, which removes the need for a separate CFG distillation stage.

    \item \textbf{Adversarial refinement via Sim-LADD.}

    The trajectory-compressed student shows textural and perceptual degradation caused by the coarse discretization of the sampling trajectory. We refine this checkpoint with Sim-LADD: unlike standard LADD, which builds training states via forward diffusion, Sim-LADD uses actual backward rollouts of the student, which reduces the train--inference mismatch (Figure~\ref{fig:rollouts}).

    We backpropagate the adversarial gradient through the final $k$ sampling steps, following DRaFT-$k$. As in the original LADD, we apply stochastic re-noising to both real and generated latents before discriminator feature extraction. This stabilizes adversarial training, reduces discriminator overfitting, and removes the need for an explicit $R_1$ gradient penalty such as the one used in APT~\citep{APT}.
\end{enumerate}

The two stages play complementary roles: $\pi$-Flow performs stable trajectory compression while preserving the teacher's semantics, motion, and diversity, and Sim-LADD then uses real data and adversarial learning to recover the high-frequency detail and perceptual sharpness lost during compression.

\subsubsection{Training Details}

\paragraph{Stage 1: $\pi$-Flow distillation.}

The student policy network is initialized directly from the pretrained teacher backbone and optimized with an on-policy imitation $\ell_2$ regression objective against the guided teacher velocity field.

Following the policy schedule described in $\pi$-Flow~\citep{piflow}, training proceeds through three regimes: an off-policy phase for stable initial convergence to the teacher distribution, followed by a mixed policy phase that prepares the student for on-policy training and lastly the on-policy phase that corrects the train--inference mismatch. We train for $4$k iterations, with the three regimes in a $1\!:\!1\!:\!2$ ratio.

\begin{itemize}[nosep]
    \item Integration grid: $n_{\text{grid}} = 10$;
    \item Policy steps: $128$;
    \item Optimizer: AdamW, learning rate $3\times10^{-6}$, batch size $64$;
    \item Student trajectory discretized into $N=10$ uniform integration intervals (10 NFE).
\end{itemize}

The resulting checkpoint initializes the adversarial refinement stage.

\begin{figure}[htbp]
  \centering
  \includegraphics[width=\textwidth]{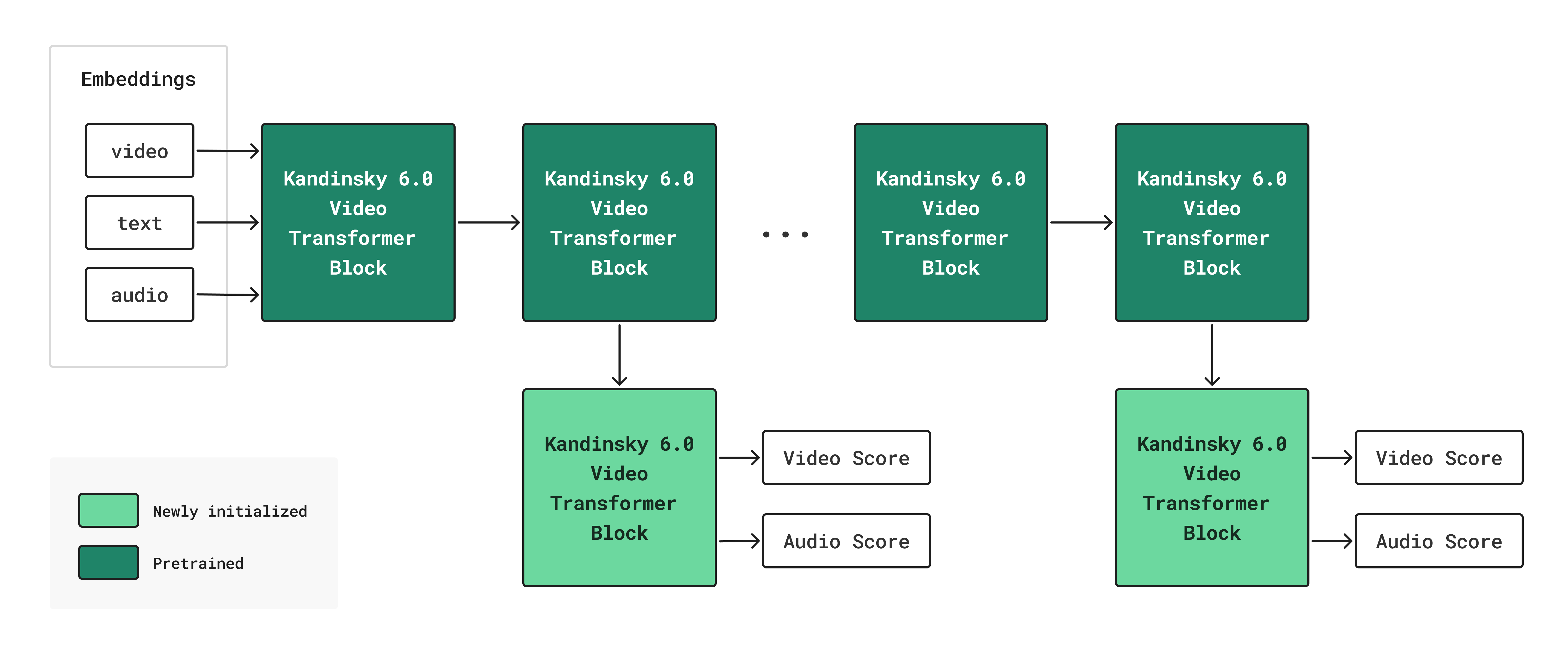}
  \caption{Discriminator architecture used in Sim-LADD. Newly initialized Kandinsky~6.0 Video transformer blocks are attached as heads to intermediate blocks of the pretrained backbone; each head outputs separate video and audio scores.}
  \label{fig:discriminator}
\end{figure}

\paragraph{Stage 2: Sim-LADD adversarial post-training.}

Adversarial refinement uses a hinge loss computed on multi-scale latent features extracted from intermediate layers of the frozen teacher backbone. Following APT~\citep{APT}, all discriminator parameters are updated during training. The discriminator heads are implemented as Fused Transformer Decoder blocks (Figure~\ref{fig:discriminator}). Image-conditioning tokens are removed from the feature sequence before the discriminator logits are computed. During the backward student rollout, gradients are propagated through the final $k=5$ solver steps, and gradient checkpointing keeps peak GPU memory consumption manageable.

Both real and generated latents are perturbed with a stochastic re-noising schedule: before discriminator feature extraction, noise levels are sampled from a fixed $\operatorname{LogitNormal}(-3.5, 1.0)$ distribution.

\begin{itemize}[nosep]
    \item Optimizer: AdamW for both the generator and the discriminator;
    \item Learning rates: $8\times10^{-7}$ (generator) and $1.5\times10^{-6}$ (discriminator); batch size 8;
    \item Gradient clipping: maximum norm $1.0$;
    \item Training length: $300$ steps.
\end{itemize}

\subsection{Super-Resolution}\label{sec:sr_training}

A single SR-DiT model is trained jointly for $\times2$ and $\times4$ video super-resolution, using training pairs from both scales with equal sampling probability. The SR-DiT and the Latent Upscalers (LU), trained separately for each scale, are optimized independently.

During SR-DiT training, the degraded LQ and HQ latents have matching spatial dimensions. For Real-ESRGAN-style degradations, this is achieved by bilinearly upsampling the low-resolution clip in pixel space before K-VAE encoding. The temporal degradation pipeline directly produces a degraded clip at the HQ resolution.

Each LU is trained to map low-resolution degraded latents to their high-resolution degraded counterparts and is used to prepare the input to SR-DiT at inference.

\subsubsection{Training Data and Degradations}\label{sec:sr_degradations}

The source videos come from the SR dataset described in
Section~\ref{sec:sr_data}.
During dataset preparation, we extract video crops of the required
spatial size, filter them, and construct degraded copies.
An accepted crop serves as the HQ target, and its degraded counterpart
provides the LQ input. Both versions are encoded with K-VAE and stored
as paired video latents.

\paragraph{Crop selection and online filtering.}
For the 4K video collection and the Kandinsky~5.0 T2V collection, we extract spatial crops of size
$512 \times 768$, $512 \times 512$, or $768 \times 512$ pixels (height $\times$ width),
retaining a sequence of frames in each crop.
Crop locations are selected using center cropping, random cropping,
optical-flow-based selection, or selection of regions with high
spatial detail.

During data preparation, dark borders are excluded from the region
available for cropping. Candidate video crops are checked for
excessive dark content, visible banding, and compression artifacts
using spatial and temporal quality measures.
The crop-selection methods are tried in a randomly sampled order,
weighted by their configured probabilities. The first candidate
that passes all checks is retained; videos for which no candidate
passes are skipped.
These crop-level checks complement the dataset-level quality
filtering described in Section~\ref{sec:sr_data}.
We use two degradation pipelines to construct LQ counterparts from the accepted HQ crops.

\paragraph{Real-ESRGAN-style degradation.}
We use a video adaptation of the two-stage high-order degradation
scheme of Real-ESRGAN~\cite{wang2021realesrgan}.
The first stage applies blur, resizing, Gaussian or Poisson noise,
JPEG compression, and video compression.
The second stage repeats blur, resizing, noise, and JPEG compression,
with weaker blur and noise settings.
A final block applies video compression and resizing to the target
low-resolution size followed by optional sinc filtering;
the order of these two components is randomized.

Blur and resize parameters and noise strength are shared across
frames, while noise is sampled independently per frame and JPEG
quality varies gradually along the video.
This pipeline degrades image quality without explicitly warping
geometry or changing the frame sequence.

The target low-resolution dimensions correspond to a
downscaling factor of 2 or 4 along each spatial axis,
depending on the encoding configuration.
For SR-DiT training, it is bilinearly upsampled back to the HQ
dimensions before K-VAE encoding.
The low-resolution version is also encoded and stored for LU training.

\paragraph{CreativeVR-style temporal degradation.}
Inspired by CreativeVR~\cite{panambur2025creativevr}, this pipeline
simulates structural and temporal artifacts in generated video.
We use the strong preset of our implementation.
A spatial grid warp is applied first, with a displacement field
interpolated between two randomly sampled grids over time.
Next, three or four operators are selected and applied in random order:
directional motion blur, adjacent-frame blending, stochastic frame
dropping with interpolation, and spatio-temporal downsampling followed
by reconstruction to the original frame count and spatial size.
Motion-blur parameters and blending coefficients follow smooth
temporal trajectories.

The degraded video retains the HQ dimensions but contains distortions
of geometry, motion, and fine detail.
This full-resolution degraded version is encoded directly for
SR-DiT training. A separate low-resolution version is obtained
by bilinear downsampling by a factor of 2 or 4 and encoded for LU training.

\subsubsection{SR-DiT Training Objective}

\begin{figure}[h!]
    \centering
    \includegraphics[width=\textwidth]{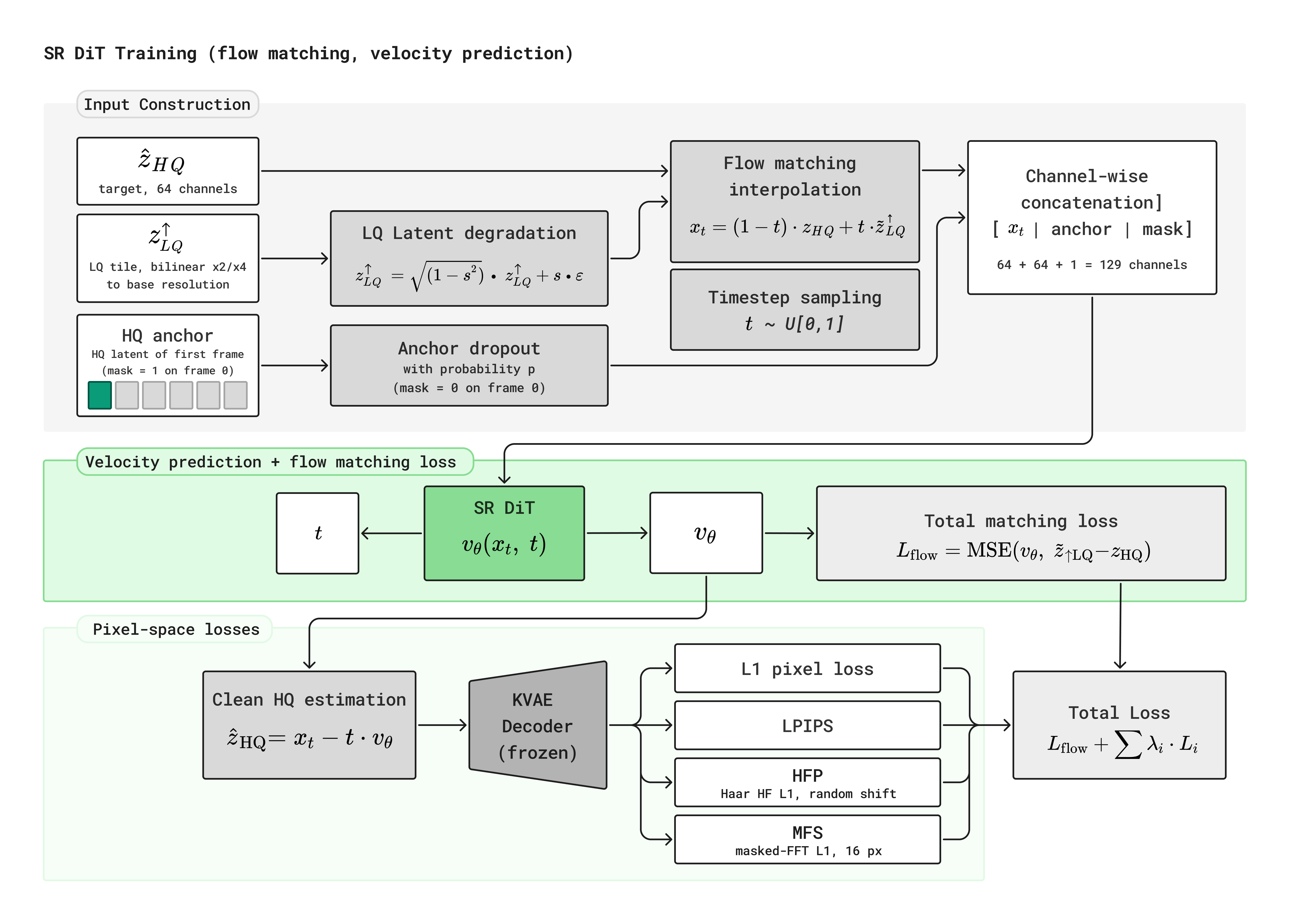}
    \caption{SR-DiT training pipeline for restoring HQ video from degraded LQ latents with first-frame HQ anchor conditioning. Training combines a latent-space flow-matching loss with pixel-space reconstruction and anti-grid losses computed through the frozen K-VAE decoder. The diagram illustrates Stages~1-2; timestep sampling uses $u \sim U[0,1]$ followed by $t = 5u/(1+4u)$.}
    \label{fig:sr_dit_training}
\end{figure}

The SR-DiT is trained with flow matching to restore HQ latents
from degraded LQ latents (Figure~\ref{fig:sr_dit_training}).
We first add Gaussian noise to the degraded LQ latent:
\[
\tilde{z}_{LQ}
= \sqrt{1-s^2}\,z_{LQ} + s\,\varepsilon,
\qquad
\varepsilon \sim \mathcal{N}(0,I),
\qquad s=0.7.
\]
The training state is constructed as
\[
x_t = (1-t)\,z_{HQ} + t\,\tilde{z}_{LQ}.
\]
In the first two training stages, we sample
$u \sim U[0,1]$ and apply the timestep transformation
$t = 5u/(1+4u)$.

Conditioned on an HQ anchor $a$ and its binary mask $m$,
the model predicts the velocity. The flow-matching loss is
\[
L_{\mathrm{flow}}
= \mathrm{MSE}\left(
v_\theta(x_t,t,a,m),\,
\tilde{z}_{LQ}-z_{HQ}
\right).
\]
All SR-DiT parameters are optimized.

\paragraph{Pixel-space losses.}
We estimate the clean HQ latent as
\[
\hat{z}_{HQ} = x_t - t\,v_\theta(x_t,t,a,m)
\]
and decode it with the frozen K-VAE decoder.
Gradients propagate through the decoder to SR-DiT.
The reconstruction is compared with the decoded HQ target.

To limit memory and computation, we select matching random
spatial crops corresponding to $256 \times 256$ pixels and
decode at most four consecutive latent frames from each of
at most two samples per GPU.
In the first two training stages, pixel-space losses are
applied only for $t \le 0.7$. We use:
\begin{itemize}
    \item Pixel L1 loss, with weight 0.6.
    \item LPIPS perceptual loss with a VGG
    backbone~\cite{zhang2018lpips}, with weight 0.35.
    \item A High-Frequency Perception (HFP) loss adapted
    from~\cite{wu2025vpdsr}: L1 between the high-frequency
    Haar subbands (LH, HL, HH) of the reconstruction and target,
    with a shared random spatial shift of 0 or 1 pixel
    along each axis. Its weight is 1.0.
    \item A masked Fourier space (MFS) loss adapted
    from~\cite{wang2026eliminating}: L1 between FFT amplitudes
    at the harmonic bins of a 16-pixel grid, matching
    the spatial compression factor of K-VAE.
    Its weight is 1.0.
\end{itemize}
HFP supervises high-frequency detail, while MFS specifically
targets periodic grid artifacts. Together, they help suppress
grid and checkerboard patterns.
The total objective is
\[
L = L_{\mathrm{flow}} + \sum_i \lambda_i L_i.
\]
The settings above apply to the first two training stages;
the distillation objective and its loss weights are described below.

\paragraph{HQ anchor.}
The anchor contains the first temporal slice of the ground-truth
HQ latent, with zeros at the remaining temporal positions.
A binary mask identifies the position containing the anchor.
The anchor and mask are concatenated with $x_t$ along the
channel dimension (Section~\ref{sec:sr_dit_arch}).

In the first two training stages, the anchor and its mask are
jointly zeroed with probability 0.15, allowing the model to
learn restoration without anchor conditioning.
The current inference pipeline uses zero anchor and mask inputs.
The conditioning interface also allows a future inference variant
to supply an HQ first-frame reference, for example using an
image super-resolution model.

\subsubsection{SR-DiT Training Stages}

The SR-DiT is trained in three stages on 8 NVIDIA H100 GPUs. All stages use AdamW with $\beta=(0.9,0.95)$, weight decay $10^{-3}$, gradient clipping at 1.0, and an LQ noise fraction of $s=0.7$. The $\times4$ and $\times2$ pair sets are sampled with equal probability, with source sequences of 20--31 latent frames. Each GPU processes a packed batch of 124 latent frames, equivalent to four 31-frame clips. Stage~3 additionally accumulates gradients over two microbatches.

\paragraph{Stage 1: training with dense attention.} The SR-DiT is initialized from the text-free Kandinsky~5.0 Video Lite (SFT) checkpoint and trained with dense attention. We use a learning rate of $10^{-5}$ with 500 warmup steps, EMA decay 0.999, and timestep scheduler scale 5.0. The checkpoint at step 7,250 is used to initialize Stage~2. This stage takes approximately 4.5 days.

\paragraph{Stage 2: NABLA fine-tuning.} To reduce attention cost, dense attention is replaced with sparse NABLA attention~\cite{mikhailov2026nabla}. We use top-CDF block selection with $P=0.8$, linearly annealed from 0.95 over the first 200 steps. This is combined with a local neighborhood following sliding-tile attention (STA)~\cite{zhang2024fast}, with a window of $11 \times 7 \times 7$ attention blocks along time, height, and width. Each spatial block contains $8 \times 8$ latent tokens. 

Training starts from the Stage~1 checkpoint at step 7,250 and runs for 6,000 steps with a learning rate of $10^{-5}$, 200 warmup steps, and the same loss settings as Stage~1. This stage takes approximately 3 days.

\paragraph{Stage 3: $\pi$-Flow distillation.} The model is distilled for inference with two SR-DiT calls using the DX policy of $\pi$-Flow~\citep{piflow}. The frozen teacher and the student backbone are initialized from the Stage~2 checkpoint at step 6,000. The student's output head is expanded to predict 10 velocity outputs and initialized by replicating the teacher's output head.

The denoising trajectory is divided into two segments. For each training sample, a segment is selected, and its starting state is constructed using the HQ and noised LQ latents, with HQ anchor conditioning as in the preceding stages. The student makes one network call at the segment start. Its 10 velocity outputs are converted into a grid of clean-latent predictions, which the DX policy interpolates to evaluate velocities within the segment without additional network calls.

The teacher is evaluated at eight states along a teacher-driven rollout within the selected segment. The DX loss is the MSE between the teacher velocity and the student policy's velocity averaged over a local integration window. The policy integration uses a resolution of 128 substeps over the full trajectory, with a minimum averaging window of three substeps.

Pixel-space losses are computed from the policy's clean-latent
estimate at the segment start, $\hat{z}_0 = x_t - \sigma_t u$,
where $\sigma_t$ is the shifted flow-matching timestep
at the segment start and $u$ is the policy velocity. These losses directly supervise the clean-latent node at the segment start; the remaining nodes are supervised through the DX loss. They are applied at all noise levels with reduced weights: L1 0.05, LPIPS 0.026, HFP 0.16, and MFS 0.033. Decoding is limited to a spatial crop corresponding to $256 \times 256$ pixels and at most four consecutive latent frames from one sample per GPU. Anchor dropout is set to 0.15 in Stages~1-2 and disabled in Stage~3.

We use a time shift of 3.5, a final segment size scale of 0.5, a learning rate of $10^{-6}$ with 100 warmup steps, gradient accumulation over two microbatches, and EMA decay 0.9999. The checkpoint used for inference is obtained after 750 distillation updates, corresponding to approximately 1.5 days.

At inference, the distilled model operates with two network function evaluations (NFE${}=2$). Within each segment, the DX policy is integrated without additional network evaluations.

\subsubsection{Latent Upscaler Training}

The LU is trained in a \emph{degraded-to-degraded} setting:
the input is the K-VAE encoding of a low-resolution degraded
video, and the target is the encoding of its full-resolution
degraded counterpart. The spatial grids differ by a factor
of 4 or 2 along each axis.
The LU is thus trained to upsample the latent representation,
while restoration of clean HQ content is handled by SR-DiT.

The weak/synthetic training mixtures contain four groups:
the Kandinsky~5.0 T2V collection with weak degradation,
the 4K collection with weak Real-ESRGAN-style degradation,
synthetic camera-motion clips constructed from images,
and the 4K collection with synthetic temporal degradation.
These mixtures contain approximately 218k pairs for $\times4$
and 194k pairs for $\times2$.

Both models are trained on 4 GPUs using packed batches
with a budget of 124 latent frames per GPU, equivalent
to four 31-frame clips.
We use AdamW with $\beta=(0.9,0.95)$,
weight decay $10^{-4}$, gradient clipping at 1.0,
and EMA decay 0.99.

\paragraph{Losses.}
The reconstruction objective combines a latent-space
Charbonnier loss ($\epsilon=10^{-3}$, weight 1.0)
with a pixel-space L1 loss (weight 1.0).
Pixel losses are computed by decoding matching crops
of the predicted and target latents with the frozen
K-VAE decoder; gradients propagate through the decoder
to the LU.

Pixel-space losses are evaluated every second training step
for one sample per GPU, using a spatial latent crop
of $16 \times 24$ cells and the first nine latent frames.
LPIPS is disabled as a training loss.

Temporal supervision matches frame-to-frame differences
between the prediction and target.
In the initial $\times4$ run, the latent temporal L1 loss
has weight 0.1. After 50k steps, its weight is increased
to 0.5, and a decoded pixel temporal L1 loss with weight 1.0
is added. The $\times2$ model uses these latter temporal
loss settings from the start.

\paragraph{$\times4$ model.}
All parameters except the output heads are initialized
from the corresponding K-VAE decoder modules,
including the input projection.
Only the output heads are optimized for the first 100 steps;
the transferred layers are then included in training.

Training consists of three consecutive runs:
50k steps with a base learning rate of $2.2 \times 10^{-4}$,
40k steps with $1.1 \times 10^{-4}$,
and a final 10k-step polishing run with $1.1 \times 10^{-5}$.
The temporal-loss changes described above are introduced
in the second run.

We use $\mu$P-inspired learning-rate scaling:
hidden and output convolution weights receive the base
learning rate multiplied by $512/C_{\mathrm{in}}$,
where $C_{\mathrm{in}}$ is the number of input channels.
Input projections, normalization and modulation parameters,
and biases retain the base rate.
Training takes approximately 60 hours on 4 GPUs.

\paragraph{$\times2$ model.}
The $\times2$ branch is initialized by copying modules
from the trained $\times4$ model
(Section~\ref{sec:latent_upscaler_arch}).
The $\times4$ path is frozen, and all parameters of
the $\times2$ branch are optimized from the first step.

The model is trained for 40k steps with a base learning rate
of $2.2 \times 10^{-4}$, followed by a 10k-step polishing run
with $2.2 \times 10^{-5}$.
The first 20k steps use the stronger-degradation pair set;
the subsequent runs use the weak/synthetic mixture
described above.
Training takes approximately two days on 4 GPUs.

\subsubsection{Validation}\label{sec:sr_validation}

\paragraph{SR-DiT metrics.}
Generated videos have no paired HQ ground truth.
We therefore monitor restoration quality using no-reference
quality metrics and additional spatial and temporal diagnostics:
\begin{itemize}
    \item \textbf{MUSIQ}~\cite{ke2021musiq} is an image quality
    metric. We use the KonIQ-10k checkpoint, score decoded
    frames independently, and average the scores over frames
    and clips; higher is better.
    As a frame-level metric, it does not directly assess
    temporal consistency.

    \item \textbf{DOVER}~\cite{wu2023exploring} is a video
    quality metric combining technical and aesthetic branches.
    The technical branch processes sampled fragments,
    while the aesthetic branch processes spatially downscaled
    video. We report the fused, normalized score in $[0,1]$;
    higher is better.

    \item \textbf{Warp error} measures motion-compensated
    differences between adjacent frames.
    We estimate forward and backward optical flow with
    RAFT~\cite{teed2020raft}, warp one frame onto the other,
    and compute the mean absolute residual over valid,
    non-occluded pixels; lower is better.
    The metric is sensitive to temporal inconsistencies,
    but also to image detail and optical-flow accuracy.
    Smooth outputs can receive low errors, so the score
    must be interpreted together with spatial quality.

    \item \textbf{Laplacian artifact score} measures excess
    high-frequency response in the SR output relative
    to the upsampled LQ input.
    Sobel edges in the LQ frame are used to exclude
    natural edges, restricting scoring to smooth regions;
    lower is better.
    The score helps identify grid and checkerboard artifacts,
    but can also respond to newly reconstructed detail.
\end{itemize}

We additionally report CLIP-IQA~\cite{wang2023clipiqa},
with higher scores indicating better predicted image quality.
MUSIQ and DOVER were developed for natural-image and
user-generated-video quality assessment, respectively,
and their scores may not fully reflect SR-specific artifacts.
We therefore use these metrics as complementary indicators,
rather than treating any single score as a complete quality measure.
Final SR quality assessment and checkpoint selection
are based on side-by-side human evaluation
at both the tile and full-frame levels.

\paragraph{SR-DiT validation sets.}
The source videos are generated by Kandinsky~5.0 Video Lite
(T2V, SFT, 5~s).
We use two separate sets of 912 video crops each.
For $\times4$, crops cover $128 \times 192$ pixels
of the source video; for $\times2$, they cover
$256 \times 384$ pixels.
Both are enlarged to the $512 \times 768$ working resolution
before K-VAE encoding.
Evaluation is performed on individual crops, without
full-frame tiling and blending.

Results are reported in Table~\ref{tab:sr_dit_stages}.
The two scales use different source crops and therefore
should be compared separately.
For the same output dimensions, the $\times4$ input contains
four times fewer pixels than the $\times2$ input and covers
a smaller region of the source video.
Consequently, the difference in absolute metric scores
between scales does not isolate model quality.

\begin{table}[htbp]
\centering
\caption{Validation metrics of the SR-DiT after each training stage on the tiled validation sets (912 videos each). Stage~1: dense attention; Stage~2: NABLA fine-tuning; Stage~3: $\pi$-Flow distillation. Values are comparable only within one scale factor.}
\label{tab:sr_dit_stages}
\begin{tabular}{l l r r r r r}
\toprule
Scale & Stage & MUSIQ $\uparrow$ & DOVER $\uparrow$ & CLIP-IQA $\uparrow$ & Lapl. artifacts $\downarrow$ & Warp error $\downarrow$ \\
\midrule
\multirow{3}{*}{$\times 4$} & Stage 1 & 48.92 & 0.5467 & 0.4426 & 0.4366 & 0.6141 \\
 & Stage 2 & 50.38 & 0.5589 & 0.4569 & 0.5337 & 0.6478 \\
 & Stage 3 & 50.41 & 0.5553 & 0.4587 & 0.4227 & 0.6781 \\
\midrule
\multirow{3}{*}{$\times 2$} & Stage 1 & 66.05 & 0.7329 & 0.5950 & 0.6062 & 0.6551 \\
 & Stage 2 & 65.29 & 0.7282 & 0.5932 & 0.6072 & 0.6531 \\
 & Stage 3 & 65.81 & 0.7348 & 0.5906 & 0.5710 & 0.6831 \\
\bottomrule
\end{tabular}
\end{table}

Within each scale, we observe the following:
\begin{itemize}
    \item At $\times4$, Stages~2 and~3 improve MUSIQ
    from 48.92 to 50.38 and 50.41, respectively,
    and also increase CLIP-IQA relative to Stage~1.
    At $\times2$, MUSIQ remains between 65.29 and 66.05,
    with no consistent improvement across stages.

    \item At $\times4$, warp error increases
    from 0.614 to 0.648 and 0.678.
    This indicates a higher motion-compensated residual.
    Since the metric also depends on spatial detail,
    the temporal-quality implications require visual assessment.

    \item At $\times4$, the Laplacian artifact score rises
    from 0.437 after Stage~1 to 0.534 after Stage~2,
    then decreases to 0.423 after Stage~3.
    This indicates reduced excess high-frequency response
    after distillation, although the metric alone cannot
    distinguish artifacts from reconstructed detail.
\end{itemize}

\paragraph{Latent Upscaler validation.}
During training, we monitor latent reconstruction error
and reference-based quality metrics on a validation subset
balanced across the four data groups.
Checkpoints are selected by PSNR between decoded predictions
and decoded full-resolution degraded targets.
We monitor latent L1 error every 2,500 steps,
and PSNR, SSIM, and LPIPS~\cite{zhang2018lpips}
with an AlexNet backbone every 10k steps.
During the final $\times2$ polishing run, decoded reference
metrics are evaluated every 5k steps.

For offline evaluation, we use 94 held-out clips whose content
was excluded from training across the data sources.
The set contains four groups:
25 clips from the Kandinsky~5.0 T2V collection with weak degradation
(source resolutions $1920 \times 1080$ and $1080 \times 1080$);
19 clips from the 4K video collection with weak
Real-ESRGAN-style degradation (mostly 4K);
25 synthetic camera-motion clips constructed from images;
and 25 clips from the 4K video collection with synthetic
temporal degradation
(source resolutions up to $7680 \times 4320$).
The $\times2$ and $\times4$ sets contain the same clips
in the same order, enabling paired comparisons between scales.
The task is degraded-to-degraded.
Evaluation uses decoded video crops of
$512 \times 512$, $512 \times 768$, or $768 \times 512$ pixels
with 89--121 frames, rather than full-resolution source videos.

\begin{table}[htbp]
\centering
\caption{PSNR (dB) of the Latent Upscaler on the held-out validation set, computed per frame on pixels decoded by K-VAE and averaged per clip. The baseline decodes the low-resolution latent and upsamples it bilinearly in pixel space; $\Delta$ is the difference between the LU and the baseline.}
\label{tab:lu_psnr}
\begin{tabular}{l r r r r r r r}
\toprule
 & & \multicolumn{3}{c}{$\times 2$} & \multicolumn{3}{c}{$\times 4$} \\
\cmidrule(lr){3-5} \cmidrule(lr){6-8}
Group & $n$ & LU & Bilinear & $\Delta$ & LU & Bilinear & $\Delta$ \\
\midrule
Kandinsky~5.0 T2V, weak & 25 & 41.60 & 42.97 & $-1.37$ & 39.19 & 39.16 & $+0.03$ \\
4K collection, weak Real-ESRGAN & 19 & 41.42 & 42.37 & $-0.95$ & 39.91 & 39.13 & $+0.78$ \\
Camera-motion clips (images) & 25 & 38.62 & 38.26 & $+0.36$ & 37.85 & 36.83 & $+1.02$ \\
4K collection, synthetic & 25 & 45.08 & 42.92 & $+2.16$ & 40.25 & 36.98 & $+3.26$ \\
\midrule
\textbf{All} & 94 & \textbf{41.70} & \textbf{41.58} & $\mathbf{+0.11}$ & \textbf{39.26} & \textbf{37.96} & $\mathbf{+1.30}$ \\
\bottomrule
\end{tabular}
\end{table}

As Table~\ref{tab:lu_psnr} shows, the $\times4$ LU achieves
higher mean PSNR than pixel-space bilinear upsampling
in all four groups.
At $\times2$, it improves PSNR on the image-derived
camera-motion clips and the 4K collection synthetic-degradation
group, while pixel-space bilinear upsampling scores higher
on the two weak-degradation video groups.
The pixel-space baseline decodes the low-resolution latent
and then bilinearly upsamples the decoded video.
By contrast, bilinear interpolation directly in latent space,
followed by decoding, yields only 21.44~dB at $\times2$
and 19.76~dB at $\times4$, showing that this simple
latent-space baseline reconstructs the target poorly.

We also measure warp error using RAFT-small~\cite{teed2020raft}
over all adjacent frame pairs, without spatial downsampling.
On the 94-clip set, the LU obtains mean errors of
0.684 at $\times2$ and 0.626 at $\times4$,
compared with 0.638 and 0.587 for pixel-space bilinear upsampling.
These differences require cautious interpretation:
warp error depends on spatial detail and optical-flow accuracy,
and smoothing can reduce the residual.
Moreover, the degraded-to-degraded target may itself contain
temporal artifacts introduced by the degradation pipeline,
so this set does not provide a clean reference
for temporal restoration quality.

We therefore additionally evaluate 24 real clips without
synthetic degradation, covering foliage, fur, structure,
and textile, with camera or object motion.
Both LU routes are evaluated on a common output canvas;
the $\times2$ route includes pixel-space pre-upsampling.
The comparison baseline encodes and decodes the source video
with the same K-VAE checkpoint, then bilinearly upsamples
the reconstruction to the evaluation canvas.

The round-trip baseline has a mean warp error of approximately
0.591. Relative to this baseline, the LU routes increase
the mean error by 0.026 at $\times2$ and 0.049 at $\times4$,
approximately 4.4\% and 8.3\%, respectively.
For the structure category, the $\times4$ difference
is approximately 0.002.
These results show a modest average increase in
motion-compensated residual relative to the round-trip baseline,
but do not uniquely attribute temporal error to either
the VAE or the LU.
Textile has the highest absolute warp error for both routes. %
\clearpage

\section[Inference]{\color{Green}Inference}\label{sec:inference}

\begin{figure}[h!]
    \centering
    \includegraphics[width=\textwidth]{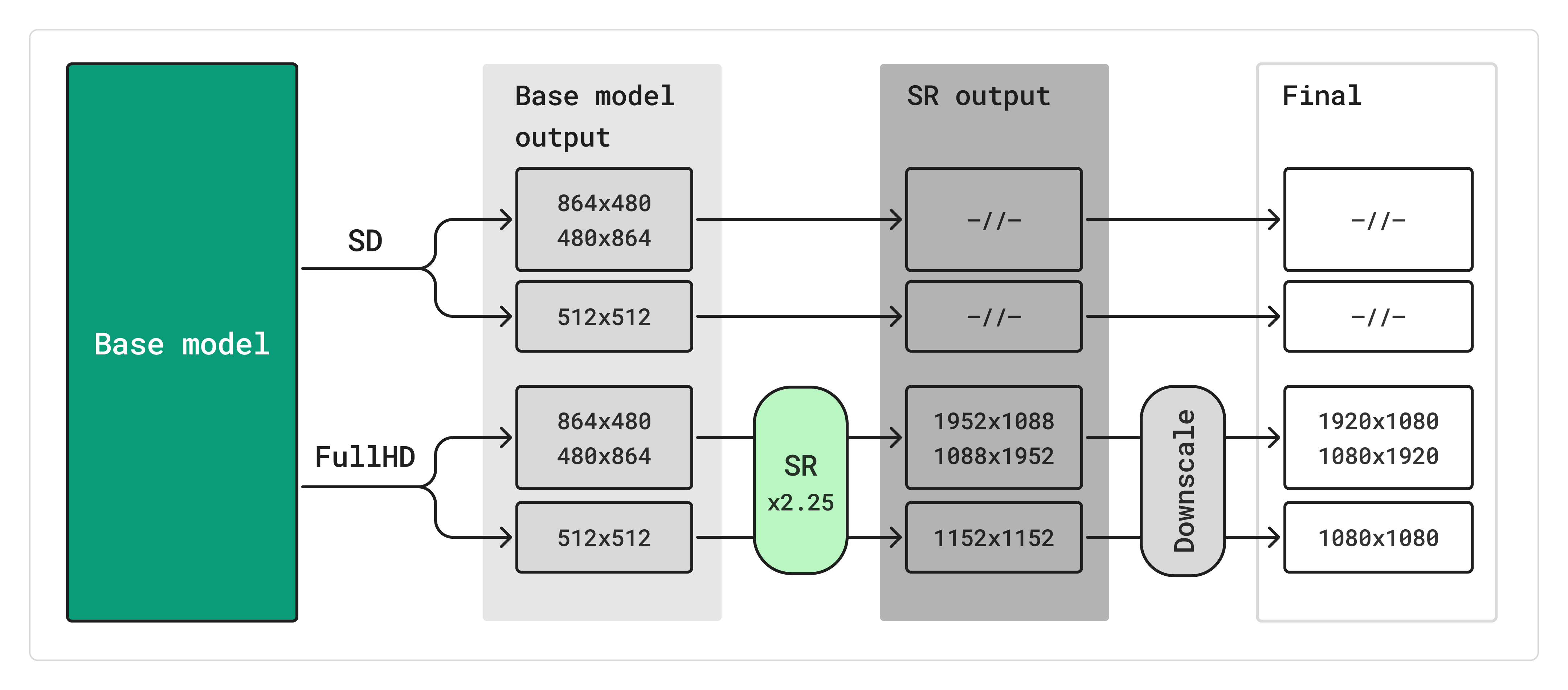}
    \caption{Output resolutions of Kandinsky~6.0 Video at inference. In SD mode, the base-model output is returned without changes; in Full HD mode, it is upscaled $\times$2.25 by the super-resolution model and then downscaled to the target Full HD resolution.}
    \label{fig:inference_resolutions}
\end{figure}

Figure~\ref{fig:inference_resolutions} summarizes the output resolutions supported at inference (given as width $\times$ height). In SD mode, the output of the base model (864$\times$480, 480$\times$864, or 512$\times$512) is returned as is. In Full HD mode, the base-model output is upscaled by a factor of 2.25 with the super-resolution model (Section~\ref{sec:SR_architecture}) and then downscaled to 1920$\times$1080, 1080$\times$1920, or 1080$\times$1080.

\begin{figure}[h!]
    \centering
    \includegraphics[width=\textwidth]{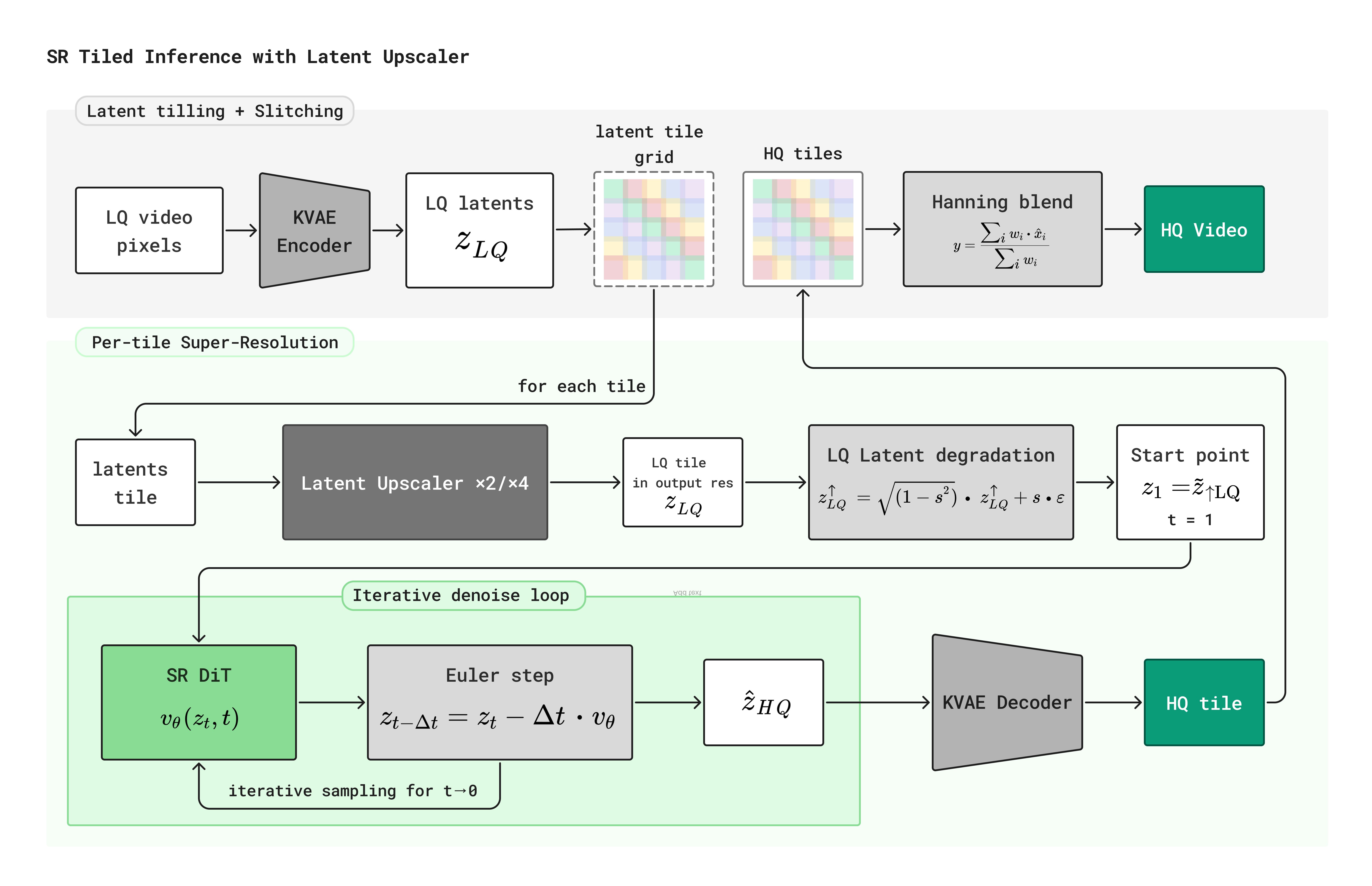}
\caption{Tiled SR inference with the Latent Upscaler. The LQ video is encoded once with K-VAE and split into overlapping latent tiles. Each tile is upscaled by the $\times2$ or $\times4$ LU, refined by SR-DiT, decoded, and merged using Hann-window blending. The denoising loop shown corresponds to the pre-distillation model; the final distilled model replaces this loop with $\pi$-Flow DX policy sampling using two network evaluations (NFE${}=2$).}
    \label{fig:sr_inference}
\end{figure}

The SR inference pipeline is shown in Figure~\ref{fig:sr_inference}.
The figure illustrates the common encoding, tiling, upscaling,
and decoding stages, with the Euler denoising loop used by
the pre-distillation model. The final distilled model replaces
this loop with $\pi$-Flow DX policy sampling.
The pipeline consists of the following steps:

\begin{enumerate}
    \item \textbf{Pre-upscaling (only for $\times2.25$).}
    The generated video is bilinearly upscaled by $\times1.125$
    in pixel space, with the dimensions aligned to the spatial
    compression factor of K-VAE (16~px).
    For example, an $864 \times 480$ video becomes
    $976 \times 544$. The $\times2$ route is then applied.

    \item \textbf{Encoding.}
    The whole clip is encoded with K-VAE once.
    A 121-frame video is mapped to 31 latent frames.

    \item \textbf{Latent tiling.}
    The latent representation is split into overlapping spatial
    tiles, using a minimum overlap of 25\%.
    Each tile covers the full temporal extent of the clip.
    For a $512 \times 768$ working resolution, input tiles contain
    $16 \times 24$ latent cells for the $\times2$ route
    and $8 \times 12$ cells for the $\times4$ route,
    corresponding to source regions of $256 \times 384$
    and $128 \times 192$ pixels, respectively.
    For square content, the corresponding input tiles contain
    $16 \times 16$ or $8 \times 8$ latent cells.

    \item \textbf{Latent upscaling.}
    The scale-specific Latent Upscaler brings each tile
    to the SR-DiT working resolution:
    $32 \times 48$ latent cells for landscape content
    or $32 \times 32$ cells for square content.
    Portrait content uses the transposed spatial dimensions.

    \item \textbf{Denoising.}
    The upscaled LQ latent is partially noised using
    the same noise fraction as in training ($s=0.7$),
    giving the starting state at $t=1$:
    \[
    z_1 = \sqrt{1-s^2}\,z^{\uparrow}_{LQ} + s\,\varepsilon,
    \qquad \varepsilon \sim \mathcal{N}(0,I).
    \]
    The Stage~2 model uses an Euler sampler, typically with 4--8 network evaluations (NFE), predicting the velocity at each step and updating the latent until $t=0$.

    The final Stage~3 model instead uses $\pi$-Flow DX policy
    sampling~\citep{piflow} with two network evaluations
    (NFE${}=2$).
    At the start of each of two trajectory segments,
    SR-DiT predicts 10 velocity outputs, which are converted
    into clean-latent nodes defining a DX policy.
    The policy is integrated within the segment using
    network-free Euler substeps: velocities are computed
    from the interpolated nodes without additional
    SR-DiT evaluations.
    Thus, NFE${}=2$ denotes two network evaluations per tile,
    rather than two Euler integration steps.

    Both models are text-free, with classifier-free guidance
    disabled. The HQ anchor and its mask are set to zero
    at inference.

    \item \textbf{Decoding and blending.}
    Each restored latent tile is decoded with K-VAE.
    The overlapping decoded tiles are combined using
    normalized Hann-window weights to reduce boundary seams.

    \item \textbf{Final resizing.}
    The assembled SR output is resized without cropping
    to the target resolution
    (Figure~\ref{fig:inference_resolutions}).
    Depending on the route, the downscaling factor ranges
    from approximately $\times1.01$
    ($1952 \times 1088 \to 1920 \times 1080$)
    to $\times1.42$
    ($2048 \times 2048 \to 1440 \times 1440$).
\end{enumerate}

Kandinsky~6.0 Video uses the $\times2.25$ routes
to produce Full HD output
(Figure~\ref{fig:inference_resolutions}).
The SR pipeline also supports 2K output
via the $\times4$ routes.

Table~\ref{tab:sr_routes} lists the supported routes and
the processing time of one 5-second clip
(121 frames at 24~fps) on a single NVIDIA H100 GPU
with NABLA attention and K-VAE decoding accelerated
by the Magi compiler.
All measurements use the distilled SR-DiT with $\pi$-Flow
DX sampling (NFE${}=2$) and the scale-specific Latent Upscaler.
The whole clip is encoded with K-VAE once.
Timings include K-VAE encoding and exclude the final resize.
Tiles are processed sequentially with a tile batch size of one.

\begin{table}[htbp]
\centering
\caption{Super-resolution routes and processing time per
5-second clip on a single H100 GPU, using the Latent Upscaler,
the distilled SR-DiT (NFE${}=2$), and the Magi compiler~\cite{magi_compiler_2026}
for K-VAE decoding. Times include K-VAE encoding and exclude
the final resize. Spatial dimensions are given as width
$\times$ height.}
\label{tab:sr_routes}
\begin{tabular}{l l c l l c r}
\toprule
Target & Source & Scale & SR output & Final & Tiles & Time, s \\
\midrule
Full HD
& $864 \times 480$ & $\times2.25$
& $1952 \times 1088$ & $1920 \times 1080$
& 9 & 39.6 \\

Full HD
& $512 \times 512$ & $\times2.25$
& $1152 \times 1152$ & $1080 \times 1080$
& 9 & 24.7 \\

2K
& $864 \times 480$ & $\times4$
& $3456 \times 1920$ & $2560 \times 1440$
& 30 & 130.8 \\

2K
& $768 \times 432$ & $\times4$
& $3072 \times 1728$ & $2560 \times 1440$
& 25 & 108.8 \\

2K
& $512 \times 512$ & $\times4$
& $2048 \times 2048$ & $1440 \times 1440$
& 25 & 68.7 \\
\bottomrule
\end{tabular}
\end{table}

Inference cost depends on the working tile resolution
and the number of tiles.
The distilled model uses two SR-DiT evaluations per tile;
DX policy integration within each segment does not require
additional network evaluations.
The assembled output is resized on the CPU to the target
resolution without cropping.

\section[Optimizations and Consumer GPU Deployment]{\color{Green}Optimizations and Consumer GPU Deployment}\label{sec:optimizations}

Throughout this section, \textbf{Lite} and \textbf{Pro} denote Kandinsky~6.0
Video Lite and Kandinsky~6.0 Video Pro, and \textbf{SD} and
\textbf{Full HD} denote the two operating modes defined in
Section~\ref{sec:inference}; a configuration such as \textbf{Pro~SD} refers
to the Kandinsky 6.0 Video Pro model in SD mode. All base-model measurements in this section are obtained with the full, non-distilled model unless explicitly marked as Distill.

Peak allocated memory during a Pro~SD run reaches 72.8\,GiB, which does not
fit on any consumer GPU. Both Lite and Pro run on 32, 24, and 16\,GB cards
through weight offloading: only two transformer blocks reside on the GPU at
a time, the rest are streamed from CPU memory during the forward pass. The
offloading mechanism itself is identical for both model sizes; only the
number of blocks and the resulting resident footprint differ. The
mathematical model is unchanged. Peak memory drops by about $3.4\times$ for
Pro at SD and about $1.8\times$ for Pro at Full HD; for Lite the
reduction is approximately $2\times$ at SD and $1.1\times$ at Full HD. Total generation time on the reference H100 changes by a few
percent.

\subsection{Block Offloading}

The transfer of the next block is overlapped with the computation of the current one on a separate CUDA stream. Embeddings, RoPE, and output projections remain resident on the GPU for the duration of the DiT stage. When MagCache skips a step, the weights of the skipped blocks are not transferred at all.

Peak PyTorch allocated memory is summarized in
Table~\ref{tab:block_offload_memory}. These values are measured with the
default SR decoding configuration; the presets described below report lower
peaks because they use a shorter SR decode segment and, at 16\,GB,
additional spatial splitting of the SR decoder.

\begin{table}[htbp]
\centering
\caption{Peak PyTorch allocated memory (GiB) on the H100 with and without
block offloading, for both model sizes and both operating modes.}
\label{tab:block_offload_memory}
\begin{tabular}{l c c}
\toprule
 & Resident & Block offload \\
\midrule
Pro, SD             & 72.8 & 21.7 \\
Pro, Full HD        & 72.8 & 39.8 \\
Lite, SD            & 43.5 & 21.7 \\
Lite, Full HD       & 44.8 & 39.8 \\
\bottomrule
\end{tabular}
\end{table}

The mode is inference-only. It is not compatible with CUDA graphs, whole-model compilation, or externally exported weights, including the Magi SR backend~\cite{magi_compiler_2026}. PyTorch 2.13 is required: with PyTorch 2.10, offloading breaks the compiled forward pass. 

Base-model outputs match the resident variant to within the numerical epsilon of the data type: video at SD, and both frames and audio at Full HD.

\subsection{VRAM Budget Presets}

Three presets are provided, for 32, 24, and 16\,GB cards. All three use block offloading. The 32 and 24\,GB presets share the same settings: extra headroom above 24\,GB does not translate into speed. The 16\,GB preset additionally splits SR decoding along the spatial dimension and quantizes the text encoder.

The settings of each preset are listed in Table~\ref{tab:vram_presets_settings}.

\begin{table}[htbp]
\centering
\caption{Settings of the VRAM budget presets for 32/24\,GB and 16\,GB GPUs.}
\label{tab:vram_presets_settings}
\begin{tabular}{l c c}
\toprule
 & 32 / 24\,GB & 16\,GB \\
\midrule
DiT block offloading      & yes & yes \\
SR decode segment length  & 4 frames & 4 frames \\
SR spatial split          & none & 2 strips \\
Text encoder              & full & NF4, released after encoding \\
\bottomrule
\end{tabular}
\end{table}

Peak allocated memory is the same for Lite and Pro: only two blocks remain
on the GPU, so model size no longer drives memory. Peak allocated memory at each preset is reported in
Table~\ref{tab:vram_presets_memory}.

\begin{table}[htbp]
\centering
\caption{Peak PyTorch allocated memory (GiB) at the 32/24\,GB and 16\,GB
presets. Peak memory is the same for Lite and Pro, because only two DiT
blocks remain resident.}
\label{tab:vram_presets_memory}
\begin{tabular}{l c c}
\toprule
 & 32 / 24\,GB & 16\,GB \\
\midrule
SD        & 21.7 & 7.9 \\
Full HD   & 21.7 & 12.3 \\
\bottomrule
\end{tabular}
\end{table}

On a real 16\,GB card the process occupies 15.1\,GiB out of 15.6\,GiB free.
Host RAM must be large enough to hold the full set of model weights together
with their pinned copies. If some of them are paged out to disk, block
streaming becomes disk-bound and generation time degrades sharply. At the
16\,GB preset, Pro requires about 58\,GiB of process RSS. Decode segments
shorter than four frames do not reduce memory further; beyond that point the
constraint is spatial, not temporal.

\subsection{Inference Acceleration}

Four mechanisms are in use.

\textbf{Regional compilation.} Base DiT blocks are compiled individually through \texttt{torch.compile}. The model is not compiled as a whole. Compilation is enabled in all configurations reported in Table~\ref{tab:performance}, for both Lite and Pro and for both the full and distilled models.

\textbf{MagCache.} MagCache~\cite{ma2025magcachefastvideogeneration} skips a subset of sampling steps based on the magnitude of the update. On Pro SD it evaluates 49 of 100 forwards and roughly halves the base stage.

\textbf{Attention backends.} On the H100 visual attention is computed with
FlashAttention~3~\cite{shah2024flashattention3fastaccurateattention}. On all
other tested GPUs --- the consumer cards (RTX 4090, 5060\,Ti, 5080, 5090), the RTX PRO 6000 workstation card, and the A100~80\,GB datacenter card --- visual self-attention is computed with SageAttention2++~\cite{zhang2025sageattention2pp}; FlashAttention~3 does not support these architectures. Text blocks and cross-attention use FlashAttention~2~\cite{dao2023flash2}. SR uses its own attention mechanism from the checkpoint, including NABLA~\cite{mikhailov2026nabla}, and does not switch together with the base.

\textbf{SR VAE compilation.} The SR VAE decoder is compiled regionally and by default roughly halves the SR stage. Disabling it saves a small amount of memory, from 22.9 to 20.3\,GiB.

\subsection{Performance}

The times below are for a 5-second clip, after warmup, excluding weight
loading and MP4 encoding. The large-memory cards (RTX PRO 6000 with 96\,GB,
A100 80\,GB, H100 with 80\,GB) run with module offloading instead of block
offloading; the consumer cards (RTX 4090, 5060\,Ti, 5080, 5090) run with
block offloading at the preset that fits their memory budget: the 32\,GB
preset for the RTX 5090, the 24\,GB preset for the RTX 4090, and the 16\,GB
preset for the RTX 5080 and RTX 5060\,Ti. 

\begin{table}[htbp]
\caption{Generation times in seconds for a 5-second clip with both the full and distilled base models, after warmup, excluding weight loading and MP4 encoding. Two operating modes (SD and Full HD) are evaluated for each configuration. }
\label{tab:performance}
\centering
\setlength{\tabcolsep}{3pt}
\begin{tabular}{l c c c c c c c}
\toprule
 & RTX 4090 & RTX 5060\,Ti & RTX 5080 & RTX 5090 & RTX PRO 6000 & A100 80\,GB & H100 \\
\midrule
Lite Distill SD       & 68.0   & 173.7  & 79.0   & 44.2   & 57.8   & 78.6   & 31.2 \\
Lite Distill Full HD  & 208.9  & 608.4  & 263.0  & 140.3  & 213.2  & 207.1  & 78.0 \\
Lite SD               & 267.9  & 834.9  & 370.0  & 201.3  & 258.7  & 371.6  & 143.2 \\
Lite Full HD          & 408.7  & 1286.7 & 555.0  & 296.2  & 414.1  & 508.0  & 192.2 \\
Pro Distill SD        & 235.0  & 717.6  & 320.0  & 174.5  & 175.6  & 236.0  & 87.8 \\
Pro Distill Full HD   & 381.0  & 1153.0 & 502.0  & 269.5  & 331.0  & 375.4  & 133.2 \\
Pro SD                & 955.4  & 3143.6 & 1404.0 & 768.0  & 727.5  & 980.4  & 351.5 \\
Pro Full HD           & 1096.7 & 3571.4 & 1585.0 & 862.8  & 882.9  & 1109.8 & 399.7 \\
\bottomrule
\end{tabular}
\end{table}

\vspace{-0.5em}
{\noindent\small\textit{Attention stack differs by architecture: RTX 4090,
RTX 5060\,Ti, RTX 5080, RTX 5090, RTX PRO 6000, and A100~80\,GB use
SageAttention2++ for visual self-attention; H100 uses FlashAttention~3.
All configurations use regional compilation. MagCache is enabled for all non-distilled configurations (both Lite and Pro) and is not used with the distilled models. Timings across different attention stacks are not directly comparable.}}
\vspace{1.5em}

Distillation provides a substantial speedup across all hardware configurations. For example, on the RTX 4090, Pro SD generation time drops from 955.4\,s (full) to 235.0\,s (distilled), representing a $4.1\times$ acceleration. Similarly, Pro Full HD time is reduced from 1096.7\,s to 381.0\,s ($2.9\times$ speedup). Importantly, this acceleration does not come at the cost of quality: as demonstrated by the side-by-side human evaluation (Section~\ref{sec:ablation-distill-vs-full}, Figure~\ref{fig:sbs-distill-vs-full}), the distilled model maintains overall quality on par with the full model, and in I2AV mode it even shows significant improvements in reference image alignment and artifact reduction. This combination of speed and fidelity makes high-quality generation feasible even on consumer-grade GPUs.

H100 timings do not transfer to consumer cards directly: the CPU--GPU bus
and the attention kernels differ. The comparison in the table is therefore
between hardware configurations, not between equivalent software stacks.

On the RTX 5090, SD and Full HD times agree to within 1\% for the base stage: they share the same base geometry. The SR stage depends on resolution, not on Lite or Pro: around 96\,s at Full HD.

\subsection{Quality Under Optimization}

SR memory settings --- segment length, spatial strips, offloading --- change the output at the level of rounding noise. Approximately 12\% of pixels differ by one brightness level, PSNR is around 57\,dB, and no visible structure or seams appear at segment boundaries.

Bitwise equality of SR frames between offloading modes to the epsilon of the data type has not been established. Disabling decoder compilation does not remove the difference.

NF4 quantization of the text encoder is the only preset setting that changes the clip itself: a different textual representation produces a different video. It is enabled only at 16\,GB. The same prompt run on CPU with full weights takes about 10\,s and reproduces the 32\,GB output. SageAttention2++ quantizes visual attention separately and is not tied to this preset.

%
\section[{Results}]{\color{Green}Results}\label{sec:evaluation}

\subsection{Benchmarking}

\begin{figure}[htbp]
    \centering
    \includegraphics[width=0.75\textwidth]{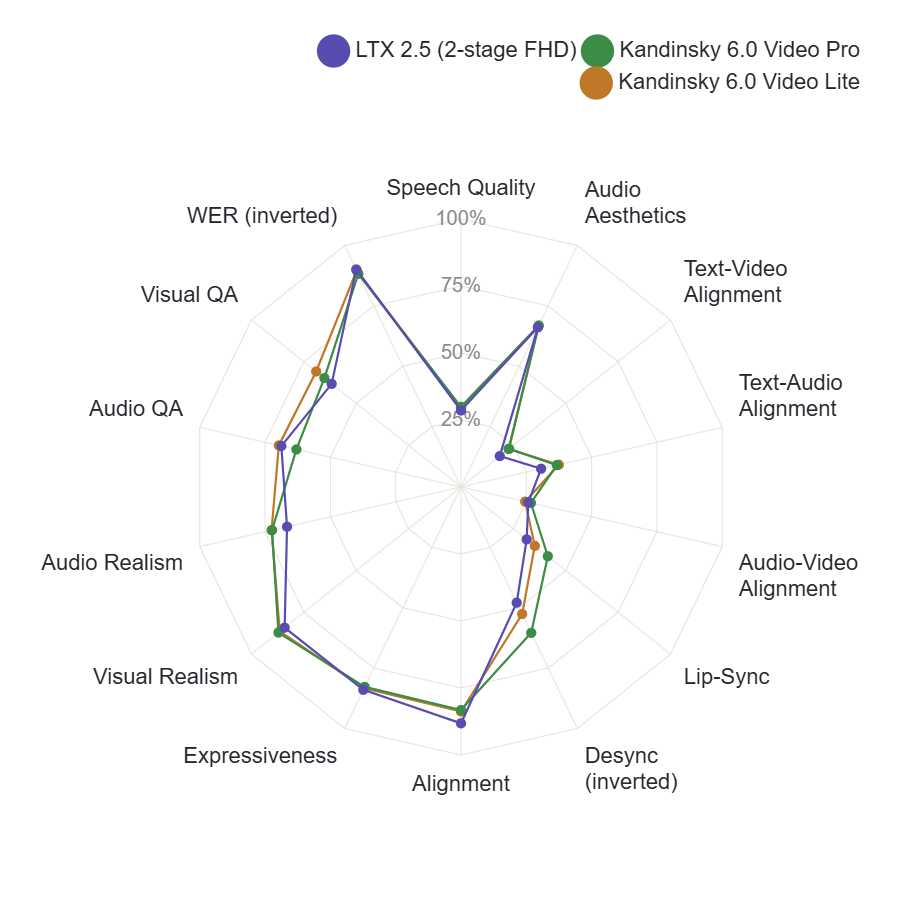}
    \caption{VABench radar chart. All metrics are normalized, with
    desync and word error rate inverted so that a larger radius always
    indicates better performance.}
    \label{fig:vabench_radar}
\end{figure}

We evaluate the final Kandinsky~6.0 Video checkpoints on
VABench~\cite{vabench2025}, a benchmark for text-to-audio-video generation
that covers audio quality, cross-modal alignment, synchronization, realism,
and expressiveness. Results are reported for \textbf{Kandinsky~6.0 Video Pro}
and \textbf{Kandinsky~6.0 Video Lite} after RL post-training, both at
$512 \times 768$ resolution, alongside \textbf{LTX~2.5} in its two-stage
Full HD configuration. Scores are listed in Table~\ref{tab:vabench} and
visualized as a radar chart in Figure~\ref{fig:vabench_radar}.

Kandinsky~6.0 Video Pro leads on speech quality, audio aesthetics, text-video and
audio-video alignment, lip-sync accuracy, desynchronization, and visual
realism. Kandinsky~6.0 Video Lite shows the strongest results on text-audio
alignment, audio realism, and the audio and visual quality scores. LTX~2.5
retains an advantage on the VLM-judged alignment and expressiveness scores,
and on word error rate.

The word error rate is not part of VABench. We compute it separately on the
speech subset: audio is transcribed with Qwen3-ASR-1.7B, punctuation is
removed from both the ASR output and the reference prompt, and WER is
computed with the \texttt{jiwer} library.
Note that this differs from the
WER reported in Section~\ref{sec:rl}, which uses Whisper-large-v3, one of
the RL reward models; the two numbers are not directly comparable and should
not be read as conflicting.

\begin{table}[htbp]
\centering
\caption{VABench results. Higher is better for all metrics except desync and
word error rate. Best per row in bold, second best underlined.}
\begin{tabular}{l c c c}
\toprule
 & LTX 2.5 & \shortstack{Kandinsky 6.0 Video Lite} & \shortstack{Kandinsky 6.0 Video Pro} \\
\midrule
speech\_qn          & 1.427 & \underline{1.477} & \textbf{1.496} \\
audio\_aes          & 3.307 & \underline{3.323} & \textbf{3.345} \\
tv\_align           & 0.185 & \underline{0.227} & \textbf{0.229} \\
ta\_align           & 0.307 & \textbf{0.374} & \underline{0.367} \\
av\_align           & \underline{0.258} & 0.245 & \textbf{0.268} \\
lipsync             & 1.567 & \underline{1.761} & \textbf{2.072} \\
desync              & 0.521 & \underline{0.474} & \textbf{0.396} \\
alignment           & \textbf{4.527} & \underline{4.351} & 4.329 \\
expressiveness      & \textbf{4.360} & \underline{4.334} & 4.312 \\
visual\_realism     & 4.366 & \underline{4.459} & \textbf{4.485} \\
audio\_realism      & 3.663 & \textbf{3.900} & \underline{3.893} \\
audio\_qa           & \underline{0.688} & \textbf{0.697} & 0.630 \\
visual\_qa          & 0.617 & \textbf{0.691} & \underline{0.652} \\
WER                 & \textbf{0.099} & \underline{0.103} & 0.117 \\
\bottomrule
\end{tabular}
\label{tab:vabench}
\end{table}

\subsection{Human Evaluation}

\subsubsection{Methodology}

The human evaluation is performed as a side-by-side (SBS) comparison of pairs of videos generated from the same prompt by two different models. The annotator is shown both videos simultaneously and can freely replay, pause, scrub, and zoom each of them in order to inspect differences in motion, detail, and audio. The videos are anonymized, their left/right positions are randomized per task, and the order of tasks is shuffled.

The evaluation is organized as a single annotation session with a fixed question sequence. Initially, audio is disabled for both videos, and the annotator answers the visual questions without sound. The annotator is then asked to enable sound and answer the remaining questions, which cover speech quality, audio-video synchronization, overall audio quality, sound-prompt correspondence, and overall task solving. For image-to-video comparisons, an additional question about correspondence to the input image is asked.

Each question targets a specific criterion and offers a fixed set of answer options, including a symmetric tie option (``Equally'', ``Both fully correct'', etc.) and, where applicable, an ``N/A'' option for cases where the criterion cannot be evaluated (e.g., no speech in either video). A majority-vote aggregation is applied across annotators. Approximately 200 or more pairwise comparisons are collected for each evaluated criterion, with the exact number depending on the comparison.

\subsubsection{Evaluation Criteria}

The annotator answers the following questions.

\textbf{Visual questions} (audio disabled):

\begin{enumerate}
    \item \textbf{Overall Visual Quality.} Which video is preferred visually, without taking the prompt into account? The annotator considers composition, lighting, colour and contrast, and overall visual appeal.
    \item \textbf{Artifacts and Semantic Breaks.} Which video contains fewer artifacts and fewer unintended semantic breaks? Artifacts include extra limbs, fingers, unrelated objects, watermarks, unrelated text, grain, flicker, texture changes without a reason, and loss of face identity. Semantic breaks include unintended object transformations and sudden appearances or disappearances of objects. The prompt is available for reference to distinguish artifacts from requested content.
    \item \textbf{Motion Realism and Physics.} Which video has more realistic object motion and better adheres to the rules of the physical world? The annotator evaluates trajectories of limbs, head, and facial expression, body coherence, and physical plausibility of object interactions. The prompt is available for reference.
    \item \textbf{Visual Prompt Following.} Which video better follows the prompt, ignoring the audio-related part of the description? The evaluation considers, in priority order, the number of objects, their quantities, properties, spatial arrangement, actions, and on-screen text.
    \item \textbf{Camera Motion.} Which video correctly reproduces the camera motion requested in the prompt? If several camera motions are requested, their count is compared. If the prompt does not specify any camera motion, the question is marked as not applicable.
\end{enumerate}

\textbf{Audio and audio-visual questions} (audio enabled):

\begin{enumerate}
    \setcounter{enumi}{5}
    \item \textbf{Speech Quality.} Which video has better speech quality? The annotator evaluates intelligibility, naturalness of voice and articulation, and, if singing is requested, whether the content is actually sung. Lip-sync and facial expression are evaluated only when the speaker is on screen.
    \item \textbf{Audio-Video Synchronization.} Which video has less desynchronization between audio and visuals? Temporal alignment is evaluated between sound events and their visual counterparts (e.g., speech and lip movement, impact and contact, door opening and sound). Desynchronization explicitly requested in the prompt is not penalized.
    \item \textbf{Overall Audio Quality.} Which video has better overall audio quality? The annotator evaluates audio defects such as muffled sound, extreme quietness, harsh clipping, unexpected interruptions, noise, and artifacts. The correspondence between audio and prompt is not evaluated here.
    \item \textbf{Sound Prompt Following.} Which video's audio better corresponds to the prompt? The annotator evaluates how accurately the requested sounds (speech, sound effects, background music, ambience) are reproduced. Missing requested sounds, wrong sounds, or wrong spoken text are counted as errors.
    \item \textbf{User Task Solving.} Which video better solves the user's task as a whole? The annotator evaluates visual and audio logic, coherence, and whether the generation can be used as a finished result. For contradictory prompts, the video offering the better creative synthesis is preferred.
\end{enumerate}

\textbf{Image-to-Video additional question}:

\begin{enumerate}
    \setcounter{enumi}{10}
    \item \textbf{Reference Image Correspondence.} Which video better preserves the input image? The annotator evaluates identity of characters, main objects and their attributes, clothing, colours, scene content, spatial arrangement, and visual style. Changes explicitly requested by the prompt, or required by the prompt, are not penalized.
\end{enumerate}

The number of questions per comparison depends on the mode: purely visual comparisons use questions 1--5 only; audio-visual comparisons add questions 6--10; image-to-video comparisons further add question 11.

\subsubsection{Kandinsky 6.0 Video Pro vs Kandinsky 5.0 Video Pro}

We compare \textbf{Kandinsky 6.0 Video Pro} against the previous generation, \textbf{Kandinsky 5.0 Video Pro}, in both text-to-video (T2V) and image-to-video (I2V) modes supported by Kandinsky 5.0. Audio is rendered, but ignored.

\begin{figure}[htbp]
    \centering
    \includegraphics[width=0.95\linewidth]{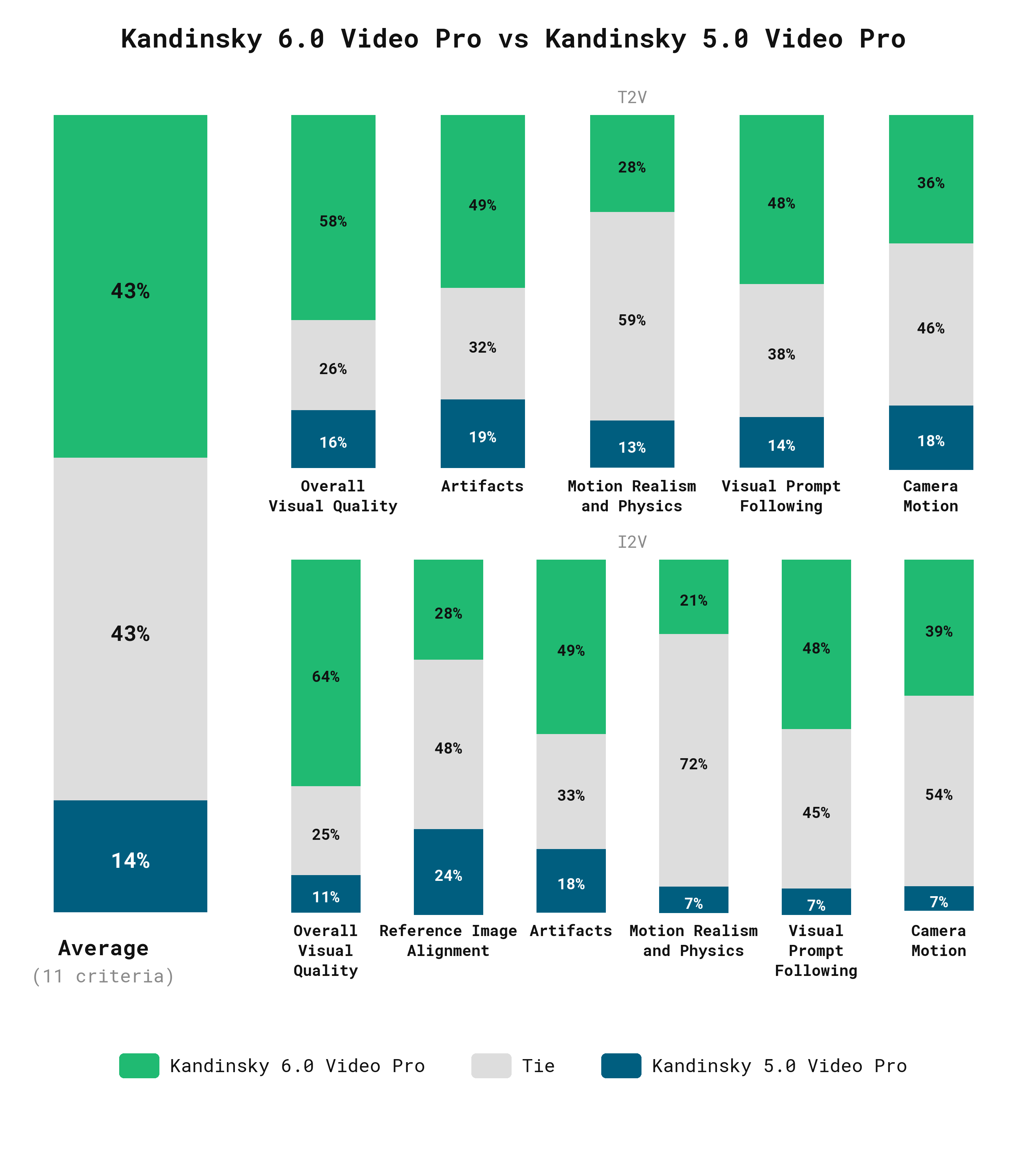}
    \caption{Side-by-side comparison of \textbf{Kandinsky 6.0 Video Pro} vs \textbf{Kandinsky 5.0 Video Pro}. Top row: Text-to-Video (T2V). Bottom row: Image-to-Video (I2V). In all figures of this section, the light green segment denotes preference for \textbf{Kandinsky 6.0 Video Pro}, gray denotes ties, and the dark segment denotes preference for the opponent model.}
    \label{fig:sbs-k6-vs-k5}
\end{figure}

The comparison (Figure~\ref{fig:sbs-k6-vs-k5}) shows a clear and consistent advantage for
\textbf{Kandinsky 6.0 Video Pro} across all perceptual and semantic criteria. In T2V mode (Figure~\ref{fig:sbs-k6-vs-k5}, top row), the new model dominates in overall visual quality, artifact suppression, motion realism, and visual prompt following, with a smaller but still positive advantage in camera motion. In I2V mode (Figure~\ref{fig:sbs-k6-vs-k5}, bottom row), the same pattern holds: Kandinsky 6.0 Video Pro leads in overall visual quality, artifacts, motion realism, visual prompt following, camera motion, and also improves on reference image correspondence --- a criterion that is otherwise challenging for video generation models. The magnitude of the gains is substantial for overall visual quality and prompt following, and moderate for motion realism.

\subsubsection{Kandinsky 6.0 Video Pro vs Kling 2.6}

We compare \textbf{Kandinsky 6.0 Video Pro} with \textbf{Kling 2.6} in T2AV and I2AV modes. In addition to the perceptual and semantic criteria, this comparison covers the full set of audio and audio-visual criteria.

\begin{figure}[htbp]
    \centering
    \includegraphics[width=0.95\linewidth]{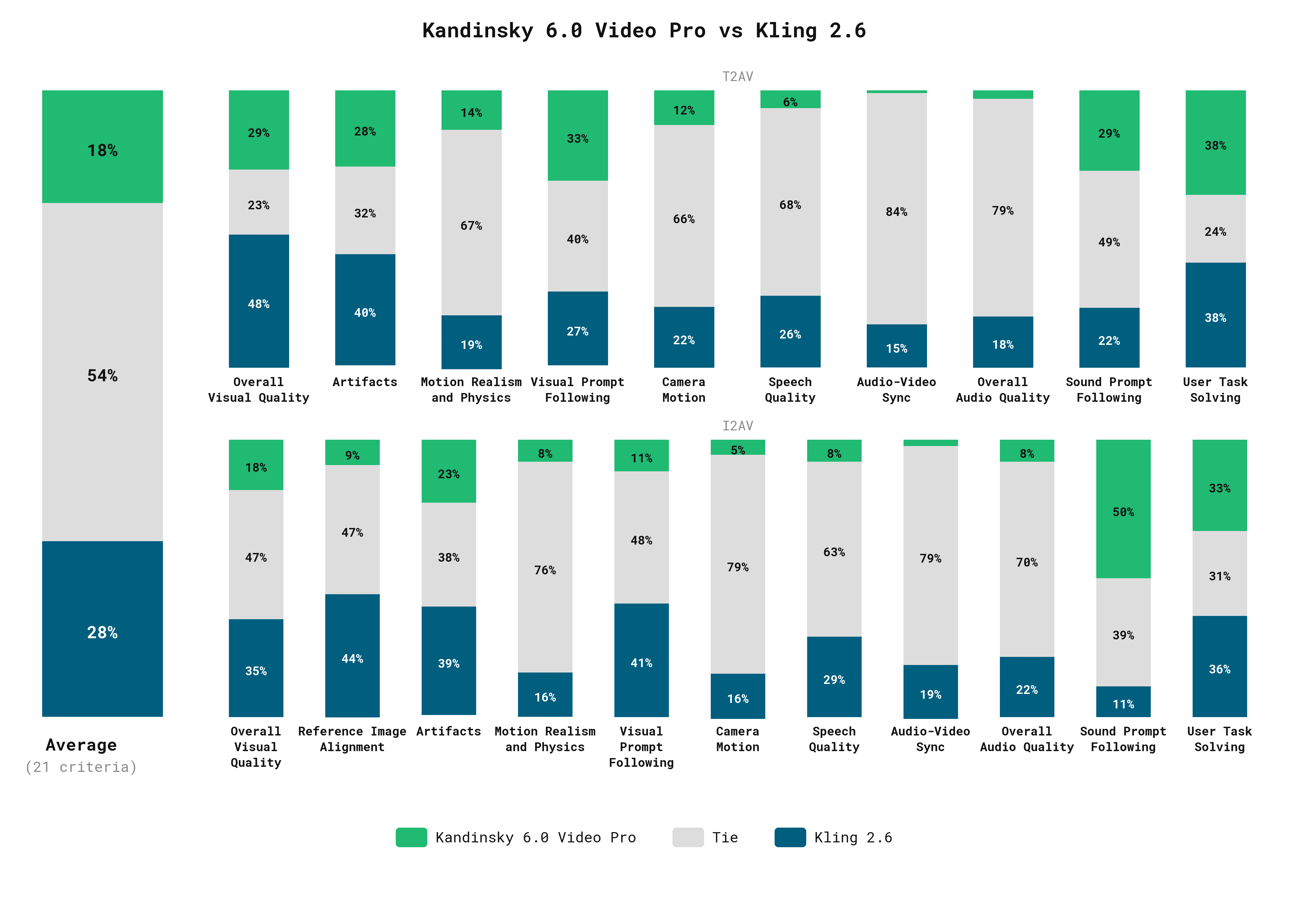}
    \caption{Side-by-side comparison of \textbf{Kandinsky 6.0 Video Pro} vs \textbf{Kling 2.6}. Top row: T2AV. Bottom row: I2AV.}
    \label{fig:sbs-k6-vs-kling}
\end{figure}

The results, presented in (Figure~\ref{fig:sbs-k6-vs-kling}), show a mixed picture. On perceptual criteria, \textbf{Kling 2.6} is preferred in overall visual quality and artifacts, while motion realism is close to parity, with most comparisons resulting in ties. On semantic criteria, visual prompt following slightly favors \textbf{Kandinsky 6.0 Video Pro} in T2AV mode and favors Kling 2.6 in I2AV mode, whereas \textbf{sound prompt following} favors \textbf{Kandinsky 6.0 Video Pro} in both modes, most clearly in I2AV mode. User-task solving is close to parity in both modes. On the audio side, the majority of comparisons for speech quality, audio-video synchronization, and overall audio quality result in ties, with the remaining preferences in favor of Kling 2.6. Overall, the two models are competitive: Kling 2.6 leads on visual aesthetics and audio quality, while Kandinsky 6.0 Video Pro leads on audio-prompt alignment.

\subsubsection{Kandinsky 6.0 Video Pro vs LTX 2.5}

We compare \textbf{Kandinsky 6.0 Video Pro} with \textbf{LTX 2.5} in T2AV and I2AV modes. As in the previous subsection, this comparison covers the full set of visual, semantic, audio, and audio-visual criteria.

\begin{figure}[htb]
    \centering
    \includegraphics[width=0.95\linewidth]{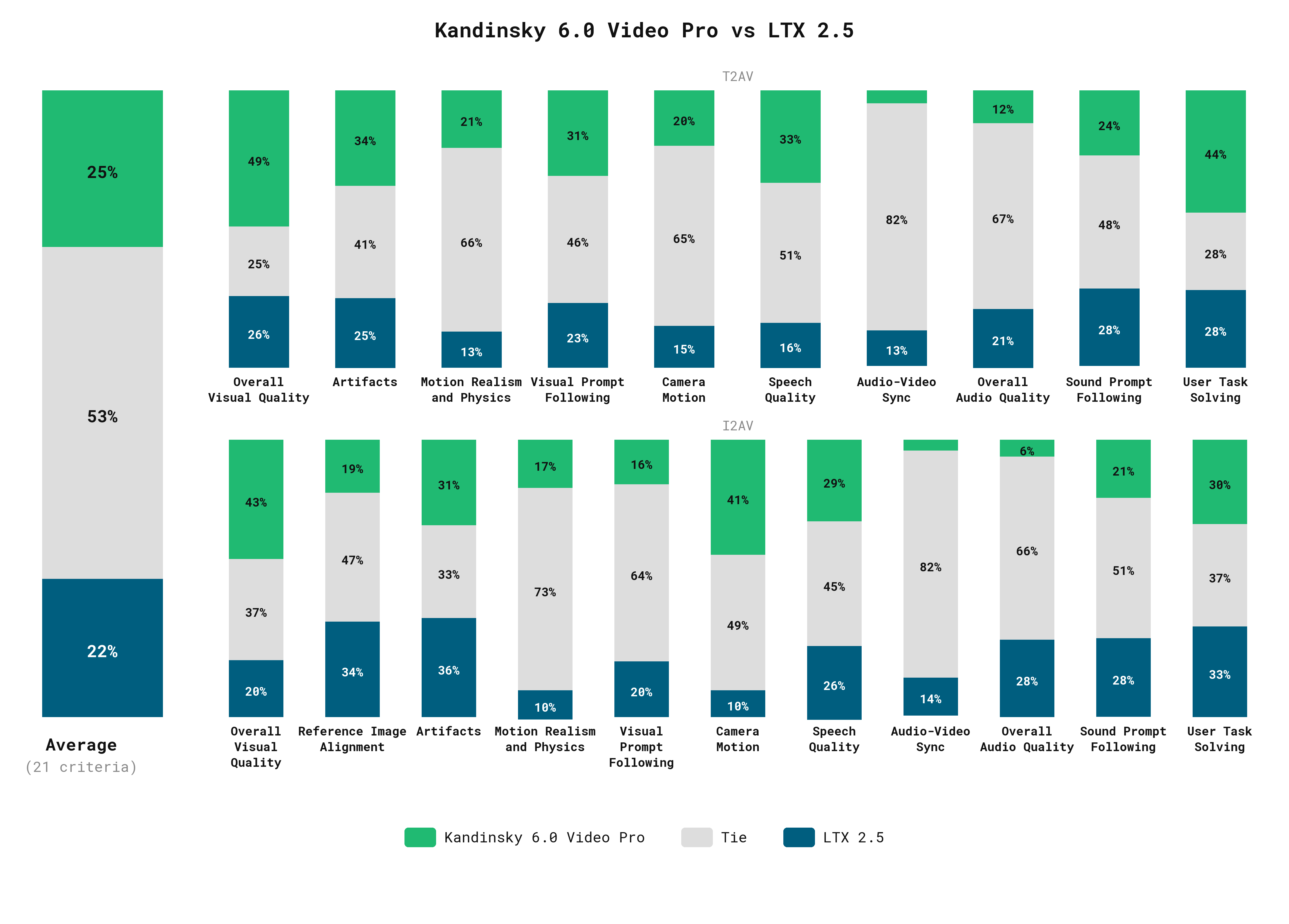}
    \caption{Side-by-side comparison of \textbf{Kandinsky 6.0 Video Pro} vs \textbf{LTX 2.5}. Top row: T2AV. Bottom row: I2AV.}
    \label{fig:sbs-k6-vs-ltx}
\end{figure}

The results are shown in (Figure~\ref{fig:sbs-k6-vs-ltx}). On the visual side, \textbf{Kandinsky 6.0 Video Pro} is clearly preferred in T2AV mode: it leads in overall visual quality, artifacts, motion realism, visual prompt following, and user-task solving, with the advantage statistically significant for overall visual quality and task solving. In I2AV mode, Kandinsky 6.0 Video Pro leads in overall visual quality, motion realism, and camera motion, while artifacts, visual prompt following, and user-task solving are close to parity. In T2AV mode, camera motion is close to parity, though the majority of prompts did not request any camera movement, so this criterion is informative for only a small subset of comparisons.

On the audio side, speech quality favors \textbf{Kandinsky 6.0 Video Pro} in both modes, with a statistically significant margin in T2AV mode, while audio-video synchronization, overall audio quality, and sound-prompt following are in favor of LTX 2.5, with a large fraction of comparisons resulting in ties. Overall, \textbf{Kandinsky 6.0 Video Pro} leads on the visual and speech-related dimensions, while LTX 2.5 retains an edge on general audio fidelity.

\subsubsection{Kandinsky 6.0 Video Pro vs Veo 3.1 Fast}

We compare \textbf{Kandinsky 6.0 Video Pro} with \textbf{Veo 3.1 Fast} in both T2AV and I2AV modes. The comparison covers the full set of criteria described above.

\begin{figure}[htb]
    \centering
    \includegraphics[width=0.95\linewidth]{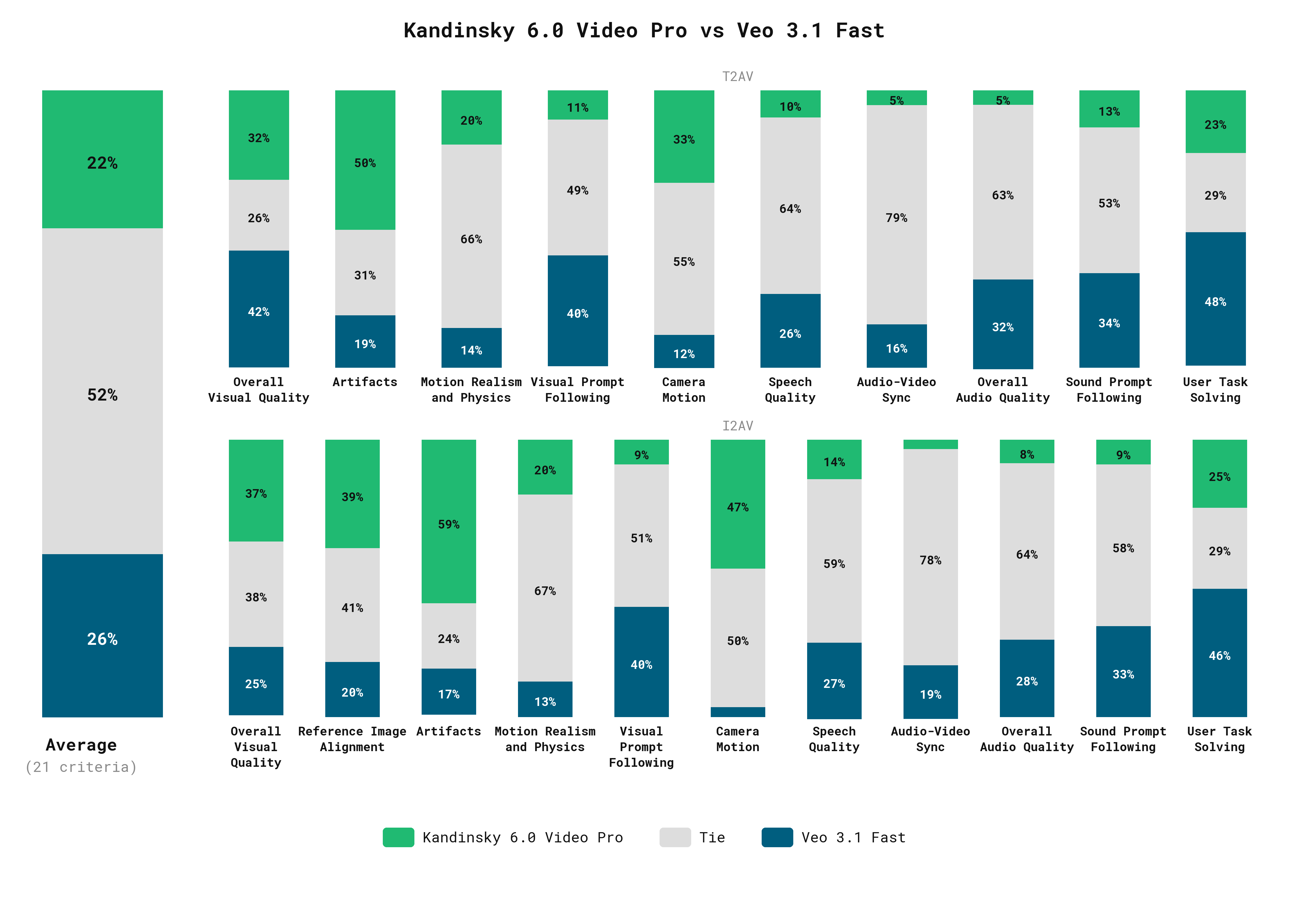}
    \caption{Side-by-side comparison of \textbf{Kandinsky 6.0 Video Pro} vs \textbf{Veo 3.1 Fast}. Top row: T2AV. Bottom row: I2AV. Light green --- Kandinsky 6.0 Video Pro, gray --- ties, dark --- Veo 3.1 Fast.}
    \label{fig:sbs-k6-vs-veo}
\end{figure}

Results are presented in (Figure~\ref{fig:sbs-k6-vs-veo}). In both modes, the two models trade strengths across different criteria, with no model dominating the entire set.

In \textbf{T2AV} mode (Figure~\ref{fig:sbs-k6-vs-veo}, top row), \textbf{Kandinsky 6.0 Video Pro} is preferred on artifacts and camera motion with statistically significant margins, and motion realism slightly favors it as well (though the sample is dominated by ties). \textbf{Veo 3.1 Fast} leads on visual prompt following, speech quality, audio-video synchronization, overall audio quality, sound-prompt following, and user-task solving --- all but speech reaching significance. Overall visual quality is close to parity with a slight edge to Veo.

In \textbf{I2AV} mode (Figure~\ref{fig:sbs-k6-vs-veo}, bottom row), \textbf{Kandinsky 6.0 Video Pro} is preferred on overall visual quality, reference-image alignment, artifacts, and camera motion --- all statistically significant. \textbf{Veo 3.1 Fast} leads on visual prompt following, audio-video synchronization, overall audio quality, sound-prompt following, and user-task solving --- also significant. Motion realism slightly favors Kandinsky 6.0 Video Pro, and for speech quality most comparisons result in ties.

Overall, the comparison reveals a clear split: \textbf{Kandinsky 6.0 Video Pro} produces cleaner visuals and better preserves the input image, while \textbf{Veo 3.1 Fast} better follows the prompt and delivers stronger audio.

\subsubsection{Kandinsky 6.0 Video Pro vs MiniMax H3}

We compare \textbf{Kandinsky 6.0 Video Pro} with \textbf{MiniMax H3} in both T2AV and I2AV modes. The comparison covers the full set of criteria described above.

\begin{figure}[htbp]
    \centering
    \includegraphics[width=0.95\linewidth]{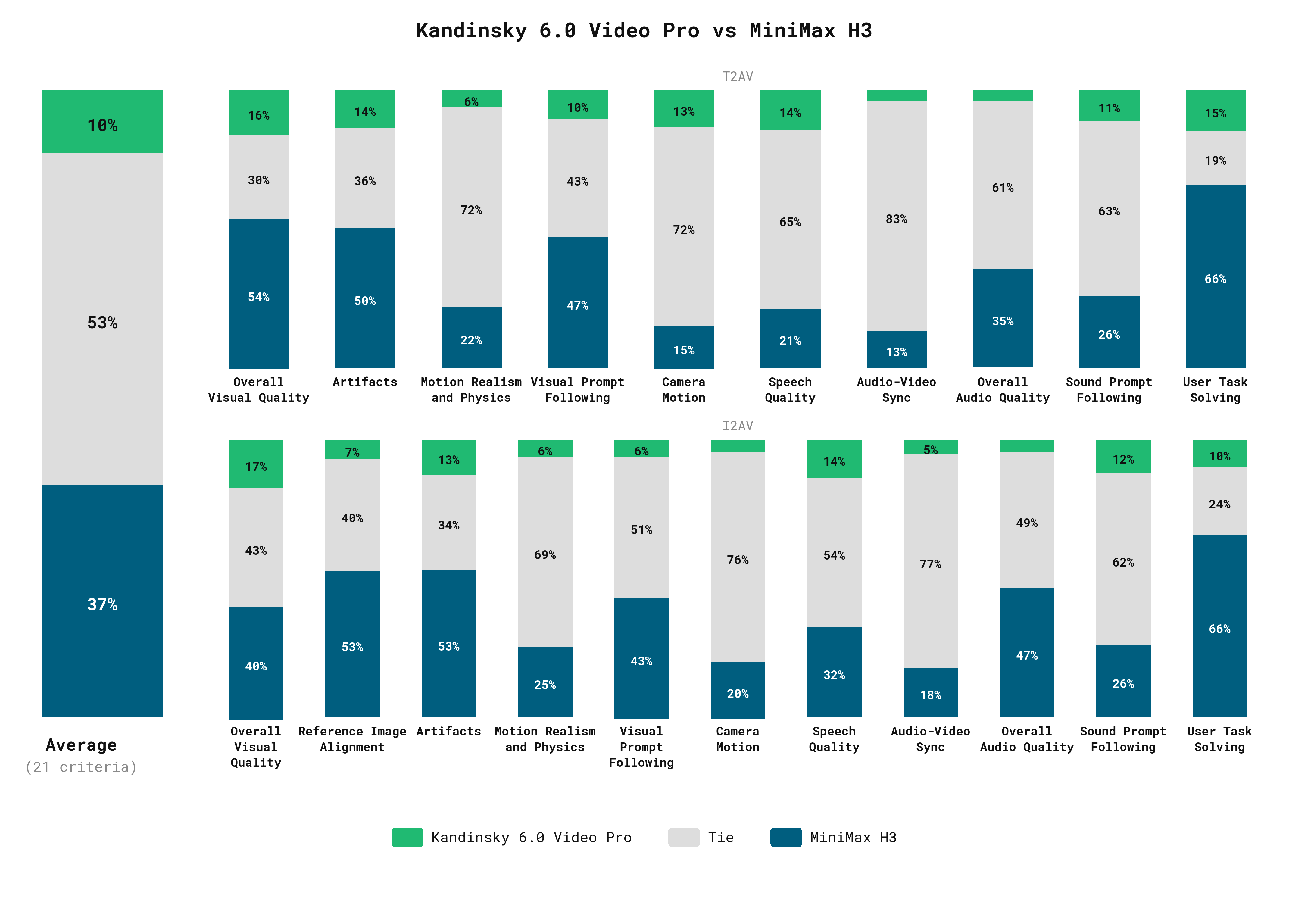}
    \caption{Side-by-side comparison of \textbf{Kandinsky 6.0 Video Pro} vs \textbf{MiniMax H3}. Top row: T2AV. Bottom row: I2AV. Light green --- Kandinsky 6.0 Video Pro, gray --- ties, dark --- MiniMax H3.}
    \label{fig:sbs-k6-vs-h3}
\end{figure}

Results are presented in (Figure~\ref{fig:sbs-k6-vs-h3}). \textbf{MiniMax H3} is preferred on the majority of criteria in both modes, with statistically significant margins on overall visual quality, artifacts, motion realism, visual prompt following, overall audio quality, sound-prompt following, and user-task solving. In \textbf{T2AV} mode, camera motion is close to parity, and for speech quality and audio-video synchronization the majority of comparisons result in ties.

The pattern is similar in \textbf{I2AV} mode: H3 leads on the visual and audio-visual criteria, while for camera motion, speech quality, and audio-video synchronization a large fraction of comparisons result in ties. As in T2AV, these criteria are applicable for only a subset of prompts.

Overall, MiniMax H3 is the stronger model on this benchmark; the closest result for Kandinsky 6.0 Video Pro is observed for camera motion in T2AV mode.

\subsubsection{Kandinsky 6.0 Video Pro vs Seedance 2.0}

We compare \textbf{Kandinsky 6.0 Video Pro} with \textbf{Seedance 2.0} in both T2AV and I2AV modes. The comparison covers the full set of criteria described above.

\begin{figure}[htbp]
    \centering
    \includegraphics[width=0.95\linewidth]{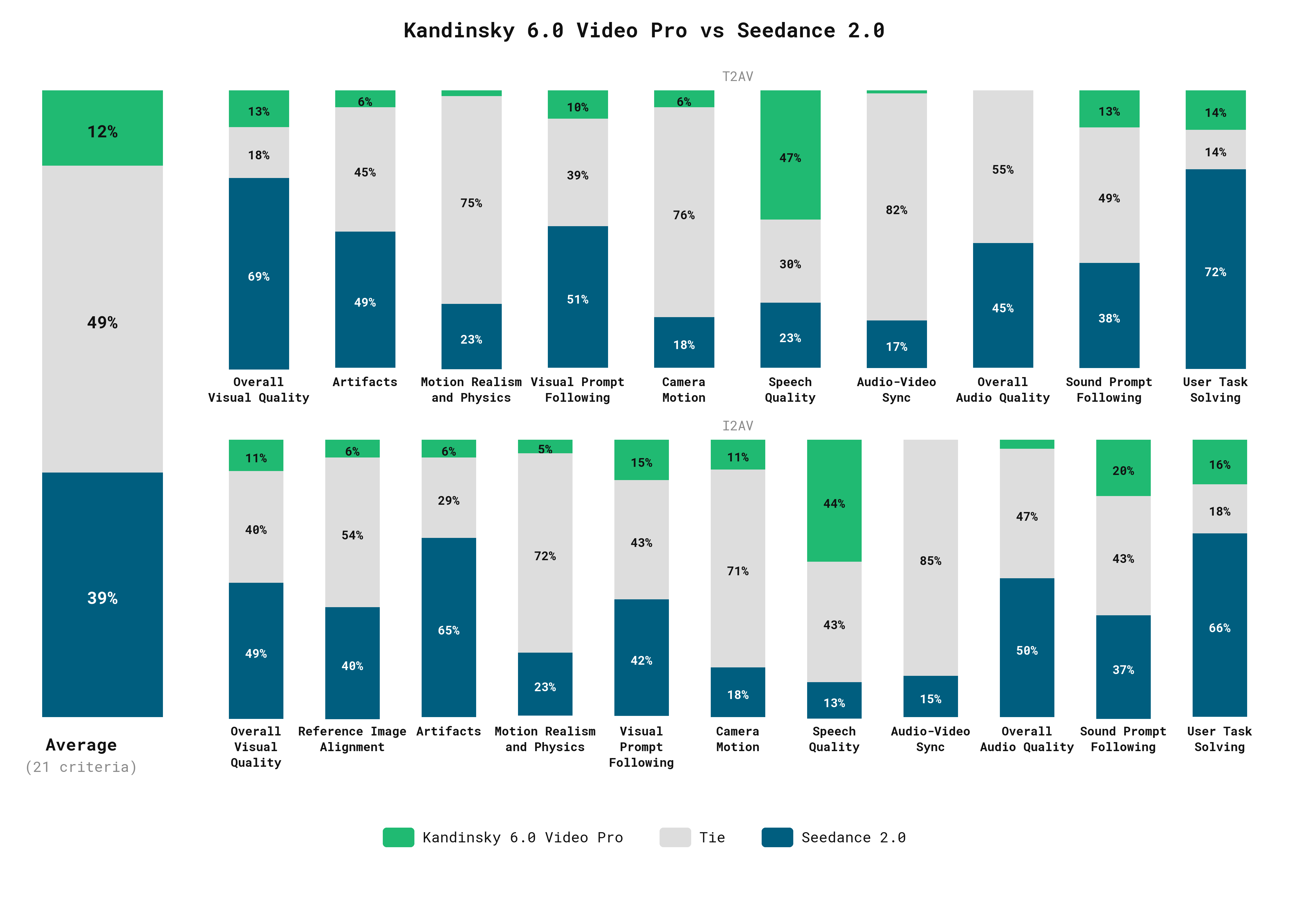}
    \caption{Side-by-side comparison of \textbf{Kandinsky 6.0 Video Pro} vs \textbf{Seedance 2.0}. Top row: T2AV. Bottom row: I2AV. Light green --- Kandinsky 6.0 Video Pro, gray --- ties, dark --- Seedance 2.0.}
    \label{fig:sbs-k6-vs-seedance}
\end{figure}

Results are presented in (Figure~\ref{fig:sbs-k6-vs-seedance}). \textbf{Seedance 2.0} is preferred on the visual criteria, with statistically significant margins on overall visual quality, artifacts, motion realism, and visual prompt following. In the audio domain, speech quality clearly favors \textbf{Kandinsky 6.0 Video Pro}, while for audio-video synchronization the large majority of comparisons result in ties.

In \textbf{I2AV} mode (Figure~\ref{fig:sbs-k6-vs-seedance}, bottom row), the pattern is similar. Seedance 2.0 leads on overall visual quality, reference-image alignment, artifacts, and visual prompt following, while for motion realism and audio-video synchronization the majority of comparisons result in ties. Speech quality again clearly favors \textbf{Kandinsky 6.0 Video Pro}.

Overall, Seedance 2.0 is the stronger model on purely visual criteria, while \textbf{Kandinsky 6.0 Video Pro} leads on speech quality in both modes. The comparison shows that the two models address different strengths: Seedance 2.0 focuses on visual fidelity, whereas Kandinsky 6.0 Video Pro performs particularly well when speech quality is the priority.

\begin{figure}[htbp]
    \centering
    \captionsetup{font=small, skip=4pt}
    \captionsetup[subfigure]{font=scriptsize, skip=0pt, justification=centering}
    \begin{subfigure}[t]{0.75\textwidth}
        \centering
        \includegraphics[width=\linewidth]{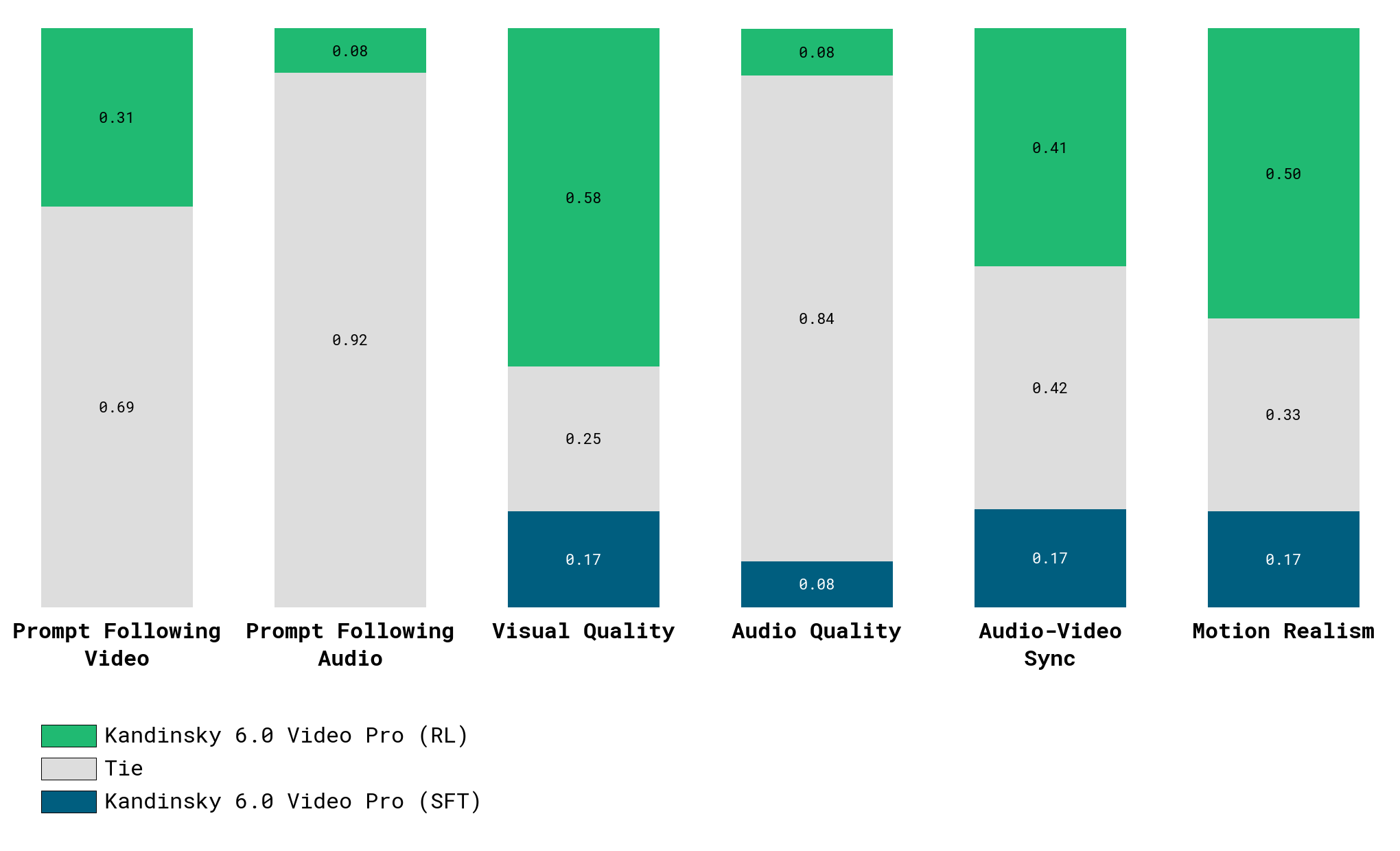}
        \caption{Text-to-Video + Audio, speech-focused}
    \end{subfigure}
    \\[0.2em]
    \captionsetup[subfigure]{font=scriptsize, skip=0pt, justification=centering}
    \begin{subfigure}[t]{0.75\textwidth}
        \centering
        \includegraphics[width=\linewidth]{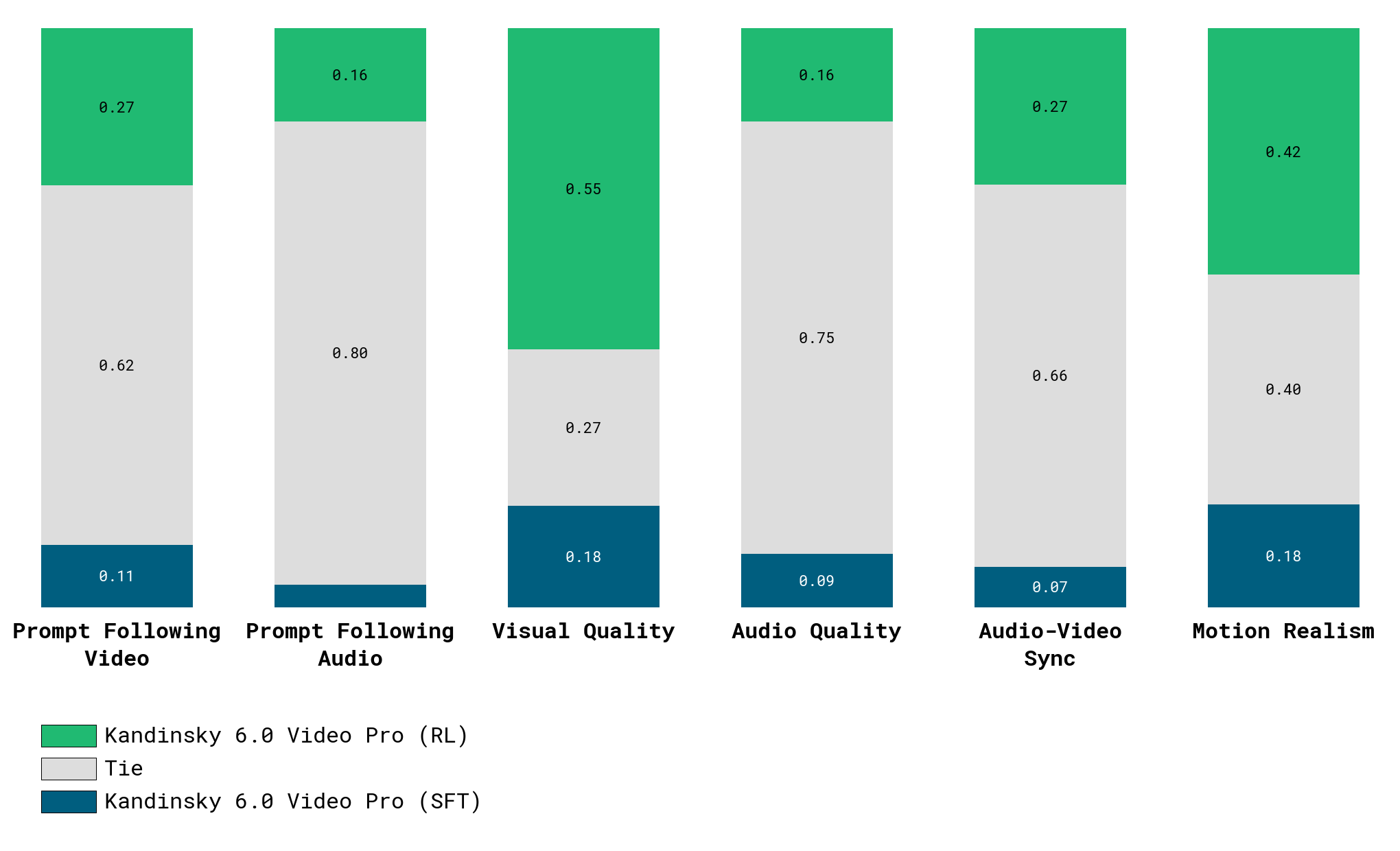}
        \caption{Text-to-Video + Audio, general domain}
    \end{subfigure}
    \\[0.2em]
    \captionsetup[subfigure]{font=scriptsize, skip=0pt, justification=centering}
    \begin{subfigure}[t]{0.75\textwidth}
        \centering
        \includegraphics[width=\linewidth]{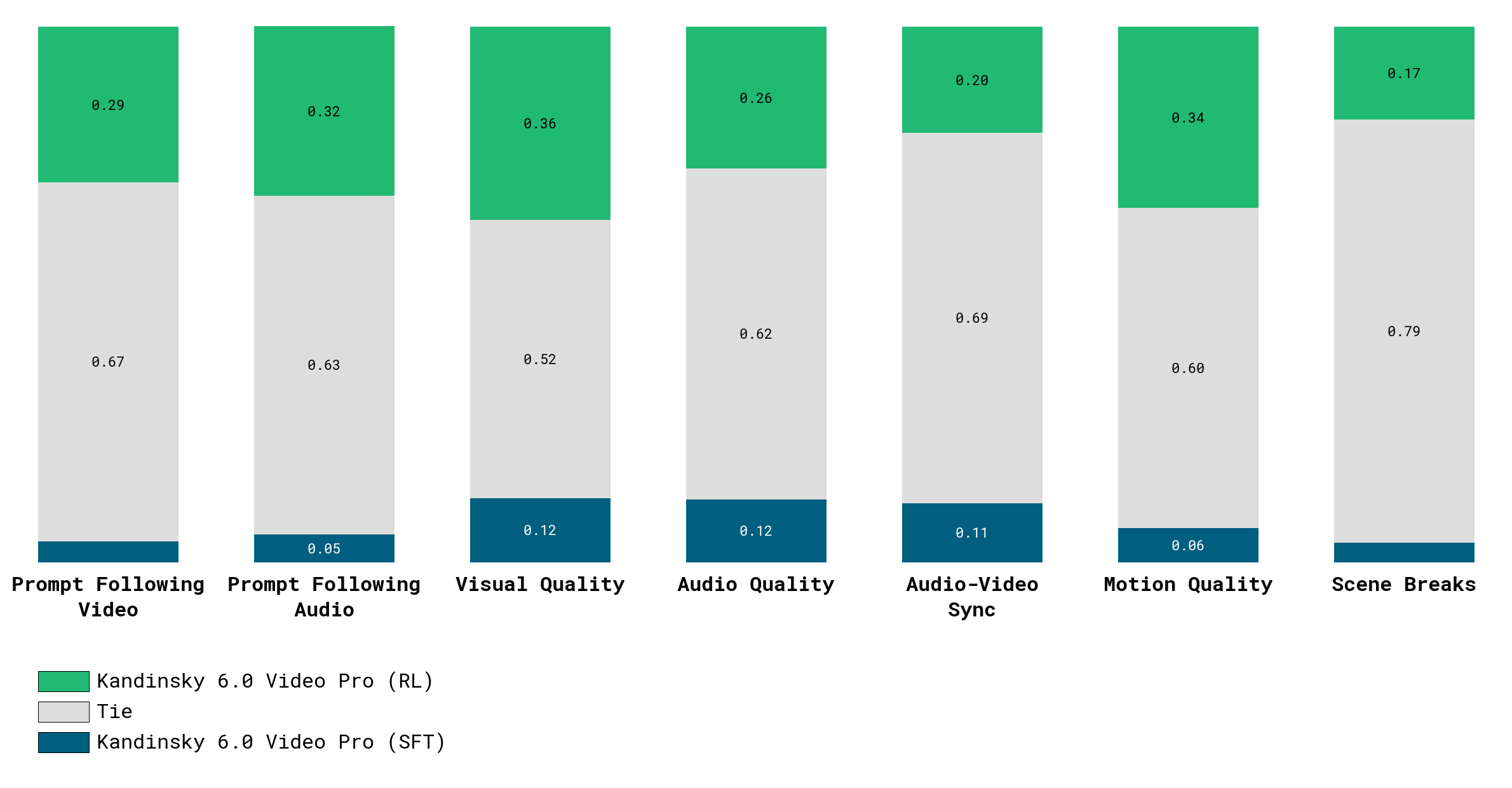}
        \caption{Image-to-Video + Audio, general domain}    
    \end{subfigure}
    \caption{Side-by-side comparison of \textbf{Kandinsky 6.0 Video Pro (RL)} vs \textbf{Kandinsky 6.0 Video Pro (SFT)} on three benchmarks (T2AV --- Text-to-Audio-Video, I2AV --- Image-to-Audio-Video). Light green --- RL, gray --- ties, dark --- SFT.}
    \label{fig:sbs-sft-vs-rl}
\end{figure}

\subsubsection{Ablation: Reinforcement Learning vs SFT}
\label{sec:ablation-rl-vs-sft}
We conduct an ablation study to isolate the effect of reinforcement learning post-training relative to the supervised fine-tuning stage. The comparison is performed in three settings: (a) T2AV mode on a curated speech-focused benchmark, (b) T2AV mode on a general-domain benchmark, and (c) I2AV mode on a general-domain benchmark. In all three settings, the \textbf{SFT} checkpoint of Kandinsky~6.0 Video Pro is compared against its \textbf{RL} checkpoint obtained after the reinforcement learning stage described in Section~\ref{sec:rl}.

Results are presented in (Figure~\ref{fig:sbs-sft-vs-rl}). The RL stage brings consistent and substantial improvements across all three settings.

\textbf{(a) T2AV, speech-focused benchmark.} Audio-video synchronization improves substantially (41\% vs 17\%), and audio prompt following shows a small gain with no comparisons in favor of SFT, although most of them are ties. The visual side benefits as well: visual quality, motion realism, and visual prompt following all improve. This indicates that, on speech-centric prompts, the RL stage improves the alignment between the audio and video components, which is one of the primary targets of the multi-objective reward stack, without degrading the visual side.

\textbf{(b) T2AV, general domain.} On the general-domain benchmark, the gains are more balanced. Visual quality, motion realism, and visual prompt following all improve substantially, while audio-related criteria show more moderate but still positive gains. This suggests that the RL stage does not specialize on speech at the expense of general-domain visual quality --- both dimensions improve, with the largest effect on the visual side.

\textbf{(c) I2AV, general domain.} In image-conditioned mode, the pattern is similar to (b): visual quality, motion quality, and audio-video synchronization improve, with prompt-following metrics also trending upward. Scene breaks decrease, indicating more stable, single-scene generations.

Across all three settings, the RL checkpoint is preferred by a clear margin, confirming that the reinforcement learning stage provides a meaningful improvement over SFT alone.

\subsubsection{Ablation: Distillation vs Full Model}
\label{sec:ablation-distill-vs-full}

\begin{figure}[htb]
    \centering
    \includegraphics[width=0.95\linewidth]{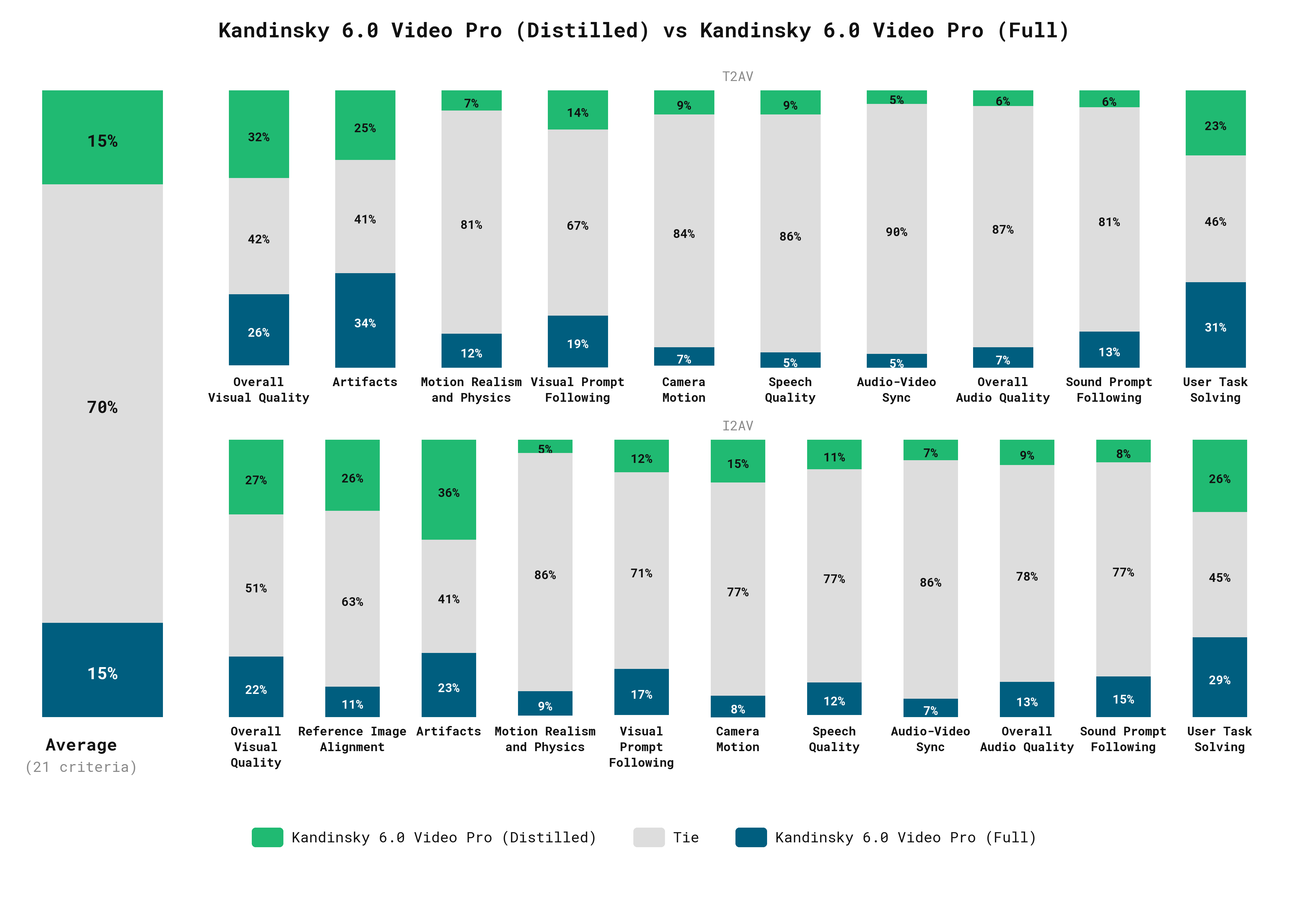}
    \caption{Side-by-side comparison of \textbf{Kandinsky 6.0 Video Pro (Distilled)} vs \textbf{Kandinsky 6.0 Video Pro (Full)}. Top row: T2AV. Bottom row: I2AV. Light green --- Distilled, gray --- ties, dark --- Full.}
    \label{fig:sbs-distill-vs-full}
\end{figure}

We compare the distilled checkpoint of Kandinsky~6.0 Video Pro against the full, non-distilled model of the same training stage in both T2AV and I2AV modes. Results are presented in (Figure~\ref{fig:sbs-distill-vs-full}).

In \textbf{T2AV} mode (Figure~\ref{fig:sbs-distill-vs-full}, top row), the two models perform on par across almost all criteria, with no individual difference reaching statistical significance at either the 0.95 or the 0.975 level. Here and below, the preference share of a model is computed with ties split equally between the two models, i.e., it equals the model's win rate plus half of the tie rate. The mean preference across all criteria is 51\% vs 49\% in favor of the full model, driven by a slight edge in artifacts and user task solving.

In \textbf{I2AV} mode (Figure~\ref{fig:sbs-distill-vs-full}, bottom row), the picture shifts in favor of the distilled model. The mean preference is 51\% vs 49\% for the distilled model. Notably, the distilled model shows statistically significant improvements (at the 0.975 level) on \textit{Reference Image Alignment} (57\% vs 43\%) and \textit{Artifacts} (57\% vs 43\%), indicating better preservation of the input image and cleaner generation. The full model retains a small, non-significant edge on motion realism, visual prompt following, overall audio quality, and sound-prompt following; the latter is the largest of these differences (15\% vs 8\% of comparisons, with 77\% ties).

Overall, distillation preserves the quality of the full model while reducing inference cost, and in the image-conditioned regime it even improves structural fidelity to the reference image.

\section[Examples]{\color{Green} Examples}\label{sec:use_cases}

This section presents qualitative results that are difficult to convey through
quantitative benchmarks alone. Video and synchronized audio generation cannot
be fully represented on the printed page: the reader cannot play back the
generated audio track or observe the temporal alignment between sound and
motion. We therefore rely on two complementary forms of illustration.

For audio-visual examples (Section~\ref{sec:examples_av}), each figure shows a
sequence of frames sampled from a single 5-second generation at 24\,fps,
together with the waveform of the corresponding audio track. The waveform makes
two properties visible at a glance. First, the audio is synchronized with the
video: impact sounds produce spikes at the exact frames where the impact
occurs, and speech amplitude follows the movements of the lips. Second, the
generated audio is semantically consistent with the visual scene: the model
produces the sound of the objects and actions it has generated, not generic
ambient noise. These properties are further supported by the VABench results reported in
Section~\ref{sec:evaluation}, where Kandinsky~6.0 Video Pro achieves the best
speech quality, lip-sync, and desynchronization scores among the evaluated
models, and by the speech quality results of the side-by-side human evaluation.

Super-resolution examples are provided in Section~\ref{sec:sr_examples}.

All examples in this section are generated by Kandinsky~6.0 Video Pro unless
stated otherwise. Examples of generated videos with their audio tracks are available in the
public release materials at \href{https://github.com/kandinskylab/kandinsky-6}{https://github.com/kandinskylab/kandinsky-6}.

\subsection{Audio-Visual Examples}
\label{sec:examples_av}

\begin{figure}[htbp]
    \centering
    \includegraphics[width=\textwidth]{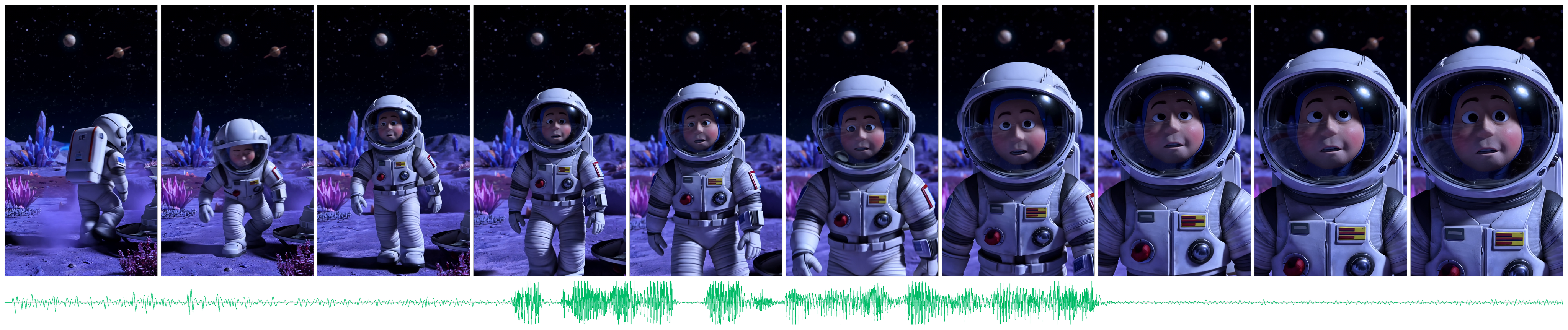}
    \caption{A 3D animated astronaut in a bulky white spacesuit walks across the
    surface of an alien planet with glowing purple crystals and two distant
    planets in the sky, turns to the camera, and speaks in awe:
    \texttt{<S>I have definitely never seen anything like this on Earth.<E>}
    \texttt{<AUDCAP>} Soft, muffled footsteps on dusty alien soil; a gentle
    ethereal hum with subtle crystalline chimes and a soft cinematic score.
    \texttt{<ENDAUDCAP>}}
    \label{fig:cosmos}
\end{figure}

\begin{figure}[htbp]
    \centering
    \includegraphics[width=\textwidth]{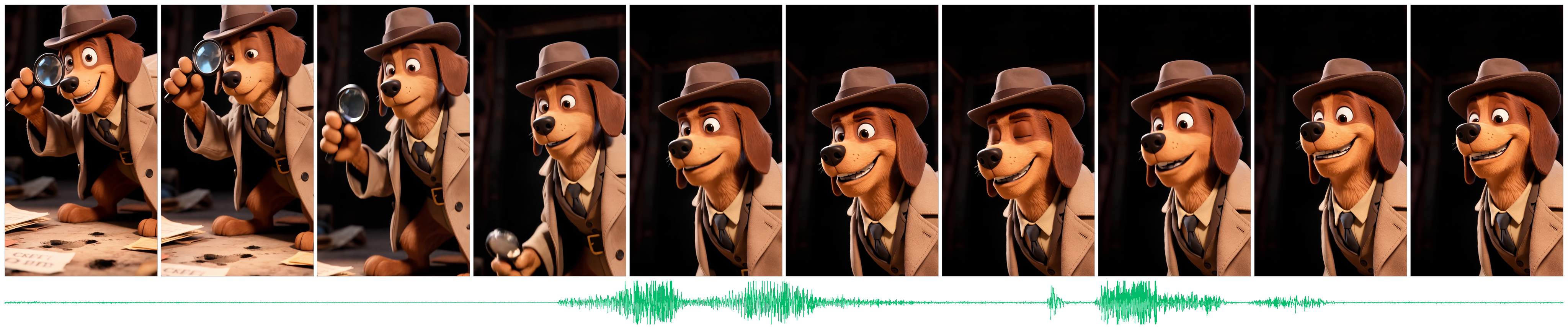}
    \caption{A cartoon dog detective in a trench coat and fedora examines a
    strange print through a magnifying glass, then looks up at the camera and
    says: \texttt{<S>I know who it is!<E>} \texttt{<AUDCAP>} Ambient night
    city sounds, distant traffic, a suspenseful underscore, and a sharp gasp
    of realization. \texttt{<ENDAUDCAP>}}
    \label{fig:dog}
\end{figure}

\begin{figure}[htbp]
    \centering
    \includegraphics[width=\textwidth]{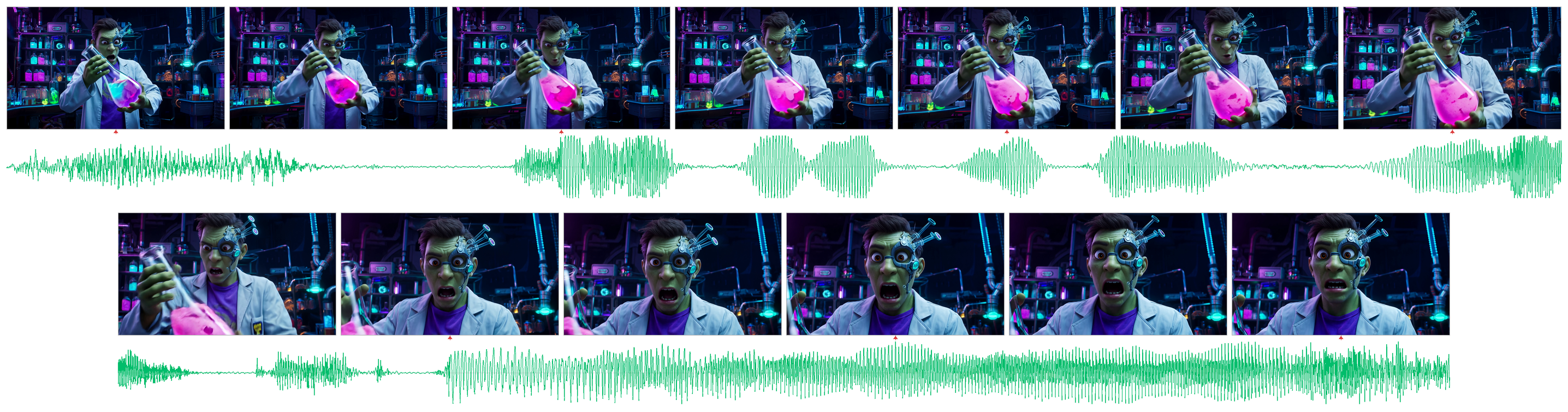}
    \caption{A cartoon mad scientist in a neon-lit laboratory mixes two glowing
    liquids in a flask; the mixture foams up, and he turns to the camera in
    panic: \texttt{<S>I mixed up the formula! How did this happen?!<E>}
    \texttt{<AUDCAP>} Hum of laboratory equipment, bubbling liquid, faint
    electrical zaps, vigorous fizzing, and a sharp gasp of panic.
    \texttt{<ENDAUDCAP>}}
    \label{fig:formula}
\end{figure}

\begin{figure}[htbp]
    \centering
    \includegraphics[width=\textwidth]{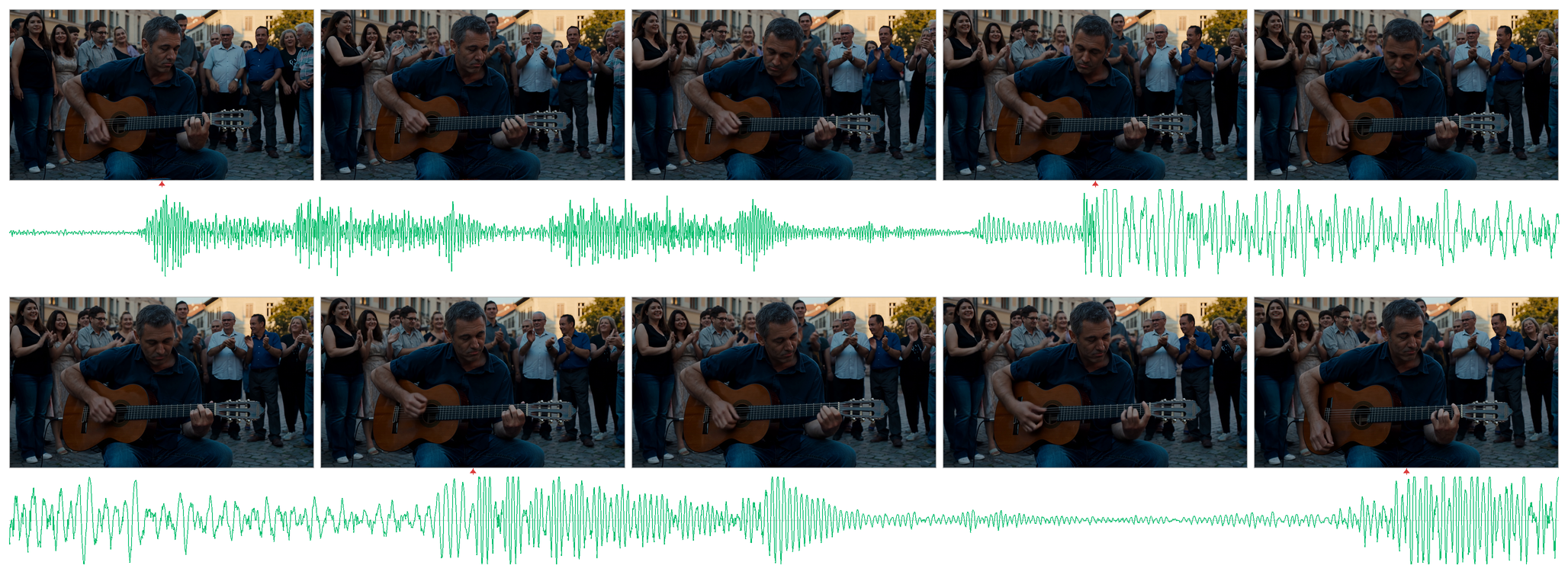}
    \caption{A street musician plays acoustic guitar on a cobblestone square
    in warm late-afternoon light while a small crowd claps along.
    \texttt{<AUDCAP>} Melodic fingerpicked acoustic guitar with rhythmic
    synchronized clapping and faint ambient sounds of a public square.
    \texttt{<ENDAUDCAP>}}
    \label{fig:guitar}
\end{figure}

\begin{figure}[htbp]
    \centering
    \includegraphics[width=\textwidth]{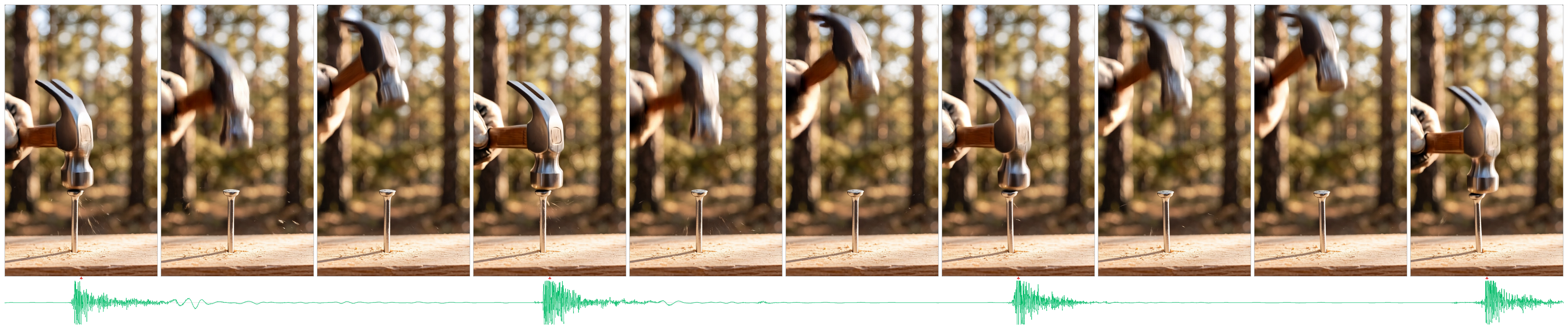}
    \caption{A gloved hand drives a metal nail into a wooden plank with a claw
    hammer in a sunlit forest. Each strike sends small wood chips flying.
    \texttt{<AUDCAP>} Loud metallic clinks and heavy resonant thuds at each
    strike, with faint forest ambience and birds in the distance.
    \texttt{<ENDAUDCAP>}}
    \label{fig:hammer_nail}
\end{figure}

\begin{figure}[htbp]
    \centering
    \includegraphics[width=\textwidth]{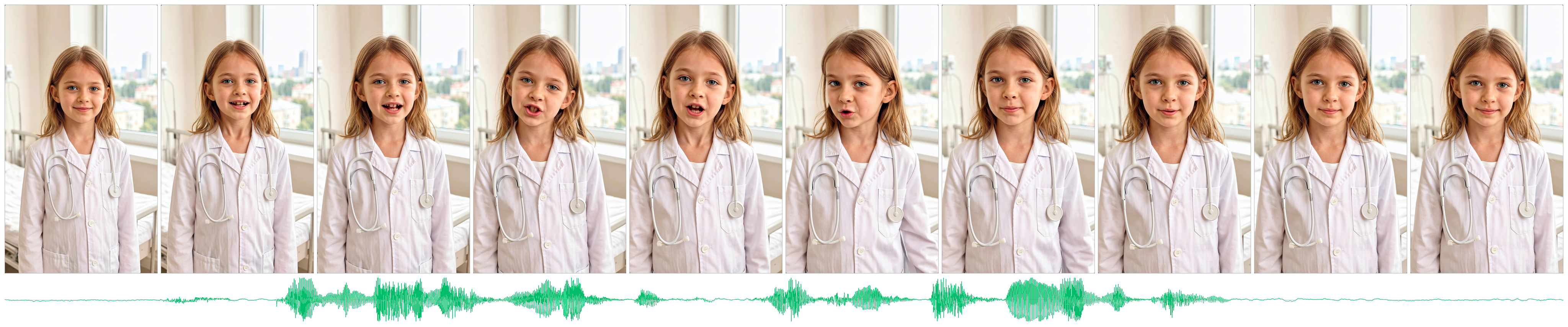}
    \caption{A young girl in a white medical coat with a stethoscope stands in
    a bright hospital room and speaks to the camera:
    \texttt{<S>When I grow up, I want to be a doctor!<E>}
    \texttt{<AUDCAP>} Quiet hospital room ambience, soft cheerful background
    music, and a clear, sweet, enthusiastic young voice.
    \texttt{<ENDAUDCAP>}}
    \label{fig:medic}
\end{figure}

\begin{figure}[htbp]
    \centering
    \includegraphics[width=\textwidth]{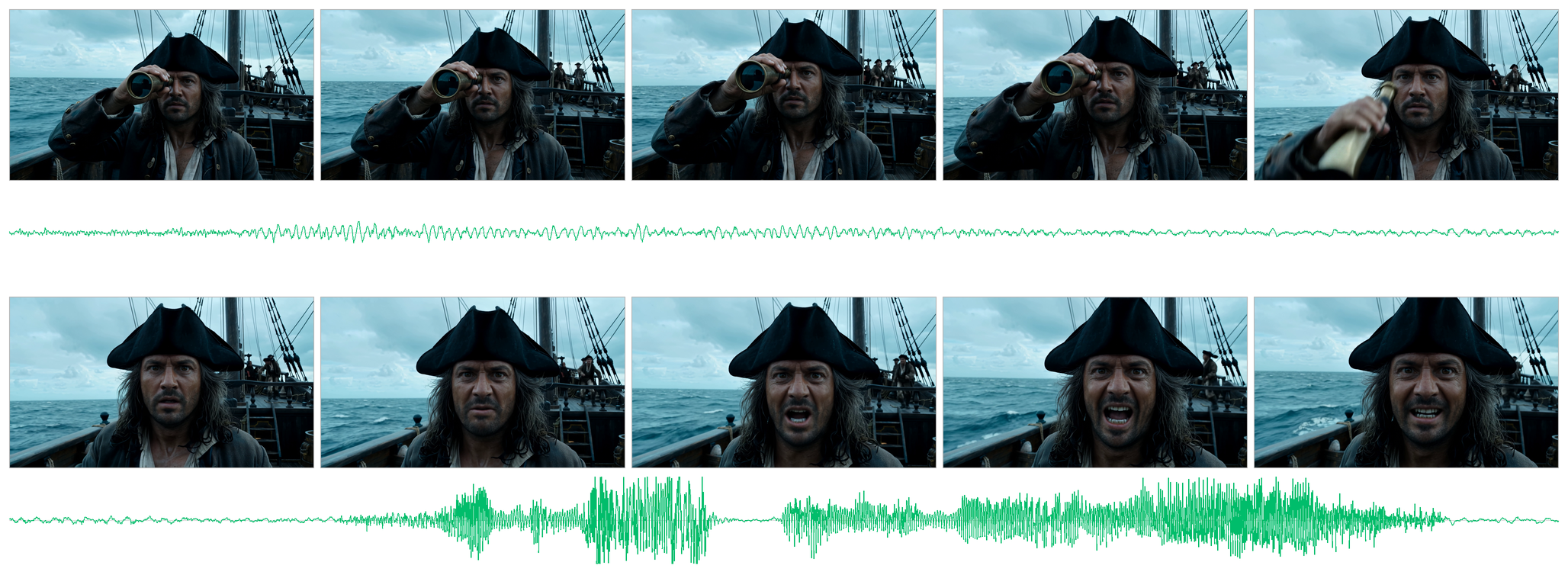}
    \caption{A pirate captain on the deck of a wooden sailing ship in rough
    seas lowers his spyglass and shouts toward the camera:
    \texttt{<S>Land ahead!<E>} \texttt{<AUDCAP>} Crashing waves against the
    hull, creaking planks and rigging, strong wind, and a loud triumphant
    shout. \texttt{<ENDAUDCAP>}}
    \label{fig:pirate}
\end{figure}

Figures~\ref{fig:cosmos}--\ref{fig:pirate} show examples in
the general and speech domains, covering animated scenes, speaking characters,
and impact-synchronized sound.


\clearpage
\subsection{Super-Resolution Examples}\label{sec:sr_examples}

\begin{figure}[htbp]
    \centering
    \includegraphics[width=\textwidth]{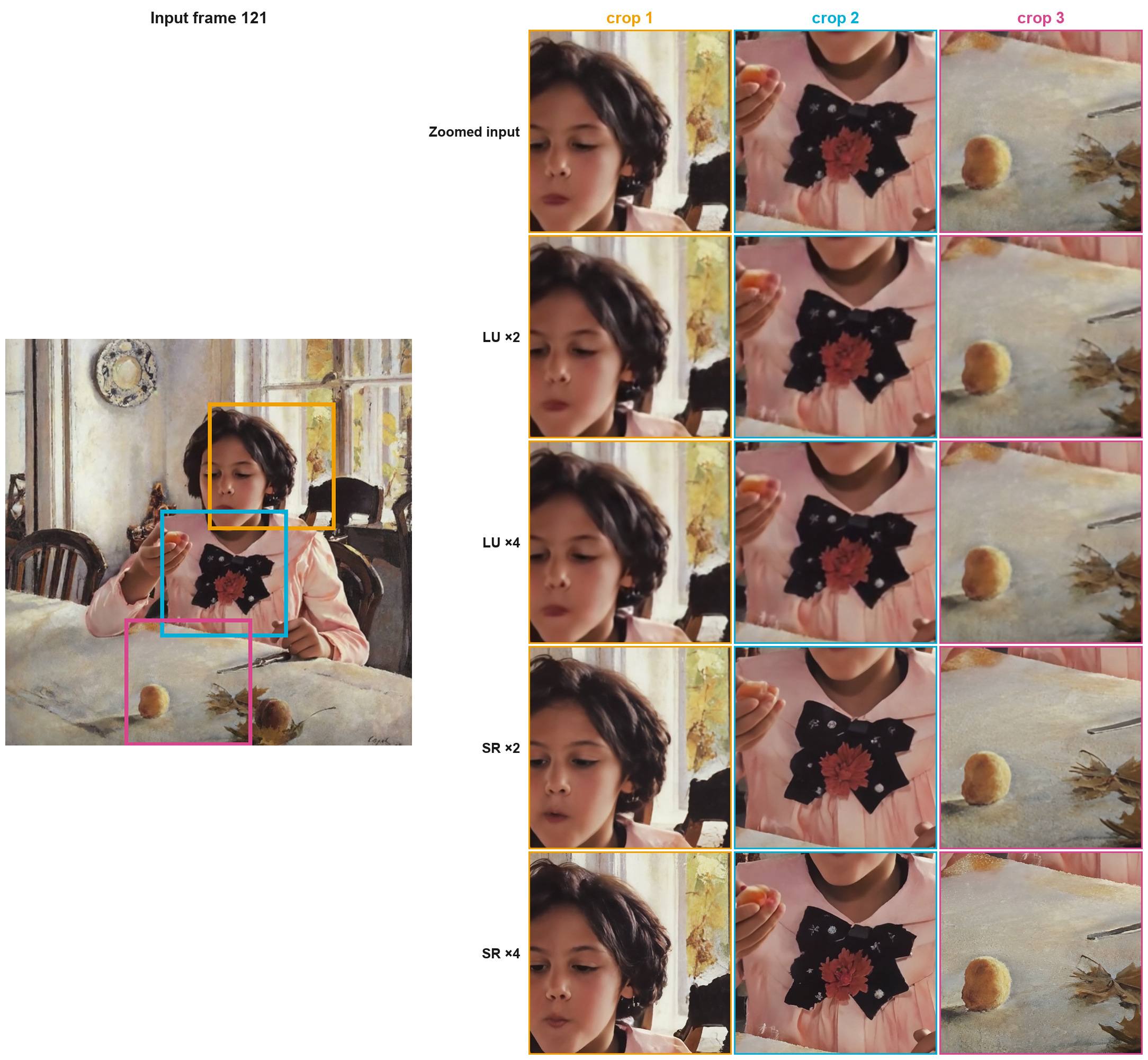}
    \caption{An animated painting based on Valentin Serov's
    \emph{Girl with Peaches} (1887); the original painting is in the public
    domain. The crops show the face and hair, the bow on the dress, and a
    fruit on the table. LU outputs remain visually close to the enlarged
    input. SR sharpens hair strands and adds texture to the bow, fruit, and
    tablecloth, with finer detail at $\times4$ than at $\times2$.}
    \label{fig:sr_133}
\end{figure}

\begin{figure}[htbp]
    \centering
    \includegraphics[width=\textwidth]{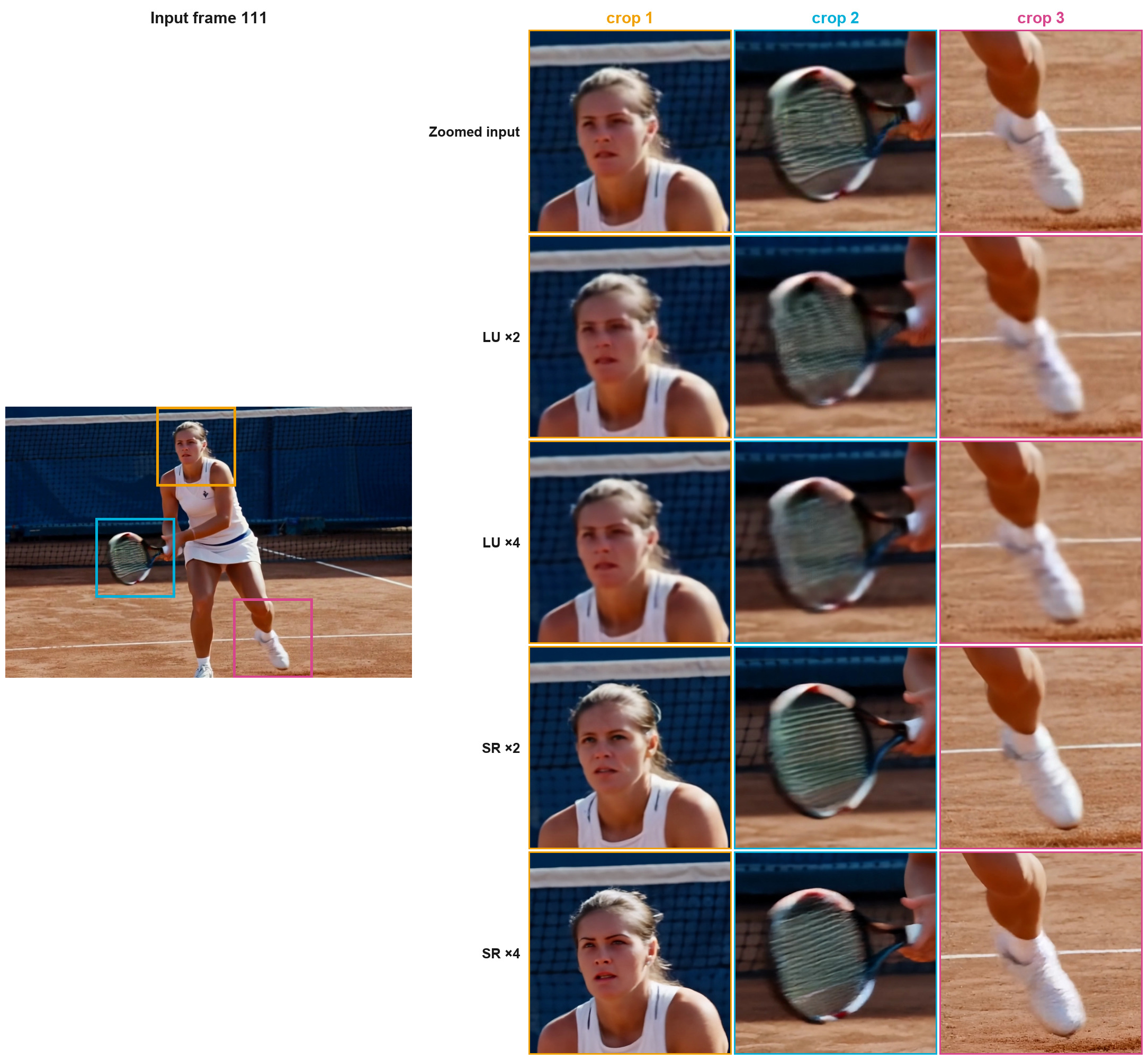}
    \caption{A tennis scene, frame 111.
The crops show the face, the racket, and the shoe with
the surrounding court surface.
LU outputs remain visually close to the enlarged input.
SR makes facial features, racket strings, and shoe and
court textures more distinct, particularly at $\times4$.}
    \label{fig:sr_122}
\end{figure}

\begin{figure}[htbp]
    \centering
    \includegraphics[width=\textwidth]{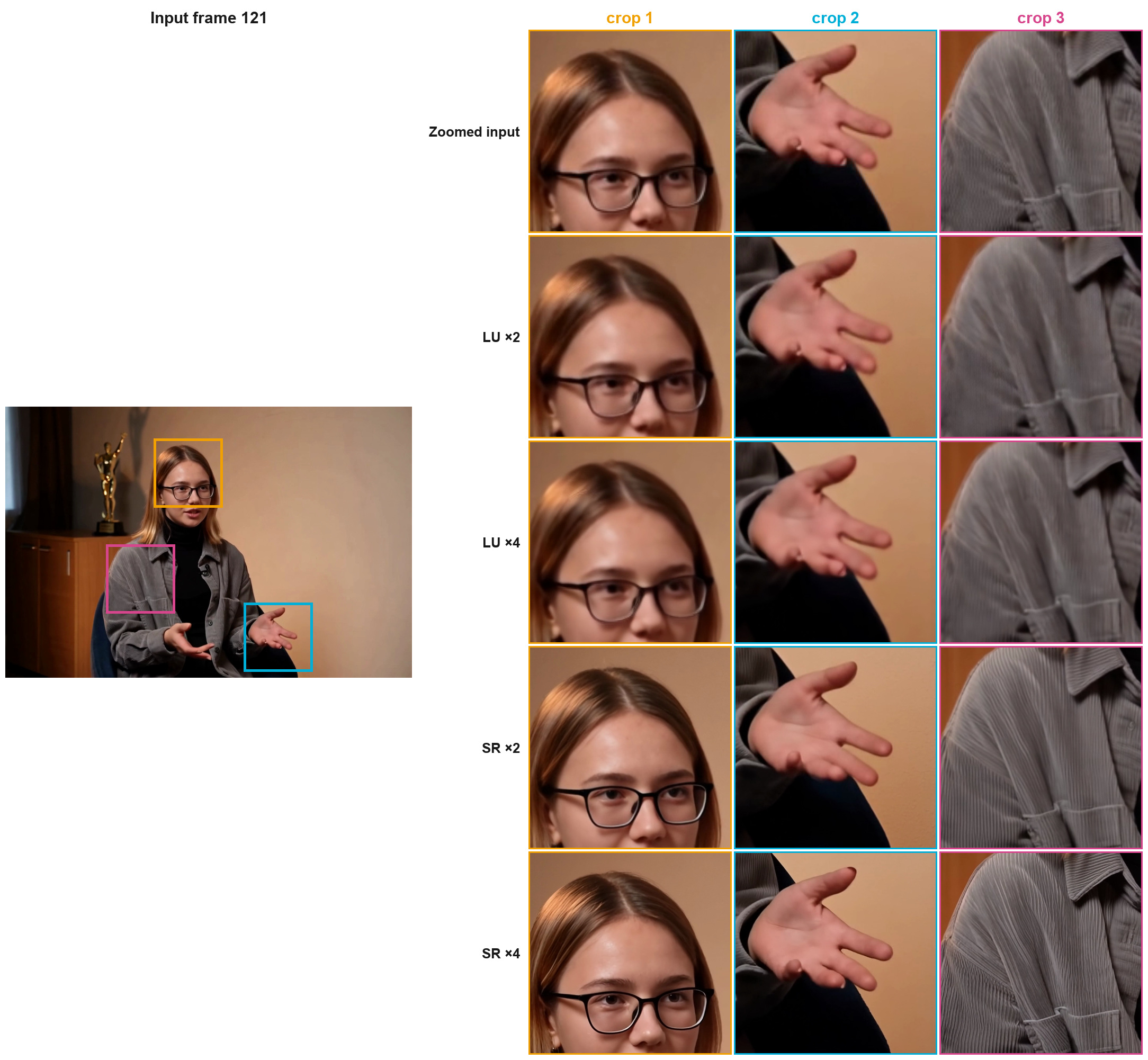}
    \caption{An interview scene, frame 121.
The crops show the face and glasses, the hand,
and the fabric of the jacket.
LU outputs remain visually close to the enlarged input.
SR sharpens facial features and glasses contours and
enhances hair and fabric detail, with more pronounced
texture at $\times4$ than at $\times2$.}
    \label{fig:sr_1004}
\end{figure}

Figures~\ref{fig:sr_133}, \ref{fig:sr_122}, and \ref{fig:sr_1004}
illustrate the effect of super-resolution on selected crops
from generated videos.
Each figure shows an input frame on the left and three
corresponding crop regions on the right.
From top to bottom, the rows show the enlarged input,
decoded LU outputs at $\times2$ and $\times4$, and outputs
of the full SR pipeline at $\times2$ and $\times4$.
The LU outputs remain visually close to the enlarged input.
The full SR pipeline produces sharper boundaries and finer
textures, with the $\times4$ route showing more pronounced
detail than the $\times2$ route in these examples.

\clearpage

\section[Related Work]{Related Work}\label{sec:related}

\subsection{Image and Video Generation}

Diffusion models~\cite{ho2020ddpm,song2021scorebased} and flow matching~\cite{lipman2023flow,liu2022flowstraightfastlearning} have become the dominant paradigm for visual generation. Latent diffusion~\cite{rombach2022high} made high-resolution synthesis computationally tractable, and transformer backbones~\cite{peebles2023dit,esser2024scalingrectifiedflowtransformers} have largely replaced convolutional U-Nets in recent systems. The Kandinsky family of text-to-image models followed this trajectory over several generations~\cite{razzhigaev2023kandinsky,arkhipkin2023kandinsky30technicalreport,arkhipkin2024kandinsky}.

Early video diffusion models extended image generators with temporal layers~\cite{ho2022imagenvideohighdefinition,blattmann2023align,blattmann2023stable}, while more recent systems train spatio-temporal diffusion transformers at scale~\cite{ma2025latte,yang2024cogvideox,genmo2024,kong2024hunyuanvideo,wan2025,polyak2024movie,seaweed2025seaweed7bcosteffectivetrainingvideo}. Within the Kandinsky line, architectural design choices for text-to-video pipelines were studied in~\cite{arkhipkin2023fusionframesefficientarchitecturalaspects,arkhipkin2025improveyourvideos}, followed by the CrossDiT-based Kandinsky~5.0 family~\cite{kandinsky5}, on which Kandinsky~6.0 Video directly builds. Other works address individual components of video generation pipelines, including efficient attention for long spatio-temporal sequences~\cite{mikhailov2026nabla,zhang2024fast,xi2025sparsevideogenacceleratingvideo}, autoencoder training without characteristic artifacts~\cite{novitskiy2025vivat}, explicit temporal control in video diffusion transformers~\cite{kuklev2026time}, and temporally correlated generative priors for video-to-video tasks~\cite{vasilev2026tcvbm}. Culture-specific generation and its evaluation are supported by dedicated datasets and benchmarks~\cite{vasilev2024craft,vasilev2025ruscode}, which underlie the Russian Cultural Code data used in this work (Section~\ref{sec:data_processing_pipeline}).

\subsection{Joint Audio-Video Generation}

Audio and video have long been generated separately or sequentially, for example by synthesizing a soundtrack for a given video~\cite{mmaudio}. Closed commercial systems such as Veo~3.1~\cite{Veo} and Sora~2~\cite{openai2024sora2} demonstrated that both modalities can be generated jointly with natural lip synchronization and sound effects. Open research models pursue the same goal with a range of designs: temporally aligned flow-based generation~\cite{syncflow2024}, hierarchical spatio-temporal synchronization priors~\cite{javis2025}, unified tri-modal transformers~\cite{3mdit2025}, and dual-stream transformers with cross-modal fusion~\cite{low2025ovi,ltx2025}. Large-scale systems such as ALIVE~\cite{alive2025}, MiniMax-H3~\cite{MiniMax}, and Cosmos~3~\cite{cosmos3} further extend joint audio-video generation. Kandinsky~6.0 Video follows the dual-stream design but initializes the video stream from a pretrained video model and trains the audio stream from scratch (Section~\ref{sec:architecture}). Joint audio-video models can also be extended with additional conditioning, such as reference-based voice cloning~\cite{mikheev2026voice}. Progress in this area is measured by dedicated benchmarks for audio-video quality and synchronization~\cite{vabench2025,t2avcompass2025,avgenbench2025}.

\subsection{Post-Training and Few-Step Sampling}

Supervised fine-tuning on curated data and averaging of fine-tuned weights (model soups)~\cite{wortsman2022modelsoup,biggs2024diffusionsoupmodelmerging} are common ways to improve the quality of pretrained generators. Reinforcement learning further aligns diffusion and flow models with human preferences and reward models~\cite{black2023trainingdiffusionmodelsreinforcement,wallace2023diffusionmodelalignmentusing,liu2025flowgrpotrainingflowmatching,liu2025improvingvideogenerationhuman}; OmniNFT~\cite{zhang2026omninft} adapts this approach to joint audio-video generation. To reduce the number of sampling steps, prior work has explored progressive distillation~\cite{salimans2022progressive}, consistency models~\cite{song2023consistency,luo2023latent,pcm}, distribution matching~\cite{dmd,dmd2,sid}, adversarial distillation~\cite{sauer2023adversarialdiffusiondistillation,ladd,APT}, and policy-based imitation distillation~\cite{piflow}.

\subsection{Video Super-Resolution}

Real-world super-resolution models are commonly trained on synthetic degradations that imitate blur, noise, resizing, and compression~\cite{wang2021realesrgan}. Diffusion-based approaches add strong generative priors and reduce the number of sampling steps~\cite{wu2025vpdsr,wang2026eliminating}, and recent work addresses structural and temporal artifacts specific to generated videos~\cite{panambur2025creativevr} as well as temporally consistent video-to-video restoration~\cite{vasilev2026tcvbm}. Our SR model operates in latent space and combines a deterministic latent upscaler with a text-free diffusion transformer distilled to two network evaluations (Section~\ref{sec:SR_architecture}). %
\section[Conclusion]{Conclusion}\label{sec:conclusion}

We presented Kandinsky~6.0 Video, a family of foundation models for synchronized text-to-audio-video and image-to-audio-video generation, comprising Kandinsky~6.0 Video Lite (3B parameters) and Kandinsky~6.0 Video Pro (29B parameters). Both models use a dual-stream CrossDiT architecture in which a video stream inherited from Kandinsky~5.0 and an audio stream trained from scratch are connected by blockwise bidirectional cross-attention. The models are trained with continuous pre-training, first separately for each modality and then jointly, followed by domain-specific supervised fine-tuning merged into a model soup, reinforcement learning adapted from OmniNFT, and two-stage distillation ($\pi$-Flow followed by adversarial refinement) to 10 function evaluations. A separate latent-space super-resolution model, consisting of a Latent Upscaler and a text-free SR-DiT distilled to two network evaluations, raises the output resolution to Full HD. Block offloading and memory presets make both models usable on consumer GPUs with 16--32~GB of memory.

In side-by-side human evaluation, Kandinsky~6.0 Video Pro clearly outperforms its predecessor, Kandinsky~5.0 Video Pro, and is preferred over LTX~2.5 in visual quality and speech quality. Compared with proprietary systems, the results are mixed: the model is competitive with Kling~2.6 and Veo~3.1 Fast, with strengths that differ by criterion, while MiniMax~H3 and Seedance~2.0 remain stronger on visual criteria; in speech quality, Kandinsky~6.0 Video Pro is preferred over Seedance~2.0. On VABench, Kandinsky~6.0 Video Pro achieves the best scores among the evaluated models (Kandinsky~6.0 Video Lite and LTX~2.5) in speech quality, audio aesthetics, text-video and audio-video alignment, lip-sync, desynchronization, and visual realism. Reinforcement learning reduces the word error rate of generated speech by 47\% for the Pro model.

The models have several limitations. They generate clips of up to 5~seconds, the base generation runs at SD resolution with Full HD obtained through super-resolution, and a gap remains to the strongest proprietary systems in visual quality and overall audio quality. Natural directions for future work include longer clips, higher native resolution, and additional reference-based conditioning.

We release the model weights, source code, and diffusers integration under the MIT license to support further research on multimodal generation.

\subsection*{Ethical Considerations}

Kandinsky~6.0 Video is intended to support research, education, and creative work, including content production, prototyping, and accessible tools for visual and audio storytelling. We believe that openly available generative models should be developed and used responsibly, with respect for human rights, personal dignity, and applicable law.

Models that produce realistic video with synchronized speech and lip movements carry specific risks. They can be misused to create misleading or deceptive content, to impersonate real individuals, to spread disinformation, or to facilitate fraud and harassment. We do not support and explicitly discourage any such use. In particular, the models should not be used to depict real people without their consent, to imitate a person's voice or appearance for deceptive purposes, to generate content that sexualizes minors or depicts non-consensual intimate imagery, to promote violence, hatred, or discrimination, or to infringe intellectual property rights.

We encourage users and downstream developers to clearly label generated content as synthetic, to comply with the laws and regulations of their jurisdictions, including those governing personal data, image and voice rights, and the disclosure of AI-generated media, and to implement appropriate safeguards, such as content moderation, when deploying the models in public-facing products.

Like other models trained on large-scale data, Kandinsky~6.0 Video may reflect biases present in its training data, including uneven representation of demographic groups, cultures, and languages. Its outputs may also contain factual, physical, or cultural inaccuracies and should not be treated as depictions of real events. We encourage the research community to study these limitations and to report problematic behavior. %
\clearpage

\section[Contributors and Acknowledgments]{Contributors and Acknowledgments}\label{sec:contributors}

\subsection*{Contributors}

\textbf{Core Contributors:}
\begin{itemize}
    \item \textbf{Pre-training:} Anna Dmitrienko, Alexander Ivanov, Denis Koposov, Eugeny Vilisov

    \item \textbf{Post-training:} Nikita Kiselev, Georgy Makarov, Ivan Mikheev, Alexander Sentsov, Sergey Taranov

    \item \textbf{Optimization:} Sofiya Bogakovskaya, Dmitry Ermilov, Mikhail Karaev, Anastasiia Kargapoltseva, Anatoliy Parpara

    \item \textbf{Distillation:} Kirill Kozlov

    \item \textbf{Super-resolution:} Vladimir Dokholyan, Georgii Fedorov, Aleksandr Gordeev

    \item \textbf{VAE:} David Chikovani, Valeria Kobenko, Kirill Malakhov

    \item \textbf{Captioning Pipeline:} Alexander Belykh, Maxim Litvinov

    \item \textbf{Metrics \& Evaluation:} Mikhail Aksyutin, Kirill Buzygin

    \item \textbf{Pre-training Data:} Kirill Chernyshev, Vladimir Chernyy, Elizaveta Muromtseva, Konstantin Reznikov, Azat Saginbaev, Artem Sherstyuk, Matvey Suprunov

    \item \textbf{Post-training Data:} Anastasia Aliaskina, Mark Bulygin, Irina Cheremnykh, Sergey Emelyanov, Andrey Inozemtsev, Vladislav Lakhtionov, Egor Malykh, Nikolai Nazarkin, Olga Varlashina

    \item \textbf{Harness:} Bulat Akhmatov, Anton Bukashkin, Mikhail Mamaev, Polina Mikhailova

    \item \textbf{Technical Report:} Vladimir Polovnikov

    \item \textbf{Track Leaders:} Julia Agafonova, Andrei Ivaniuta, Ivan Kirillov, Yury Kolabushin, Anatoly Korobov, Denis Krzhivokolskiy, Alexey Letunovskiy, Dmitrii Mikhailov, Lev Novitskiy, Nikita Samsonov, Andrey Shutkin, Irina Tolstykh, Viacheslav Vasilev

    \item \textbf{Project Supervisors:} Vladimir Arkhipkin, Denis Parkhomenko, Vladimir Korviakov, Nikolai Gerasimenko, Denis Dimitrov
\end{itemize}

\paragraph{Contributors:} Grigorii Alekseenko, Olga Androsova, Anna Averchenkova, Serafima Bocharova, Mikhail Chernyshov, Georgy Daniltsev, Polina Gavrilova, Maria Kovaleva, Konstantin Kuklev, Alexander Kunitsyn, Sergey Kuzin, Alexander Lyulkov, Tatiana Nikulina, Stanislav Onuchin, Nikita Osterov, Nikita Shaimov, Egor Silvestrov, Bulat Suleimanov, Tatiana Trofimuk, Ilya Trushkin, Aleksandra Tsybina, Ilya Vasiliev, Sergey Yakubson, Konstantin Zakharov

\subsection*{Acknowledgments}

We would like to thank the following people for their help, advice, and support:

\paragraph{TagMe Team:} Sergey Gorelov, Olga Pavlova, Dmitry Popov, Alexey Potapov, Alexander Potemkin

\paragraph{GigaData Team:} Andrey Berezin, Andrey Evtikhov, Yulduz Fattakhova, Vyacheslav Fedotov, Alexey Sandakov, Sergey Volkov

\paragraph{Cloud.ru:} Alena Drobyshevskaia, Nikita Kochetkov, Sergey Kovylov, Alyona Kuznetsova, Alexander Naumov, Vladislav Shidlovsky, Viktor Strelkov

\paragraph{Model Risk Management Team:} Alexey Isakov, Viktor Panshin, Stepan Ponomarev, Vladislav Rodionov, Oleg Yangalichin

\paragraph{Kandinsky Office:} Alexey Bondarenko, Vladimir Kozlov, Margarita Kusakina, Elizaveta Raubishko

\vspace{0.6cm}

as well as Anton Frolov, Andrey Korystin, and Fyodor Minkin. %

\bibliographystyle{unsrt}
\bibliography{references}

\end{document}